\documentclass[10pt,twocolumn,letterpaper]{article}

\usepackage{iccv}
\usepackage{times}
\usepackage{epsfig}
\usepackage{graphicx}
\usepackage{amsmath}
\usepackage{amssymb}
\usepackage{subcaption}
\usepackage{enumitem}

\usepackage{comment}
\usepackage{pifont}
\usepackage{booktabs}
\usepackage{multirow}
\usepackage{color, colortbl}
\usepackage[dvipsnames]{xcolor}
\usepackage{float}

\usepackage[breaklinks=true,bookmarks=false]{hyperref}

\usepackage[capitalize]{cleveref}
\crefname{section}{Sec.}{Secs.}
\Crefname{section}{Section}{Sections}
\Crefname{table}{Table}{Tables}
\crefname{table}{Tab.}{Tabs.}

\iccvfinalcopy %

\def\iccvPaperID{10882} %

\ificcvfinal\fi

\begin{document}

\title{\textbf{\textit{SuS-X}}: Training-Free Name-Only Transfer of Vision-Language Models}

\author{Vishaal Udandarao\\
University of Cambridge\\
{\tt\small vu214@cam.ac.uk}
\and
Ankush Gupta\\
DeepMind, London\\
{\tt\small ankushgupta@google.com}
\and
Samuel Albanie\\
University of Cambridge\\
{\tt\small sma71@cam.ac.uk}
}
\maketitle
\ificcvfinal\thispagestyle{empty}\fi

\begin{abstract}
Contrastive Language-Image Pre-training (CLIP) has emerged as a simple yet effective way to train large-scale vision-language models. CLIP demonstrates impressive zero-shot classification and retrieval performance on diverse downstream tasks.
However, to leverage its full potential, fine-tuning still appears to be necessary.
Fine-tuning the entire CLIP model can be resource-intensive and unstable.
Moreover, recent methods that aim to circumvent this need for fine-tuning still require access to images from the target task distribution.
In this paper, we pursue a different approach and explore the regime of training-free ``name-only transfer'' in which the only knowledge we possess about the downstream task comprises the names of downstream target categories. 
We propose a novel method, \textbf{\textit{SuS-X}}, consisting of two key building blocks---``SuS'' and ``TIP-X'', that requires neither intensive fine-tuning nor costly labelled data.
\textit{\textbf{SuS-X}} achieves state-of-the-art (SoTA) zero-shot
classification results on 19 benchmark datasets.
We further show the utility of TIP-X in the training-free few-shot setting, where we again achieve SoTA results over strong training-free baselines.
Code is available at \href{https://github.com/vishaal27/SuS-X}{https://github.com/vishaal27/SuS-X}.

\end{abstract}

\section{Introduction}
\label{sec:intro}

Vision-language pre-training has taken the machine learning community by storm.
A broad range of vision-language
models (VLMs)~\cite{radford2021learning, li2022blip, yao2021filip, alayrac2022flamingo, jia2021scaling} exhibiting exceptional transfer on tasks like classification~\cite{zhang2022tip, zhou2022learningcoop}, cross-modal retrieval~\cite{udandarao2020cobra, bain2022clip} and segmentation~\cite{shin2022reco, ghiasi2021open} have emerged.
These models are now the \textit{de facto} standard for downstream task transfer in the field of computer vision.

\begin{figure}[!ht]
    \centering
    \includegraphics[width=0.45\textwidth]{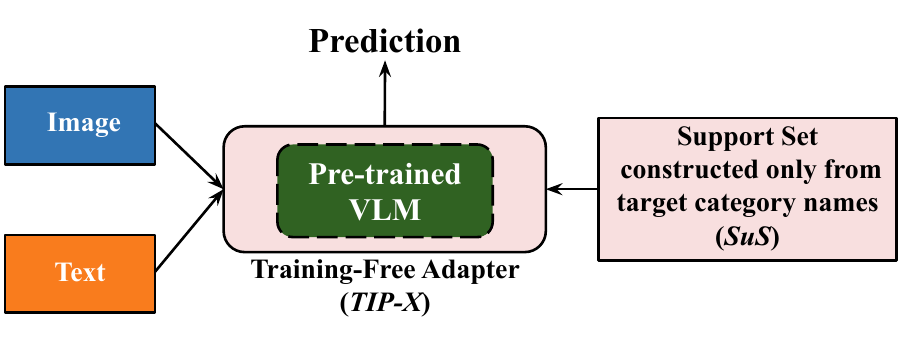}
    \caption{
    \textbf{Training-free name-only transfer.}
    We propose \textbf{\textit{SuS-X}}, a framework for enhancing the zero-shot transfer abilities of VLMs like CLIP~\cite{radford2021learning}, BLIP~\cite{li2022blip} and TCL~\cite{yang2022visiontcl}, without training.
    To achieve this, we propose a novel method \textit{TIP-X}, which adapts these VLMs using a curated \textit{support set} (\textit{SuS}) that is \textit{not drawn} from the target distribution. Our
    \textit{SuS} leverages one key piece of information about the task at hand: the names of the target categories.
    }
    \label{fig:teaser}
    \vspace{-3mm}
\end{figure}

\setlength{\tabcolsep}{3pt}
\begin{table*}[!ht]
  \footnotesize
  \centering
    \caption{\textbf{Taxonomy of CLIP adaptation methods for downstream classification.} We underline the Zero-Shot CLIP model to signify that it is the base model that all others build on top of. $^*$This method considers access to all test-set samples simultaneously, hence we still consider it zero-shot. $^\dagger$This method additionally uses class hierarchy maps.
    } 
  \begin{tabular}{clccc}
    \toprule
    & \multirow{2}{*}{\textbf{Method}} & \textbf{Does not require} & \textbf{Does not require} & \textbf{Does not require} \\
    & & \textbf{training} & \textbf{labelled data} & \textbf{target data distribution} \\
    \midrule
    \multirow{10}{*}{\begin{minipage}[t]{0.2\textwidth}
          \centering
         \textit{Few-shot fine-tuning}\\
          \textit{methods}
        \end{minipage}} & LP-CLIP~\cite{radford2021learning} & \textcolor{black}{\ding{55}} & 
    \textcolor{black}{\ding{55}} & 
    \textcolor{black}{\ding{55}} \\
    & CoOp~\cite{zhou2022learningcoop} & \textcolor{black}{\ding{55}} & 
    \textcolor{black}{\ding{55}} & 
    \textcolor{black}{\ding{55}} \\
    & PLOT~\cite{chen2022promptplot} & \textcolor{black}{\ding{55}} & 
    \textcolor{black}{\ding{55}} & 
    \textcolor{black}{\ding{55}} \\
    & LASP~\cite{bulat2022languagelasp} & \textcolor{black}{\ding{55}} & 
    \textcolor{black}{\ding{55}} & 
    \textcolor{black}{\ding{55}} \\
    & SoftCPT~\cite{ding2022promptsoftcpt} & \textcolor{black}{\ding{55}} & 
    \textcolor{black}{\ding{55}} 
    & \textcolor{black}{\ding{55}} \\
    & VT-CLIP~\cite{zhang2021vtclip} & \textcolor{black}{\ding{55}} & 
    \textcolor{black}{\ding{55}} 
    & \textcolor{black}{\ding{55}} \\
    & VPT~\cite{derakhshani2022variational} & \textcolor{black}{\ding{55}} & 
    \textcolor{black}{\ding{55}} & 
    \textcolor{black}{\ding{55}} \\
    & ProDA~\cite{proda} & 
    \textcolor{black}{\ding{55}} & 
    \textcolor{black}{\ding{55}} & 
    \textcolor{black}{\ding{55}} \\
    & CoCoOp~\cite{zhou2022conditional} & \textcolor{black}{\ding{55}} & 
    \textcolor{black}{\ding{55}} & 
    \textcolor{black}{\ding{55}} \\
    & CLIP-Adapter~\cite{gao2021clip} & \textcolor{black}{\ding{55}} & 
    \textcolor{black}{\ding{55}} & 
    \textcolor{black}{\ding{55}} \\
    \midrule
    \multirow{6}{*}{\begin{minipage}[t]{0.2\textwidth}
          \centering
         \textit{Intermediate}\\
          \textit{methods}
        \end{minipage}} & TIP-Adapter~\cite{zhang2022tip} & \textcolor{black}{\ding{51}} & \textcolor{black}{\ding{55}} & 
    \textcolor{black}{\ding{55}} \\
    & UPL~\cite{huang2022unsupervisedupl} & \textcolor{black}{\ding{55}} & \textcolor{black}{\ding{51}} & \textcolor{black}{\ding{55}} \\
    & SVL-Adapter~\cite{pantazis2022svladapter} & \textcolor{black}{\ding{55}} & 
    \textcolor{black}{\ding{51}} &
    \textcolor{black}{\ding{55}} \\
    & TPT~\cite{shu2022tpt} & 
    \textcolor{black}{\ding{55}} & \textcolor{black}{\ding{51}} & \textcolor{black}{\ding{51}} \\
    & CLIP+SYN~\cite{he2022syntheticclipsyn} & \textcolor{black}{\ding{55}} & \textcolor{black}{\ding{51}} & \textcolor{black}{\ding{51}} \\
    & CaFo~\cite{zhang2023prompt} & \textcolor{black}{\ding{55}} & \textcolor{black}{\ding{51}} & \textcolor{black}{\ding{51}} \\
    \midrule
    \multirow{3}{*}{\begin{minipage}[t]{0.2\textwidth}
        \centering
        \textit{Zero-shot}\\
        \textit{methods}
        \end{minipage}} & \underline{Zero-Shot CLIP}~\cite{radford2021learning} & \textcolor{black}{\ding{51}} & \textcolor{black}{\ding{51}} & \textcolor{black}{\ding{51}} \\    
       & CALIP~\cite{guo2022calip} & \textcolor{black}{\ding{51}} & \textcolor{black}{\ding{51}} & \textcolor{black}{\ding{51}} \\
       & {CLIP+DN}~\cite{zhou2023distribution}$^*$ & \textcolor{black}{\ding{51}} & \textcolor{black}{\ding{51}} & \textcolor{black}{\ding{51}} \\
       \midrule
    \multirow{4}{*}{\begin{minipage}[t]{0.2\textwidth}
        \centering
        \textit{Training-free name-only}\\
        \textit{transfer methods}
        \end{minipage}} & 
    CuPL~\cite{pratt2022doescupl} & \textcolor{black}{\ding{51}} & \textcolor{black}{\ding{51}} &
    \textcolor{black}{\ding{51}} \\
    & VisDesc~\cite{menon2022visual} & \textcolor{black}{\ding{51}} & \textcolor{black}{\ding{51}} &
    \textcolor{black}{\ding{51}} \\
    & CHiLS~\cite{novack2023chils}$^\dagger$ & \textcolor{black}{\ding{51}} & \textcolor{black}{\ding{51}} &
    \textcolor{black}{\ding{51}} \\
    & \textbf{\textit{SuS-X}} (ours) & 
    \textcolor{black}{\ding{51}} & \textcolor{black}{\ding{51}} & \textcolor{black}{\ding{51}} \\
    \bottomrule
  \end{tabular}
  \vspace{-2mm}
  \label{tab:taxonomy}
\end{table*}

One such prominent model, CLIP~\cite{radford2021learning}, is trained on a web-scale corpus of 400M image-text pairs using a contrastive loss that maximises the similarities of paired image-text samples. CLIP pioneered the notion of \textit{zero-shot transfer} in the vision-language setting\footnote{This idea of zero-shot transfer is distinct from the traditional zero-shot classification setup introduced by Lampert et al.~\cite{lampert2009learning} in which the task is to generalise to classes not seen during training.}: classification on unseen datasets. 
For a given classification task, CLIP converts the class labels into classwise textual prompts. An example of such a prompt is ``A photo of a $<$\texttt{CLASS}$>$.'', where $<$\texttt{CLASS}$>$ is replaced by the ground-truth text label for each class. It then computes similarities between the query image and text prompts of all classes. The class whose prompt yields the maximal similarity with the query image is then chosen as the predicted label.

The zero-shot performance of CLIP is however limited by its pre-training distribution~\cite{feuer2022caption, santurkar2022caption, fang2022data,nguyen2022quality}. 
If the downstream dataset distribution diverges too strongly from the distribution of images seen during pretraining, CLIP's zero-shot performance drastically drops~\cite{fang2022data}.
To mitigate this, several lines of work propose to adapt CLIP on diverse downstream tasks---\cref{tab:taxonomy} provides a brief summary of these methods.
Most of them
employ fine-tuning on either labelled or unlabelled subsets of data from the target task.
However, fine-tuning such an over-parameterised model can be unstable and lead to overfitting~\cite{couairon2022embedding, gao2021clip}.
Furthermore, having access to the true distribution of the target task can be prohibitive in data-scarce environments~\cite{christie2018functional, beery2018recognition, kermany2018identifying} and online learning settings~\cite{cossu2022continual, srinivasan2022climb}.

To alleviate these issues, in this paper, we aim to adapt CLIP and other VLMs for downstream classification in a \textit{name-only} (requires only category names\footnote{We use category and class interchangeably in this paper.}, but no samples from the target task) and \textit{training-free} fashion.
We propose \textbf{\textit{SuS-X}} (see~\cref{fig:teaser}), consisting of two novel building blocks: (i) {\textit{SuS}} (\underline{Su}pport \underline{S}ets), our dynamic \textit{support set} curation strategy that forgoes the need for samples from the target task, and (ii) {\textit{TIP-X}}, our main framework for performing zero-shot classification while being training-free.
For a given downstream task, we first curate a \textit{support set} by leveraging the task category labels, either in a parametric manner \ie, generating images from large-scale text-to-image models (\eg, Stable Diffusion~\cite{stablediffusion}) or non-parametric manner \ie, retrieving real-world images from a large vision-language data bank (\eg, LAION-5B~\cite{schuhmann2022laion}). We then use the curated \textit{support set} as a proxy few-shot dataset to inform our downstream predictions using \textit{TIP-X}, in a similar vein to recent few-shot adaptation methods~\cite{gao2021clip, zhang2022tip}. 

Our extensive experiments show that \textit{\textbf{SuS-X}} outperforms zero-shot methods on 19 benchmark datasets across three VLMs, namely, CLIP, BLIP and TCL by 4.60\%, 5.97\% and 11.37\%
absolute average accuracy respectively.
We further extend the \textit{TIP-X} framework to the few-shot regime, outperforming previous SoTA methods in the \textit{training-free} domain. Our main contributions are three-fold: (1) We propose \textit{\textbf{SuS-X}}, a SoTA method in the \textit{training-free name-only transfer} setting for downstream adaptation of VLMs, (2) We present \textit{SuS}, an effective strategy for curating \textit{support sets} using parametric or non-parametric methods to mitigate the lack of data samples available from the target task distribution, and (3) We propose \textit{TIP-X}, a novel training-free method for adapting VLMs to downstream classification in both the \textit{name-only} transfer and few-shot regimes.

\section{Related Work}
\label{sec:related_work}

\noindent \textbf{Vision-Language (VL) Foundation Models.}
In the past few years, there has been a Cambrian explosion in large-scale VL foundation models~\cite{bommasani2021opportunities}. 
In a seminal work, Radford et al.~\cite{radford2021learning} introduced CLIP, a large VLM trained on a massive corpus (400M image-text pairs acquired from the web) that exhibits strong downstream visual task performance.
The introduction of CLIP inspired 
further development of VLMs~\cite{li2022blip, alayrac2022flamingo, jia2021scaling, desai2021virtex, zhang2020contrastive, yu2022coca, yang2022visiontcl, chen2022prototypical, wang2021simvlm, gao2022pyramidclip, cyclip, li2021supervision, ma2022x, you2022learning}, each pre-trained on web-scale datasets to learn joint image-text representations.
These representations can then be applied to tackle downstream tasks like semantic segmentation~\cite{shin2022reco, ghiasi2021open}, object detection~\cite{gu2021open, du2022learning}, image captioning~\cite{mokady2021clipcap, barraco2022unreasonable} and generative modelling~\cite{stablediffusion, ramesh2022hierarchical}, 
In this work, we adapt such VLMs in a training-free setting to diverse downstream tasks.\\
\noindent \textbf{Adaptation of VL models.}
The paradigm shift introduced by CLIP is its ability to do image classification in a zero-shot transfer setting~\cite{radford2021learning}.
In this setup, none of the target dataset classes are known \textit{a-priori} and the task is to adapt implicitly at inference time to a given dataset.
Since CLIP's training objective drives it to assign appropriate similarities to image-text pairs, it acquires the ability to perform zero-shot classification directly.

Inspired by CLIP's zero-shot success, further work has sought to improve upon its performance.
In~\cref{tab:taxonomy}, we characterise some of these methods along three major axes: (i) if the method requires training, (ii) if the method requires labelled samples from the target task, and (iii) if the method requires samples from the target task distribution\footnote{Note that (iii) subsumes (ii). (ii) refers to access to labelled data samples from the target dataset whereas (iii) refers to a more general setting where the samples from the target dataset can be unlabelled. We distinguish between the two for clarity.}.

In this work, we focus on the \textit{training-free name-only transfer} regime---our
goal is to adapt
VLMs to
target tasks without explicit training or access to samples from the target distribution.
Instead, we assume access only to category names of target tasks.
This formulation was recently considered 
for
semantic segmentation, where it was 
called~\textit{name-only transfer}~\cite{shin2022namedmask}---we likewise adopt this terminology.
To the best of our knowledge, only two other concurrent approaches, CuPL~\cite{pratt2022doescupl} and VisDesc~\cite{menon2022visual},
operate in this regime.
They use pre-trained language models to enhance textual prompts for zero-shot classification.
By contrast, \textit{\textbf{SuS-X}} pursues a \textit{support set} curation strategy to adapt
VLMs using knowledge of category names.
These approaches are complementary, and we find that they can be productively combined.
Two other related works operating purely in the zero-shot setting are: (1) CALIP~\cite{guo2022calip}, which uses parameter-free attention on image-text features, and (2) CLIP+DN~\cite{zhou2023distribution}, which uses distribution normalisation. 
We compare
with these four baselines in \cref{sec:results}.

\vspace{-1mm}
\section{\textbf{\textit{SuS-X}}: Training-Free Name-Only Transfer}\label{susx}
\begin{figure*}
    \centering
    \includegraphics[width=\textwidth]{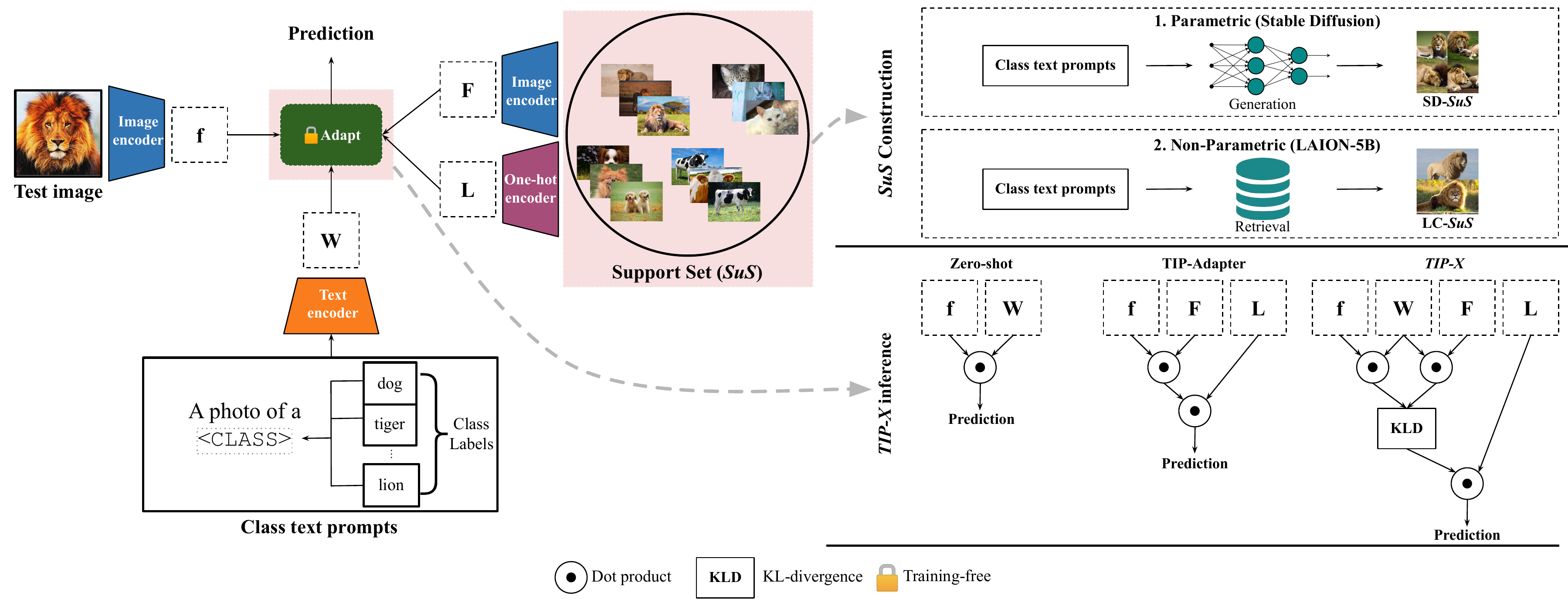}
    \caption{\textbf{\textbf{\textit{SuS-X}} for training-free name-only transfer.}
    {\textit{SuS-X}} consists of two core building blocks.
    (1) \textit{SuS} (top right), a dynamic \textit{support set} that we construct to infuse visual information into the VLM based only on knowledge of target category names.
    We construct support sets either in a parametric (generating images using Stable Diffusion) or non-parametric (retrieving images from LAION-5B) manner. 
    (2) \textit{TIP-X} (bottom right), our novel training-free method that leverages image-text distances to compute similarities between the \textit{support set} and the test images. These similarities act as attention weights for the \textit{support set} labels, and can directly be combined with the original logits from the VLM for classification.
    }
    \vspace{-3mm}
    \label{fig:susx}
\end{figure*}

We describe the two main building blocks of \textbf{\textit{SuS-X}}---(1)~Support Set (\textit{SuS}) construction, and (2)~training-free inference using our novel \textit{TIP-X} method. 
\cref{fig:susx} depicts our overall \textit{training-free name-only} transfer framework.

\subsection{\textbf{\textit{SuS}} Construction} 
\label{method:subsec:sus}

We follow recent adaptation methods~\cite{zhang2022tip, gao2021clip} that use a small collection of labelled images to provide visual information to CLIP.
However, differently from these methods, rather than accessing labelled images from the target distribution, we propose two methods (described next) to construct such a \textit{support set} (\textit{SuS}) without such access. %

\noindent \textbf{(I) Stable Diffusion Generation.} 
Our first method leverages the powerful text-to-image generation model, \emph{Stable Diffusion}~\cite{stablediffusion}.
We employ specific prompting strategies for generating salient and informative support images. %
Concretely, given a set of downstream textual class labels, $\mathcal{T}=\{t_1, t_2, \dots, t_C\}$, where $C$ denotes the number of categories, we prompt Stable Diffusion to generate $N$ images per class. In this way, we construct our \textit{support set} of size $NC$, with each image having its associated class label.

By default, we prompt Stable Diffusion using the original CLIP prompts, \ie, ``A photo of a $<$\texttt{CLASS}$>$.'', where $<$\texttt{CLASS}$>$ is the class text label.
To further diversify the generation process, we follow CuPL~\cite{pratt2022doescupl} to first generate customised textual prompts for each class by prompting GPT-3~\cite{brown2020language} to output descriptions of the particular class. We then feed this customised set of prompts output by GPT-3 into Stable Diffusion for generating images. For example, to generate images from the ``dog'' class, we prompt GPT-3 to describe ``dogs'', and then prompt Stable Diffusion with the resulting descriptions. In section~\ref{analysis}, we compare the performance of the default (called \textit{Photo}) and this augmented prompting procedure (called \textit{CuPL}). Unless otherwise specified, all our experiments with Stable Diffusion \textit{support sets} use the \textit{CuPL} strategy.

\noindent \textbf{(II) LAION-5B Retrieval.} 
Our second method leverages the large-scale vision-language dataset,
\textit{LAION-5B}~\cite{schuhmann2022laion}. 
It contains 5.85 billion image-text pairs,
pre-filtered by
CLIP.
Using LAION-5B, we retrieve task-specific
images
using class text prompts for constructing the \textit{support set}. 
Concretely, given textual class labels, $\mathcal{T}=\{t_1, t_2, \dots, t_C\}$, we rank all images in LAION-5B by their CLIP image-text similarity to each text class label $t_{i}$, where $i\in[1, C]$. 
We then use the top $N$ image matches as our \textit{support set} for class $i$,
resulting in an $NC$-sized \textit{support set} of images with their associated class labels. 
Note that curating supporting knowledge by search is a classical technique in computer vision~\cite{fergus2005learning} that was recently revisited in the task of semantic segmentation~\cite{shin2022reco}.
Here we adapt this idea to the \textit{name-only transfer} classification setting.
For efficient retrieval, we leverage the approximate nearest neighbour indices released by the authors\footnote{\href{https://huggingface.co/datasets/laion/laion5B-index}{https://huggingface.co/datasets/laion/laion5B-index}}.
Similar to the Stable Diffusion generation approach, we experiment with both \textit{Photo} and \textit{CuPL} prompting strategies for curating our LAION-5B \textit{support set} (see~\cref{analysis}). By default, we use \textit{Photo} prompting for all our experiments with LAION-5B \textit{support sets}.

\noindent \textbf{Remark.} %
Note that \textit{SuS} can be seen as a visual analogue to CuPL~\cite{pratt2022doescupl}, where, for each class, we augment VLMs with rich, relevant images, instead of the customised textual descriptions generated in CuPL.

\subsection{\textbf{\textit{TIP-X}} Inference}
\label{method:subsec:tip-x}
Given our \textit{support set} from the previous section, our task is to now leverage it in a training-free inference scheme to inform CLIP's zero-shot predictions. 
We first briefly review the zero-shot CLIP classification pipeline, discuss the recently proposed TIP-Adapter~\cite{zhang2022tip} for training-free adaptation, and highlight a critical shortcoming in its method due to uncalibrated intra-modal embedding distances, which we address in our method---\textit{TIP-X}.

\noindent\textbf{Zero-shot CLIP.}
For classification into $C$ classes, CLIP converts class labels into text prompts and encodes them with its text encoder.
Collectively, the encoded prompt vectors can be interpreted as a classifier weight matrix $W \in \mathbb{R}^{C{\times}d}$, where $d$ is
embedding dimension.
For a
test set ${T}{=}{\{y_1, y_2, ..., y_t\}}$ comprising $t$ test images, CLIP's image encoder is applied to produce test image features:
\begin{equation}
\begin{gathered}
f_i = \texttt{CLIPImageEncoder}(y_i), i\in[1, t], f_i\in\mathbb{R}^{d}\\
f = \texttt{Concat}([f_1, f_2, \dots, f_t]), f\in\mathbb{R}^{t\times d}
\end{gathered}
\end{equation}
Using $W$ and $f$, CLIP performs classification by computing zero-shot logits ($\texttt{ZSL}$) via a dot product:
\begin{equation}
\texttt{ZSL} = fW^T
\end{equation}

\noindent\textbf{TIP-Adapter.}
Given a $CK$-sized $K$-shot labelled dataset $D = \{x_1, x_2, \dots, x_{CK}\}$\footnote{Note that a $K$-shot labelled dataset for $C$ classes has a size $CK$.} from the target domain, TIP-Adapter~\cite{zhang2022tip} encodes $D$ using CLIP's image encoder:
\begin{equation}
\begin{gathered}
F_i = \texttt{CLIPImageEncoder}(x_i), i\in[1, CK], F_i\in\mathbb{R}^d\\
F = \texttt{Concat}([F_1, F_2, \dots, F_{CK}]), F\in\mathbb{R}^{CK\times d}
\end{gathered}
\end{equation}
It then converts each of the few-shot class labels to one-hot vectors $L\in \mathbb{R}^{CK{\times}C}$. Next, it computes an affinity matrix to capture the similarities between $F$ and $f$: 
\begin{equation}
\label{tip-affinity}
A = \exp(-\beta(1-fF^T))
\end{equation} where $\beta$ is a hyperparameter that modulates ``\textit{sharpness}''.
Finally, these affinities are used as attention weights over $L$ to produce logits that are blended with $\texttt{ZSL}$ using a hyperparameter, $\alpha$:
\begin{equation}
\texttt{TL} = \alpha AL + fW^T
\end{equation}

\noindent
\textbf{Motivating \textit{TIP-X}.} TIP-Adapter gains from the affinity computation between the test and few-shot image samples (see~\cref{tip-affinity}). This similarity is computed in CLIP's image space. However, prior research~\cite{yu2022multimodal, liang2022mind, udandarao2022understanding} has demonstrated the existence of a \textit{modality gap} between CLIP's image and text spaces. This leads us to question if doing image-image similarity comparisons in CLIP's image space is optimal.

\begin{figure}[h]
  \centering
  \begin{subfigure}[b]{\linewidth}
    \centering
    \includegraphics[width=0.96\linewidth]{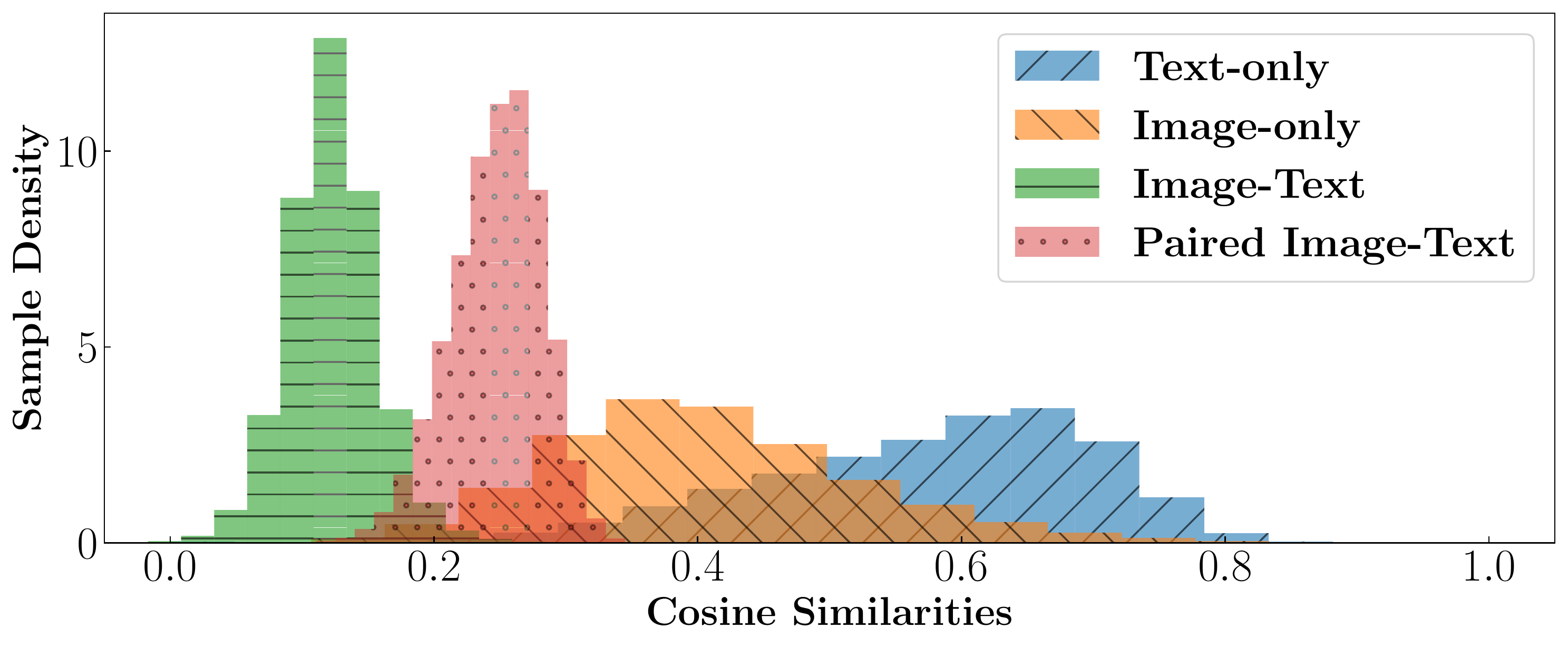}
    \caption{\textbf{Intra-modal and inter-modal CLIP cosine similarities.}
    We observe quite distinct intra-modal and inter-modal cosine similarity distributions.}
    \label{fig: cosine-sims}
  \end{subfigure}\\
  \begin{subfigure}[b]{\linewidth}
    \centering
    \includegraphics[width=\linewidth]{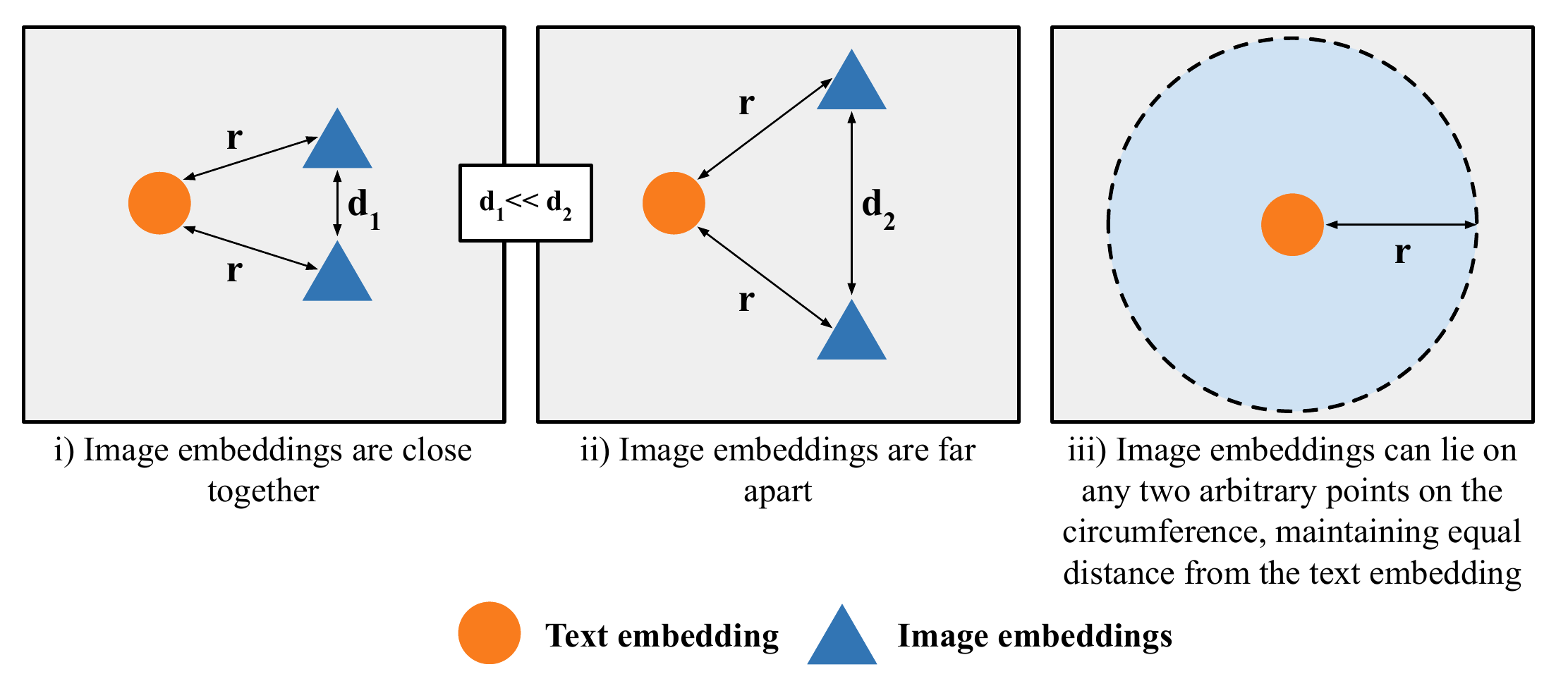}
    \caption{\textbf{Intra-modal degrees of freedom}.
    Different intra-modal similarities can satisfy same inter-modal constraints, leaving room for poor calibration.}
    \label{fig:motivation-tipx}
  \end{subfigure}
  \caption{\textbf{Our two-fold analysis motivating \textit{TIP-X}}}
  \label{fig:motivating-tipx}
\end{figure}

~\cref{fig: cosine-sims} shows the pairwise image-image, text-text and image-text cosine similarities of the ImageNet validation set CLIP embeddings.
Clearly, the intra-modal and inter-modal similarities are distributed differently---the inter-modal similarities have small variance and mean, whereas the intra-modal similarities  have larger means and variances.
This mismatch happens because \textit{contrastive training of CLIP maximises the inter-modal cosine similarities of paired samples without regard to intra-modal similarities}.
This implies that the intra-image CLIP embedding similarities employed by TIP-Adapter may not reflect the true intra-image similarities.
\cref{fig:motivation-tipx} illustrates this idea with a simple example. Consider
two image embeddings that are required to be a distance $r$ away from a particular text embedding. The two image embeddings can satisfy this condition by being very close to each other or very far apart from each other. 
\cref{fig:motivation-tipx} shows that this constraint can be satisfied by any two arbitrary points on a hypersphere of radius $r$. 
While we expect loose constraints to be imposed via transitivity, we nevertheless expect a lower quality of calibration in intra-modal (\eg, image-image) comparisons.

\noindent \textbf{\textit{TIP-X} to the rescue.}
To get around the problem of uncalibrated intra-modal embedding distances in TIP-Adapter, we propose to use inter-modal distances as a bridge.
More specifically, rather than computing similarities between the test features ($f{\in}\mathbb{R}^{t{\times}d}$) and few-shot features ($F{\in}\mathbb{R}^{CK\times d}$) in the image embedding space ($fF^T$), we use the image-text space.
We first construct signatures by computing similarities of $f$ and $F$ with the text classifier weights $W$: 
\begin{equation}
\begin{gathered}
S = \texttt{softmax}(FW^T), S\in\mathbb{R}^{CK\times C}\\
s = \texttt{softmax}(fW^T), s\in\mathbb{R}^{t\times C}
\end{gathered}
\end{equation}
These signatures comprise probability distributions encoding inter-modal affinities between the few-shot features and class text vectors, and likewise for the test features.
We then construct our affinity matrix $M \in \mathbb{R}^{t\times CK}$ by measuring the KL-divergence between the signatures as follows:

\begin{equation}
\begin{gathered}
    M_{i, j} = \texttt{KL}(s_{i}||S_{j}), i\in[1, t] , j\in[1, CK]
\end{gathered}
\end{equation}
where $s_i$ represents the $i^{th}$ test signature for the $t$ test samples, and $S_j$ represents the $j^{th}$ few-shot signature. Since we are working with discrete probability distributions, we compute the KL-divergence as $\texttt{KL}(P||Q) = \sum_{i} P_{i} \log\frac{P_{i}}{Q_{i}}$.

The construction of the affinity matrix $M$ can be seen as analogous to the affinity computation in TIP-Adapter (\cref{tip-affinity}).
However, our affinity matrix construction removes direct reliance on the uncalibrated image-image similarities.

Finally, before using our affinity matrix $M$ as attention weights for $L$ (one-hot encoded class labels), we rescale (denoted by $\psi$) the values of $M$ to have the same range (min, max values) as the TIP-Adapter affinities ($A$).
Further, since our affinity matrix $M$ consists of KL-divergence values, the most similar samples will get small weights since their KL-divergence will be low (close to 0).
To mitigate this, we simply negate the values in $M$.
We then blend our predicted logits with $\texttt{TL}$ using a scalar $\gamma$:
\begin{equation}
\label{txl-logits}
\texttt{TXL} = fW^{T} + \alpha AL + \gamma\psi(-M)L 
\end{equation}
The entire \textit{TIP-X} method is shown in 
~\cref{fig:susx} (bottom right).

\subsection{\textbf{\textit{SuS-X}}: Combining \textbf{\textit{SuS}} and \textbf{\textit{TIP-X}}}
Since our constructed \textit{support sets}
act
as pseudo few-shot datasets, we directly replace the few-shot features $F$ in the \textit{TIP-X} framework with the features of our
\textit{support set}. We call our 
method \textbf{\textit{SuS-X-LC}} if we combine \textit{TIP-X} with the LAION-5B curated \textit{support set}, and \textbf{\textit{SuS-X-SD}} when combined with the Stable Diffusion generated \textit{support set}. These methods enable \textit{training-free name-only} adaptation of zero-shot VLMs.

\section{Experiments}
\label{sec:results}

First, we evaluate \textbf{\textit{SuS-X}} against strong baselines in the \textit{training-free zero-shot/name-only} transfer regimes, across three VLMs.
Next, we illustrate the adaptation of \textit{TIP-X} into the few-shot training-free regime.
Finally, we ablate and analyse our method to provide additional insights.

\subsection{Training-free name-only transfer evaluation}\label{training-free-section}

\noindent
\textbf{Datasets.} 
For a comprehensive evaluation, we test on 19
datasets spanning a wide range of object, scene and fine-grained categories: ImageNet~\cite{deng2009imagenet}, StanfordCars~\cite{krause20133d}, UCF101~\cite{soomro2012ucf101}, Caltech101~\cite{fei2004learning}, 
Caltech256~\cite{griffin2007caltech},
Flowers102~\cite{nilsback2008automated}, OxfordPets~\cite{parkhi2012cats}, Food101~\cite{bossard2014food}, SUN397~\cite{xiao2010sun}, DTD~\cite{cimpoi2014describing}, EuroSAT~\cite{helber2019eurosat}, FGVCAircraft~\cite{maji2013fine},
Country211~\cite{radford2021learning},
CIFAR-10~\cite{cifar},
CIFAR-100~\cite{cifar},
Birdsnap~\cite{berg2014birdsnap},
CUB~\cite{wah2011caltech}, ImageNet-Sketch~\cite{wang2019learning} and ImageNet-R~\cite{hendrycks2021many}. 
Previous few-shot adaptation methods~\cite{zhang2021tip, gao2021clip, zhou2021learning} benchmark on a subset of 11 of these 19 datasets. 
We report results on the
19-dataset suite in the main paper and compare results using only the 11-dataset subset in the supp. mat.

\noindent
\textbf{Experimental Settings.}\label{experimental-settings-clip} 
We compare against six baselines.
For zero-shot CLIP, we use prompt ensembling with 7 different prompt templates following~\cite{radford2021learning,zhang2022tip}\footnote{The 7 prompt templates are: ``itap of a $<$\texttt{class}$>$.'', ``a origami $<$\texttt{class}$>$.'', ``a bad photo of the $<$\texttt{class}$>$.'', ``a photo of the large $<$\texttt{class}$>$.'', ``a $<$\texttt{class}$>$ in a video game.'', ``art of the $<$\texttt{class}$>$.'', and ``a photo of the small $<$\texttt{class}$>$.''.}. 
We run CuPL\footnote{\href{https://github.com/sarahpratt/CuPL}{https://github.com/sarahpratt/CuPL}}, VisDesc\footnote{\href{https://github.com/sachit-menon/classify_by_description_release}{https://github.com/sachit-menon/classify\_by\_description\_release}} (\textit{name-only} transfer) and CLIP+DN\footnote{\href{https://github.com/fengyuli2002/distribution-normalization}{https://github.com/fengyuli2002/distribution-normalization}} (\textit{zero-shot} transfer) using their official code.
We also experiment with augmenting the CuPL prompts with the original prompt ensemble, and call it CuPL+e.
For CALIP (\textit{zero-shot} transfer), in the absence of public code at the time of writing, we aim to reproduce their results using our own implementation. 
For our proposed methods, we report results using both \textbf{\textit{SuS-X-LC}} and \textbf{\textit{SuS-X-SD}}.
For both methods, we use a fixed number of support samples per dataset (see supp. mat. for details). 
For CALIP and \textbf{\textit{SuS-X}}, we conduct a hyperparameter search on the dataset validation sets.
In~\cref{analysis} we perform a hyperparameter sensitivity test for a fair evaluation.
By default, we use the ResNet-50~\cite{he2016deep} backbone as CLIP's image encoder for all models.

\begin{table*}
\footnotesize
  \centering
    \caption{\textbf{Training-free adaptation of CLIP on 19 datasets with RN50 visual backbone}.
    The best and second best results for each dataset are \textbf{bolded} and \underline{underlined}, respectively. 
    Individual results for all 19 datasets are available in the supp. mat. $^*$Average reported across 19 datasets. $^\dagger$Our re-implementation.
    }
    \vspace{-3mm}
  \begin{tabular}{c|c|c|cccccc}
    \toprule
     &
    \textbf{Method} & \textbf{Average}$^*$ & \textbf{ImageNet~\cite{deng2009imagenet}} &\textbf{ImageNet-R~\cite{hendrycks2021many}} & \textbf{ImageNet-Sketch~\cite{wang2019learning}} & \textbf{EuroSAT~\cite{helber2019eurosat}} & \textbf{DTD~\cite{cimpoi2014describing}} & \textbf{Birdsnap~\cite{berg2014birdsnap}} \\
    \midrule
    \multirow{4}{*}{\textit{Zero-shot}} &
    Zero-shot CLIP~\cite{radford2021learning} & 52.27 & 60.31 & 59.34 & 35.42 & 26.83 & 41.01 & 30.56 \\
    & CALIP~\cite{guo2022calip} & -- & 60.57 & -- & -- & 38.90 & 42.39 & -- \\    
    & CALIP~\cite{guo2022calip}$^\dagger$ & 52.37 & 60.31 & 59.33 & {36.10} & 26.96 & 41.02 & 30.68 \\ 
    & CLIP+DN~\cite{zhou2023distribution} & 53.02 & 60.16 & 60.37 & 35.95 & 28.31 & 41.21 & 31.23 \\ 
    \midrule
    \multirow{5}{*}{\textit{Name-only}} & CuPL~\cite{pratt2022doescupl} & 55.50 & 61.45 & 61.02 & 35.13 & 38.38 & 48.64 & 35.65 \\ 
    & CuPL+e & 55.76 & 61.64 & 61.17 & 35.85 & 37.06 & 47.46 & 35.80 \\ 
    & VisDesc~\cite{menon2022visual} & 53.76 & 59.68 & 57.16 & 33.78 & 37.60 & 41.96 & 35.65  \\
    &\textit{ SuS-X-SD} (ours) & \underline{56.73} & \underline{61.84} & \underline{61.76} & \underline{36.30} & \textbf{45.57} & \textbf{50.59} & \underline{37.14} \\    
    & \textit{SuS-X-LC} (ours) & \textbf{56.87} & \textbf{61.89} & \textbf{62.10} & \textbf{37.83} & \underline{44.23} & \underline{49.23} & \textbf{38.50} \\
\bottomrule
  \end{tabular}
  \label{tab:main-result-table-with-baselines}
  \vspace{-2mm}
\end{table*}

\begin{figure*}[h]
  \centering
  \begin{subfigure}[b]{0.3\linewidth}
    \centering
    \hspace*{-2mm}\includegraphics[height=1.6in]{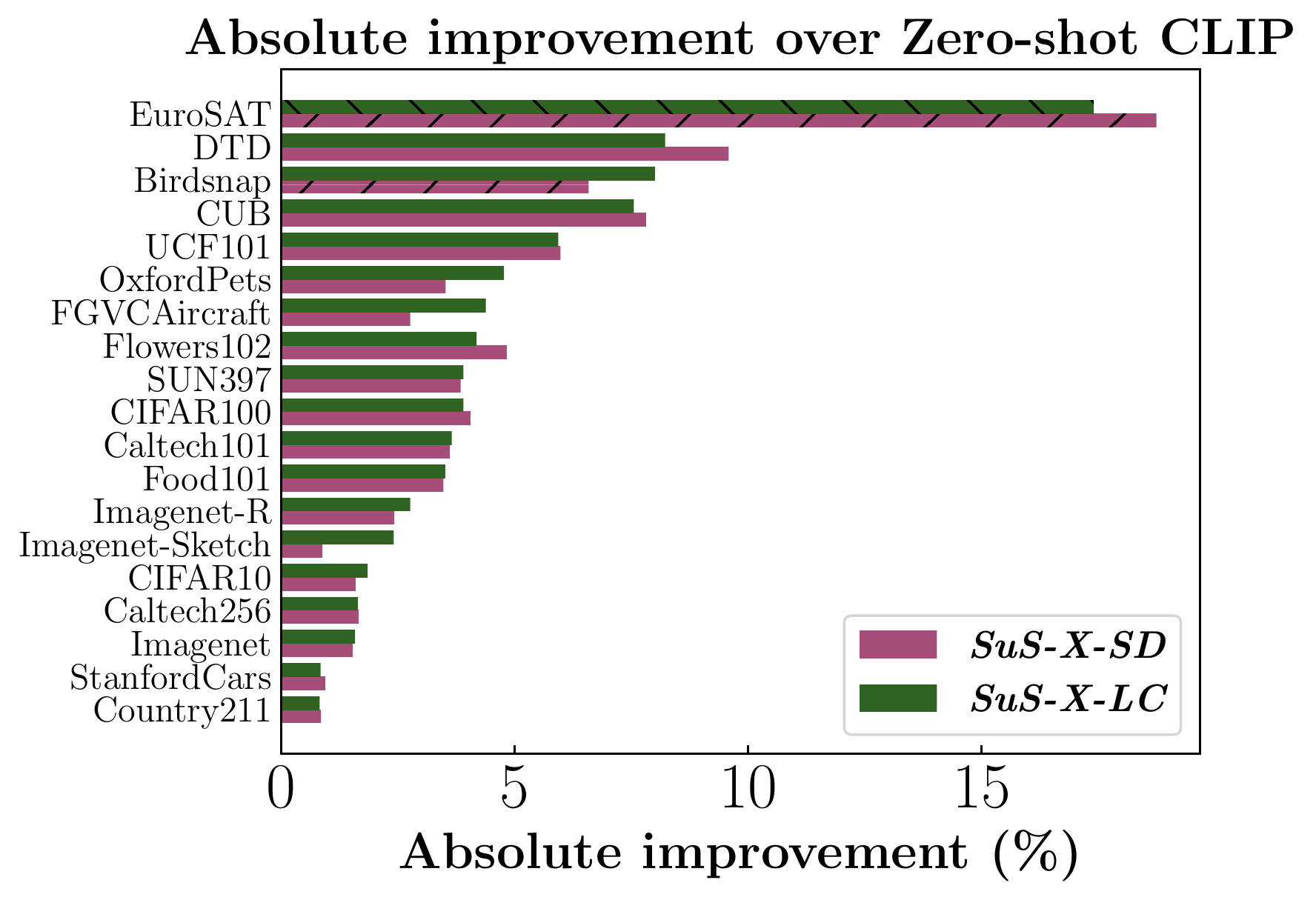}\vspace{-2mm}
    \caption{}
    \label{fig:sus-x-zs-comparison}
  \end{subfigure}
  \quad
  \begin{subfigure}[b]{0.3\linewidth}
    \centering
    \includegraphics[height=1.6in]{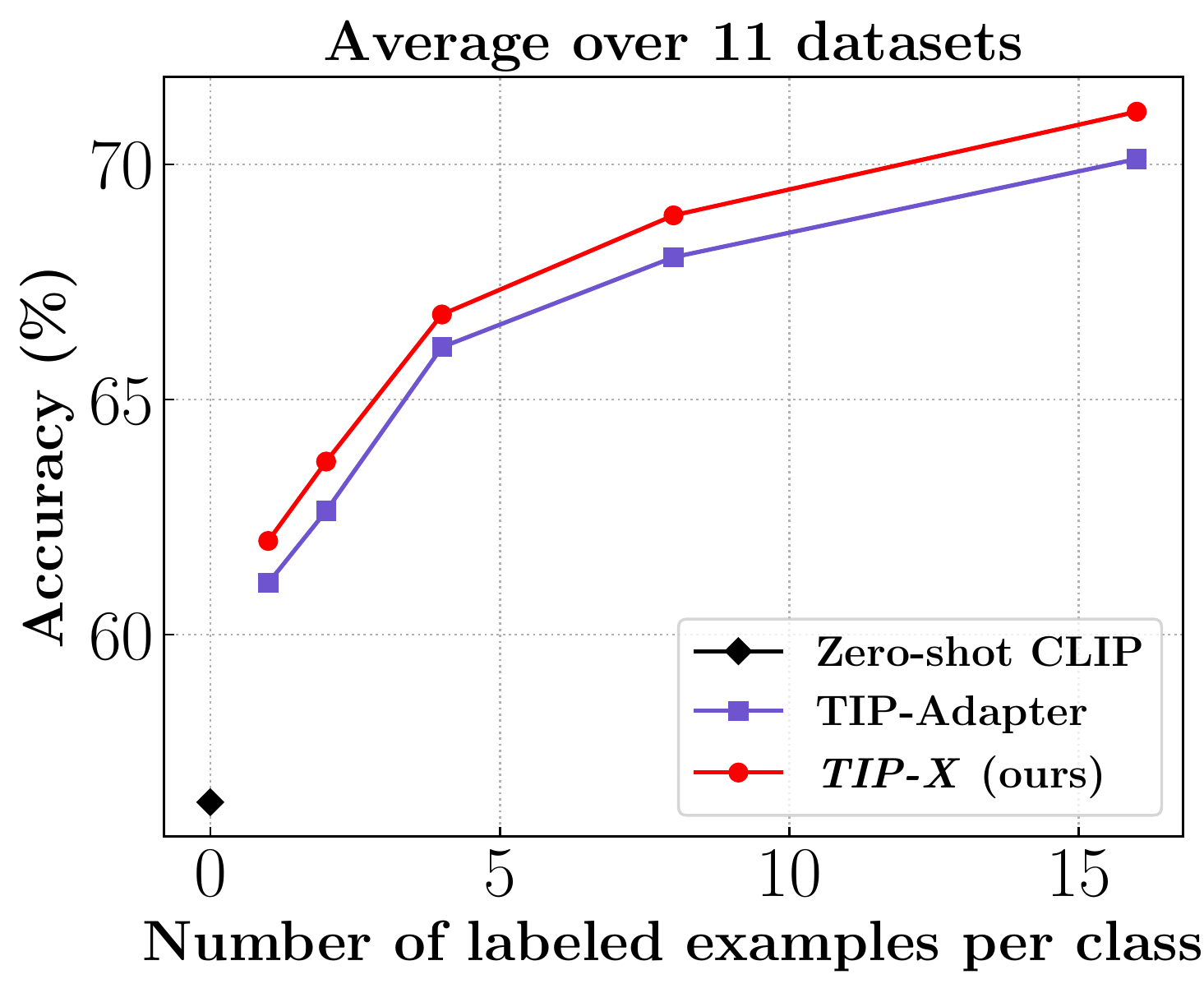}\vspace{-2mm}
    \caption{}
    \label{fig:tip-x-few-shot}
  \end{subfigure}
  \quad
  \begin{subfigure}[b]{0.3\linewidth}
    \centering
    \includegraphics[height=1.6in]{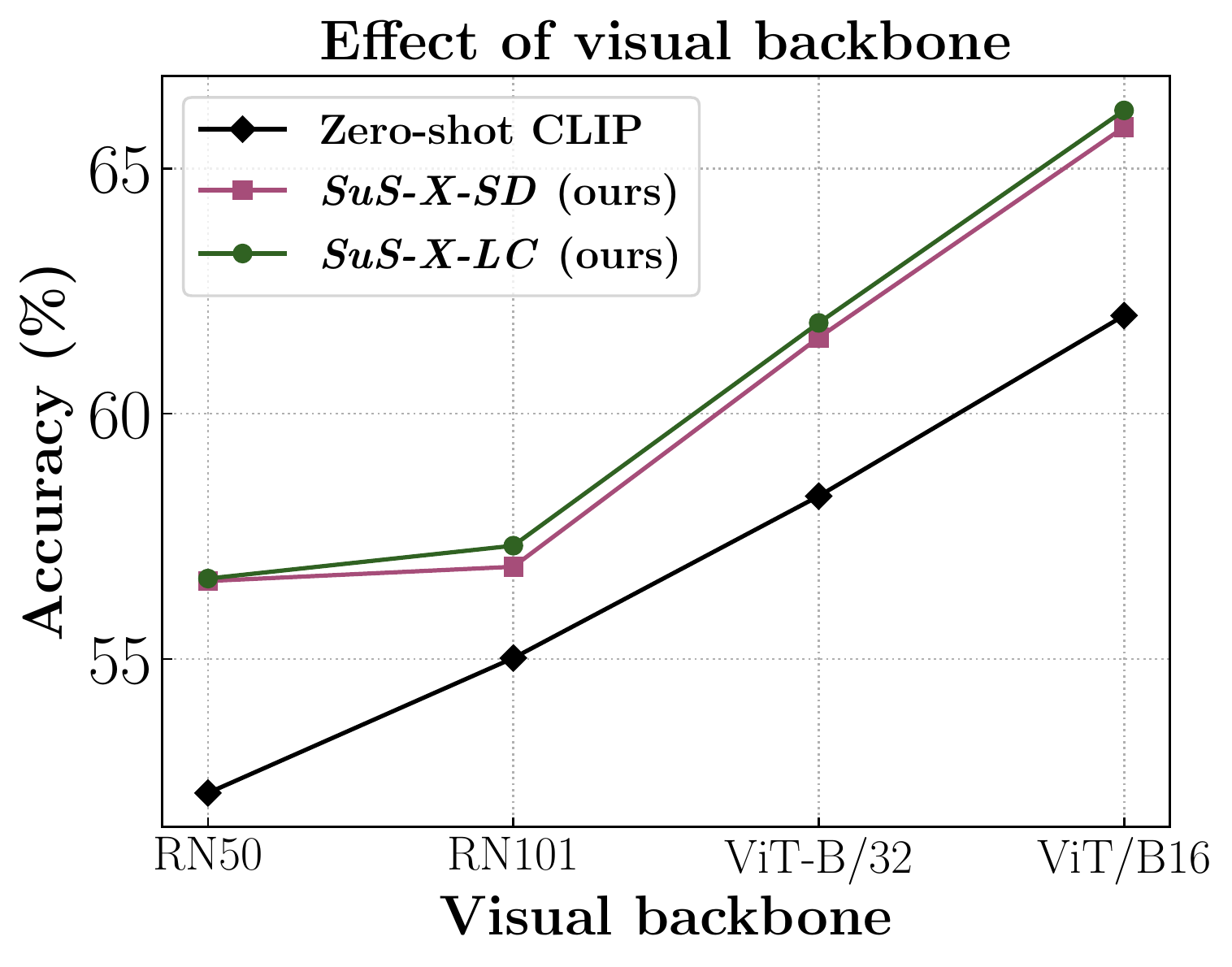}\vspace{-2mm}
    \caption{}
    \label{fig:visual-backbone-ablation}
  \end{subfigure}
  \vspace{-2mm}
  \caption{\textbf{(a)} Comparison of \textbf{\textit{SuS-X}} with Zero-shot CLIP. \textbf{(b)} Results of training-free few-shot classification. \textbf{(c)} Performance comparison of \textbf{\textit{SuS-X}} across visual backbones.}
  \label{fig:tip-x-few-shot-zs-clip}
\end{figure*}

\noindent \textbf{Main Results.}\label{main-results} 
In~\cref{tab:main-result-table-with-baselines}, we compare both variants of \textbf{\textit{SuS-X}} with the 
baselines.
We report an average across 19 datasets.
We also include results on ImageNet, EuroSAT, DTD, Birdsnap, ImageNet-R and ImageNet-Sketch (results on all 19 datasets in the supp. mat.).
\textbf{\textit{SuS-X}} methods outperform zero-shot CLIP by 4.6\% on average across all 19 datasets. 
We observe striking gains of 18\%, 8\% and 7\% on EuroSAT, DTD and Birdsnap respectively.
We also outperform the SoTA training-free adaptation methods---CuPL+ensemble and VisDesc by 1.1\% and 3.1\% on average respectively.
To further probe where we attain the most gains, we plot the absolute improvement of our models over zero-shot CLIP in~\cref{fig:sus-x-zs-comparison}.
We observe large gains on fine-grained (Birdsnap, CUB, UCF101) and specialised (EuroSAT, DTD) datasets, demonstrating the utility of \textbf{\textit{SuS-X}} in injecting rich visual knowledge into zero-shot CLIP (additional fine-grained classification analysis in supp. mat.). We further compare \textbf{\textit{SuS-X}} to few-shot methods that use labelled samples from the true distribution in the supp. mat.---despite being at a disadvantage due to using no target distribution samples, \textbf{\textit{SuS-X}} is still competitive with these methods.

\subsection{Transfer to different VLMs}
\label{transfer-vlms}
We evaluate transfer to VLMs other than CLIP, namely TCL~\cite{yang2022visiontcl} and BLIP~\cite{li2022blip}. We only retain image and text encoders of these models for computing features, while preserving all other experimental settings from~\cref{experimental-settings-clip}. \cref{tab:other-architecture-results} shows our \textbf{\textit{SuS-X}} methods strongly outperform all baseline methods across both VLMs---we improve on zero-shot models by 11.37\% and 5.97\% on average across 19 datasets. This demonstrates that our method is not specific to CLIP, but can improve performance across different VLMs.

\setlength\tabcolsep{0.5pt}
\begin{table}
\footnotesize
\centering
\caption{\textbf{\textit{SuS-X} generalises to different VLMs.} $^*$Average reported across 19 datasets.}
\vspace{-3mm}
\setlength\tabcolsep{1.4pt}
\begin{tabular}{c|c|c|cccc}
\toprule
\textbf{VLM} & \textbf{Method} & \textbf{Average$^*$} & \textbf{ImageNet} & \textbf{EuroSAT} & \textbf{DTD} & \textbf{Birdsnap}\\
\midrule
\multirow{6}{*}{\textit{\textbf{TCL}}} & Zero-shot  & 31.38 & 35.55 & 20.80 & 28.55 & 4.51 \\
& CuPL & 34.79 & 41.60 & 26.30 & 42.84 & 6.83 \\
& CuPL+e & 32.79 & 41.36 & 25.88 & 41.96 & 6.60 \\
& VisDesc & 33.94 & 40.40 & 21.27 & 34.28 & 5.69 \\
& \textit{SuS-X-SD} & \underline{41.49} & \underline{52.29} & \underline{28.75} & \textbf{48.17} & \underline{13.60} \\
& \textit{SuS-X-LC} & \textbf{42.75} & \textbf{52.77} & \textbf{36.90} & \underline{46.63} & \textbf{17.93} \\
\midrule
\multirow{6}{*}{\textit{\textbf{BLIP}}} & Zero-shot & 48.73 & 50.59 & 44.10 & 44.68 & 10.21 \\
& CuPL & 51.11 & 52.96 & 39.37 & 52.95 & 12.24 \\
& CuPL+e & 51.36 & 53.07 & 41.48 & 53.30 & 12.18 \\
& VisDesc & 49.91 & 50.94 & 42.25 & 47.45 & 11.69 \\
& \textit{SuS-X-SD} & \underline{53.20} & \underline{55.93} & \underline{45.36} & \textbf{56.15} & \underline{16.95} \\
& \textit{SuS-X-LC} & \textbf{54.64} & \textbf{56.75} & \textbf{51.62} & \underline{55.91} & \textbf{23.78}\\
\bottomrule
\end{tabular}
\vspace{-3mm}
\label{tab:other-architecture-results}
\end{table}

\subsection{Adapting to the few-shot regime}
\label{adapt-tipx-few-shot}
A key component of our \textbf{\textit{SuS-X}} method is \textit{TIP-X}.
In the previous section, we showcased SoTA results in the training-free name-only transfer regime. 
Due to its formulation, \textit{TIP-X} can directly be extended to the few-shot regime, where our \textit{support sets} are labelled samples from the target dataset rather than curated/generated samples.
To evaluate \textit{TIP-X} on such real-world \textit{support sets}, we conduct training-free few-shot classification using \textit{TIP-X}.
We compare against the SoTA method in this regime---TIP-Adapter~\cite{zhang2022tip}.
We report results on the 11-dataset subset used by TIP-Adapter on five different shot settings of the $K$-shot classification task: 1, 2, 4, 8 and 16.

We present average accuracy results on all shots in~\cref{fig:tip-x-few-shot}---\textit{TIP-X} outperforms both Zero-shot CLIP and TIP-Adapter (absolute gain of 0.91\% across shots). Notably, on OxfordPets, we achieve 2.1\% average gain. This further demonstrates the generalisability of the \textit{TIP-X} method in transferring to the few-shot training-free setting.

\subsection{Analysis}
\label{analysis}

We conduct several ablations and provide additional visualisations to offer further insight into the \textit{\textbf{SuS-X}} method.

\noindent \textbf{Component Analysis.} \textit{\textbf{SuS-X}} consists of two major building blocks---\textit{SuS} construction and \textit{TIP-X}. We compare the performance difference (with average accuracy across 19 datasets) of using \textit{SuS} with TIP-Adapter instead of \textit{TIP-X} in~\cref{tab:component_analysis}. We use both default ensemble prompts and CuPL prompts for CLIP's text classifier to break down the performance gains further.
We note that both \textit{SuS} and \textit{TIP-X} are crucial for achieving the best results.

\begin{figure*}
  \centering
  \subfloat[Dishwasher]{%
  \begin{minipage}{0.48\linewidth}
  \centering
  \includegraphics[width=0.3\textwidth]{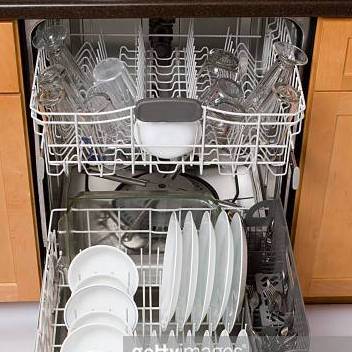}
  \includegraphics[width=0.3\textwidth]{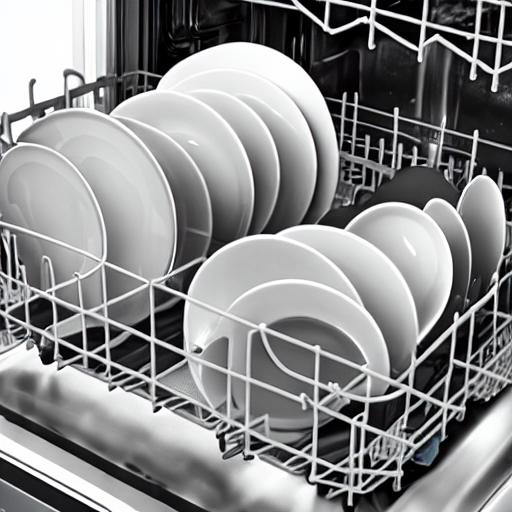}
  \includegraphics[width=0.3\textwidth]{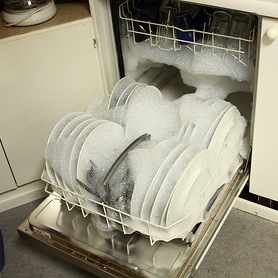}
  \end{minipage}%
  }
  \centering
  \subfloat[Split Rail Fence]{%
  \begin{minipage}{0.48\linewidth}
  \centering
  \includegraphics[width=0.3\textwidth]{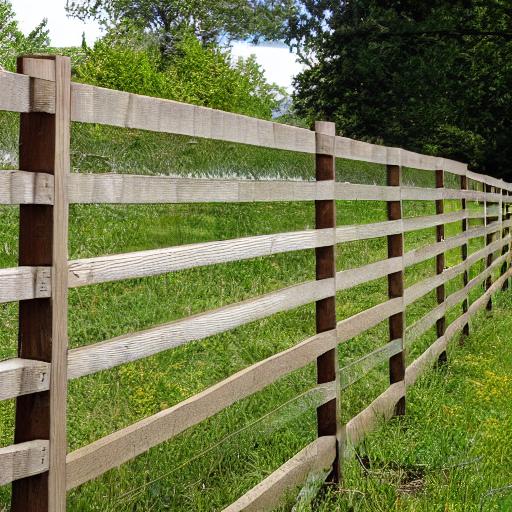}
  \includegraphics[width=0.3\textwidth]{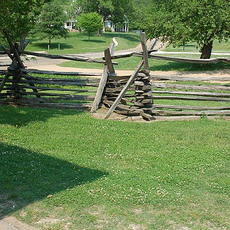}
  \includegraphics[width=0.3\textwidth]{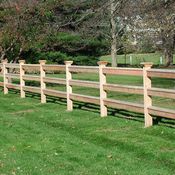}
  \end{minipage}%
  }
  \par
  \centering
  \subfloat[Australian Kelpie]{%
  \begin{minipage}{0.48\linewidth}
  \centering
  \includegraphics[width=0.3\textwidth]{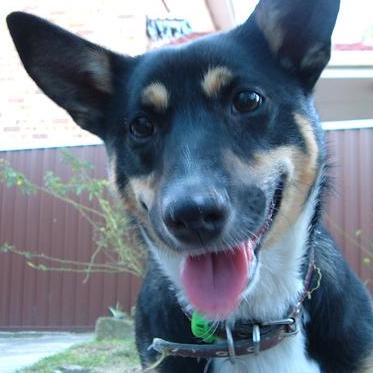}
  \includegraphics[width=0.3\textwidth]{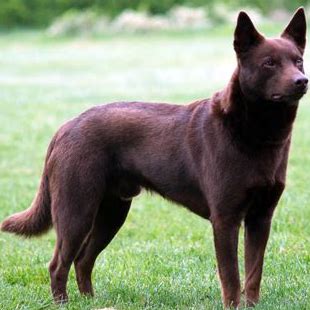}
  \includegraphics[width=0.3\textwidth]{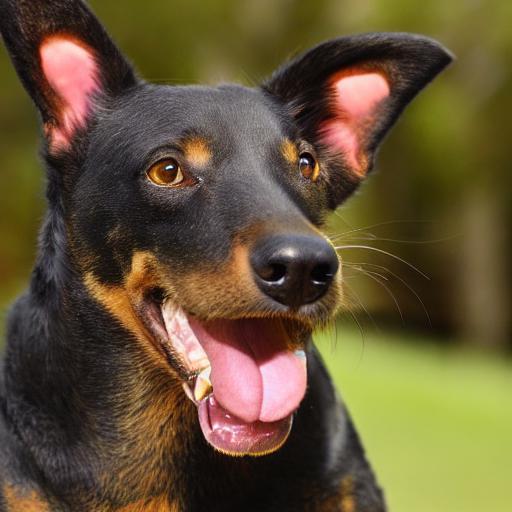}
  \end{minipage}%
  }
  \centering
  \subfloat[Bulbul]{%
  \begin{minipage}{0.48\linewidth}
  \centering
  \includegraphics[width=0.3\textwidth]{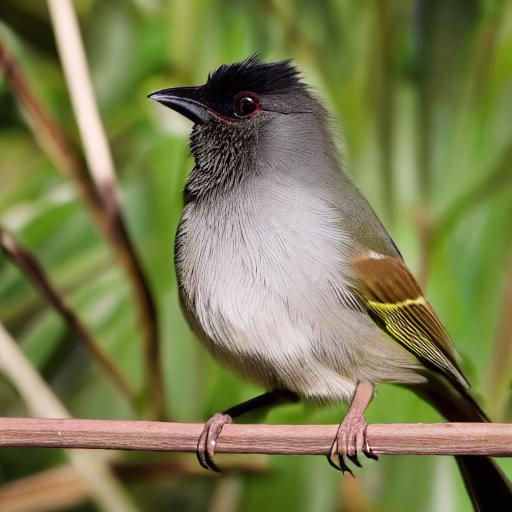}
  \includegraphics[width=0.3\textwidth]{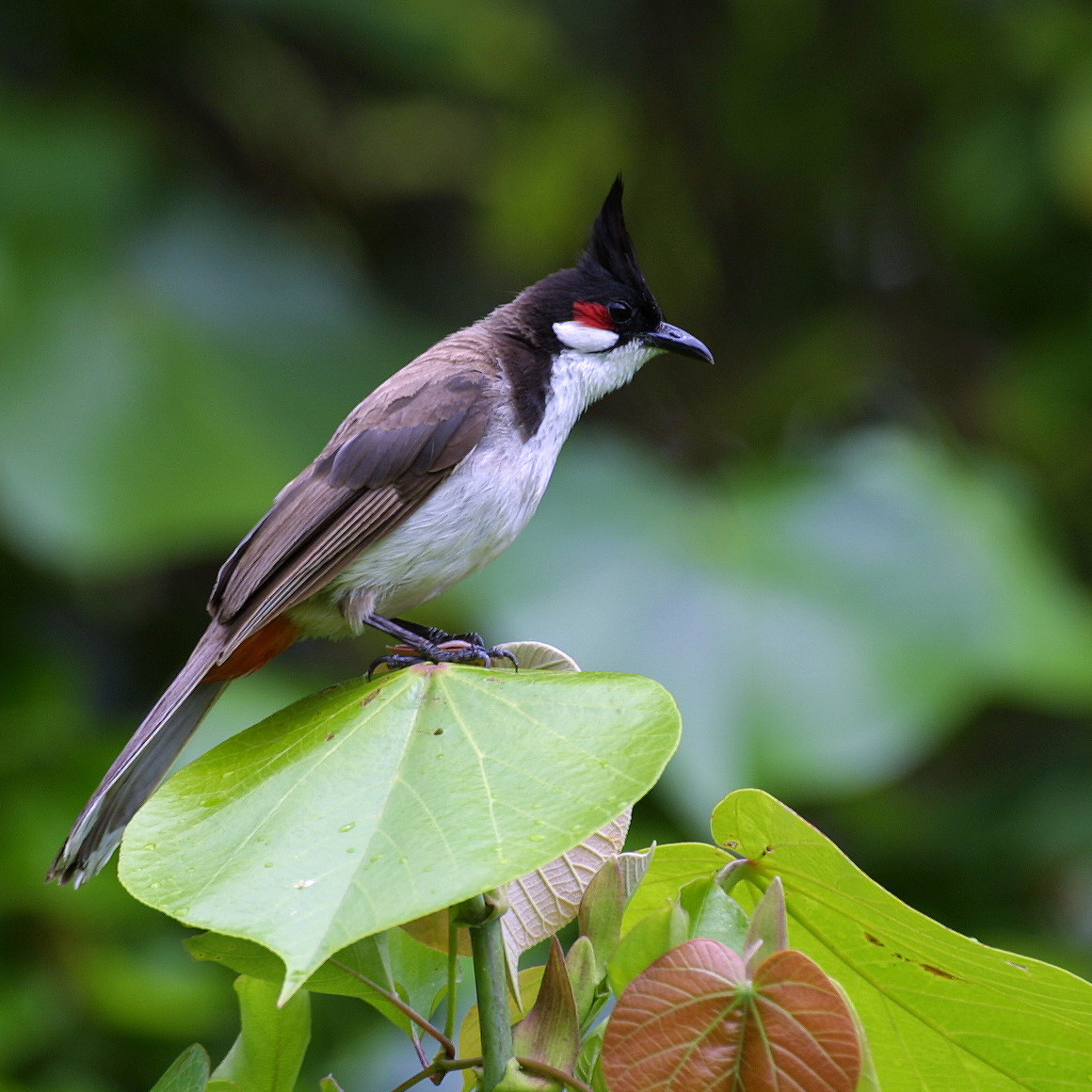}
    \includegraphics[width=0.3\textwidth]{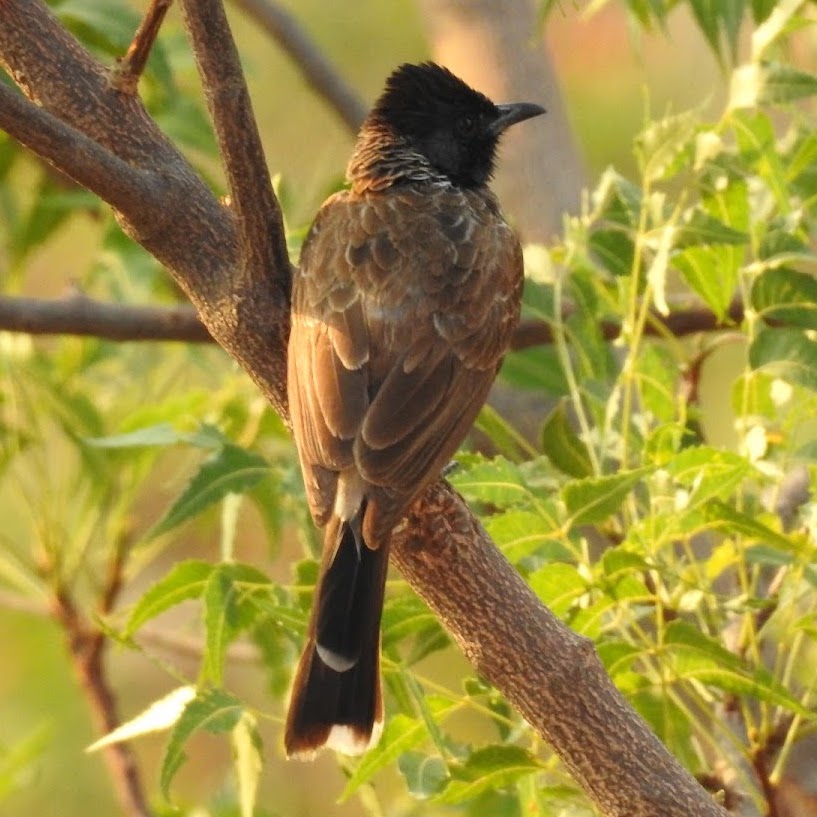}
  \end{minipage}%
  }
  \vspace{-2mm}
  \caption{\textbf{Support samples from the generated \textit{SuS-SD}, retrieved \textit{SuS-LC} and true training distribution for ImageNet.} 
  By randomising the image order in each subfigure, we pose a challenge question---can you match the three images for each subfigure to their source \textit{i.e.} \textit{SuS-SD}, \textit{SuS-LC} or ImageNet train set?
  The answers are provided at the bottom of the page\protect\footnotemark.
  }
  \vspace{-3.5mm}
  \label{fig: sus-examples}
\end{figure*}

\noindent \textbf{Transfer to different visual backbones.} We evaluate the scalability of our model across different CLIP visual backbones---~\cref{fig:visual-backbone-ablation} shows that both \textbf{\textit{SuS-X}} variants consistently improve upon zero-shot CLIP across ResNet and VisionTransformer backbones of varying depths and sizes.

\begin{figure}    
  \subfloat[Tasks where larger support sets are beneficial]{%
  \begin{minipage}{\linewidth}
  \centering
  \includegraphics[width=0.5\textwidth]{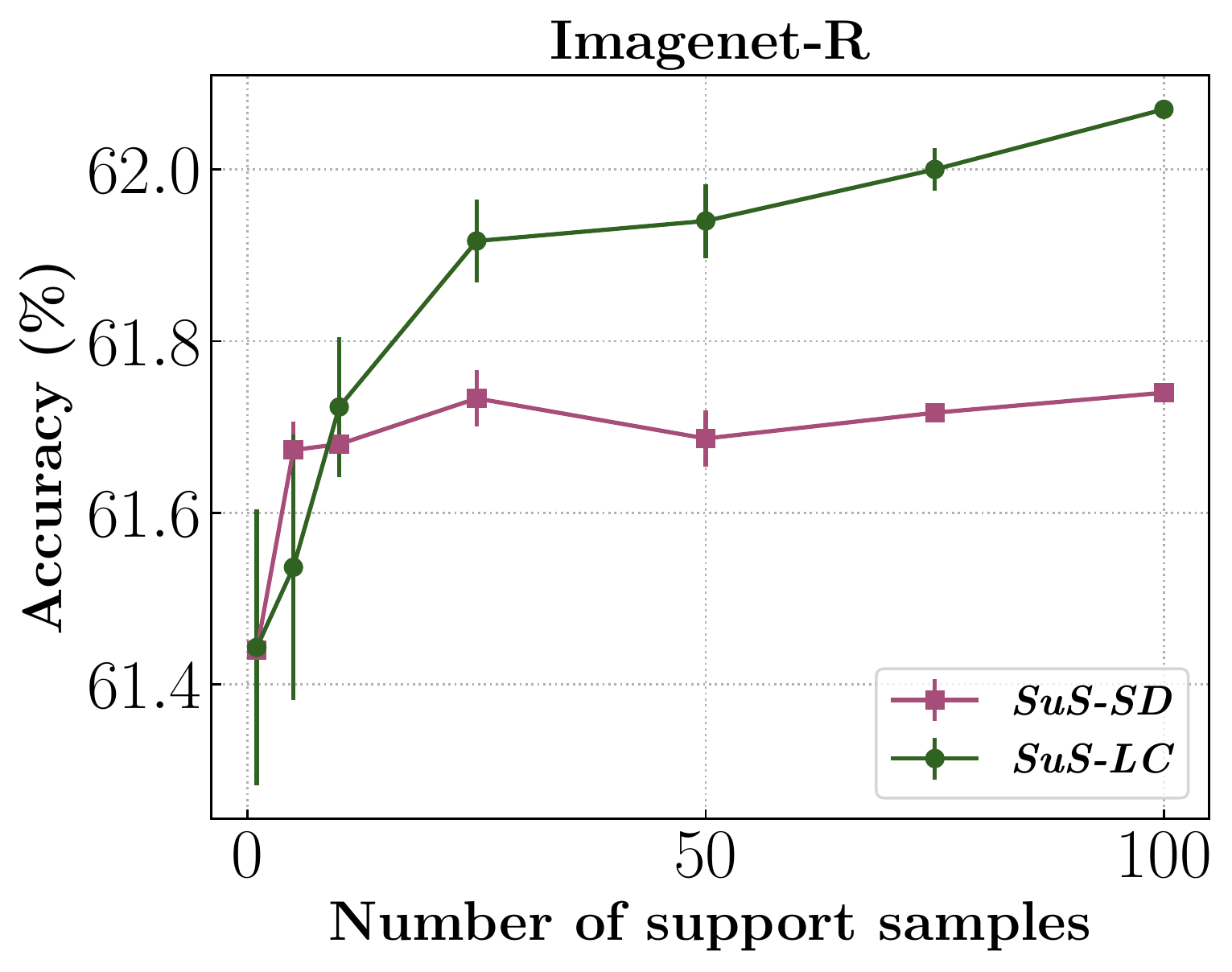}
  \hspace*{-0.2cm}
  \includegraphics[width=0.5\textwidth]{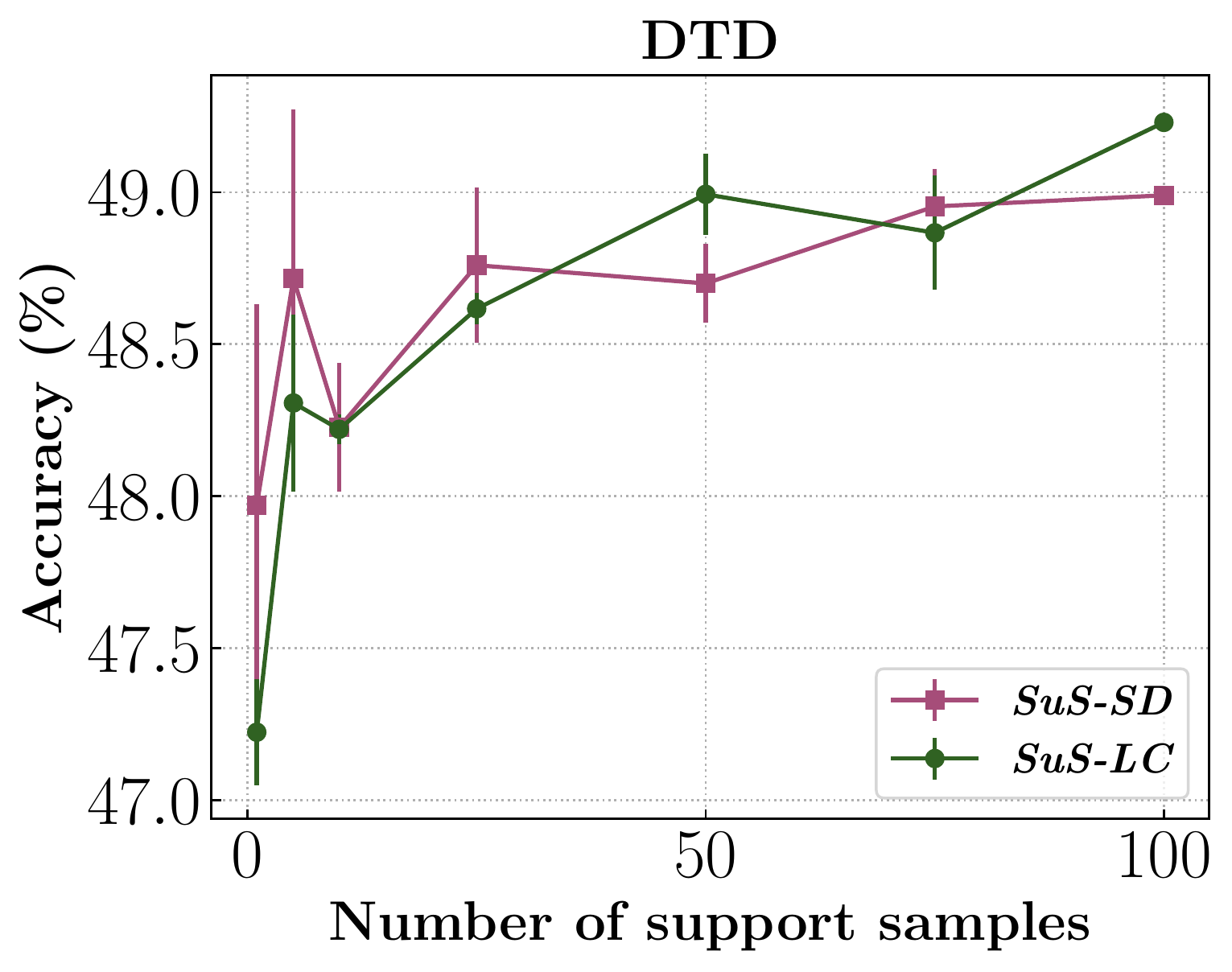}
  \end{minipage}%
  }\\
  \subfloat[Tasks where larger support sets are harmful]{%
  \begin{minipage}{\linewidth}
  \centering
  \includegraphics[scale=0.27]{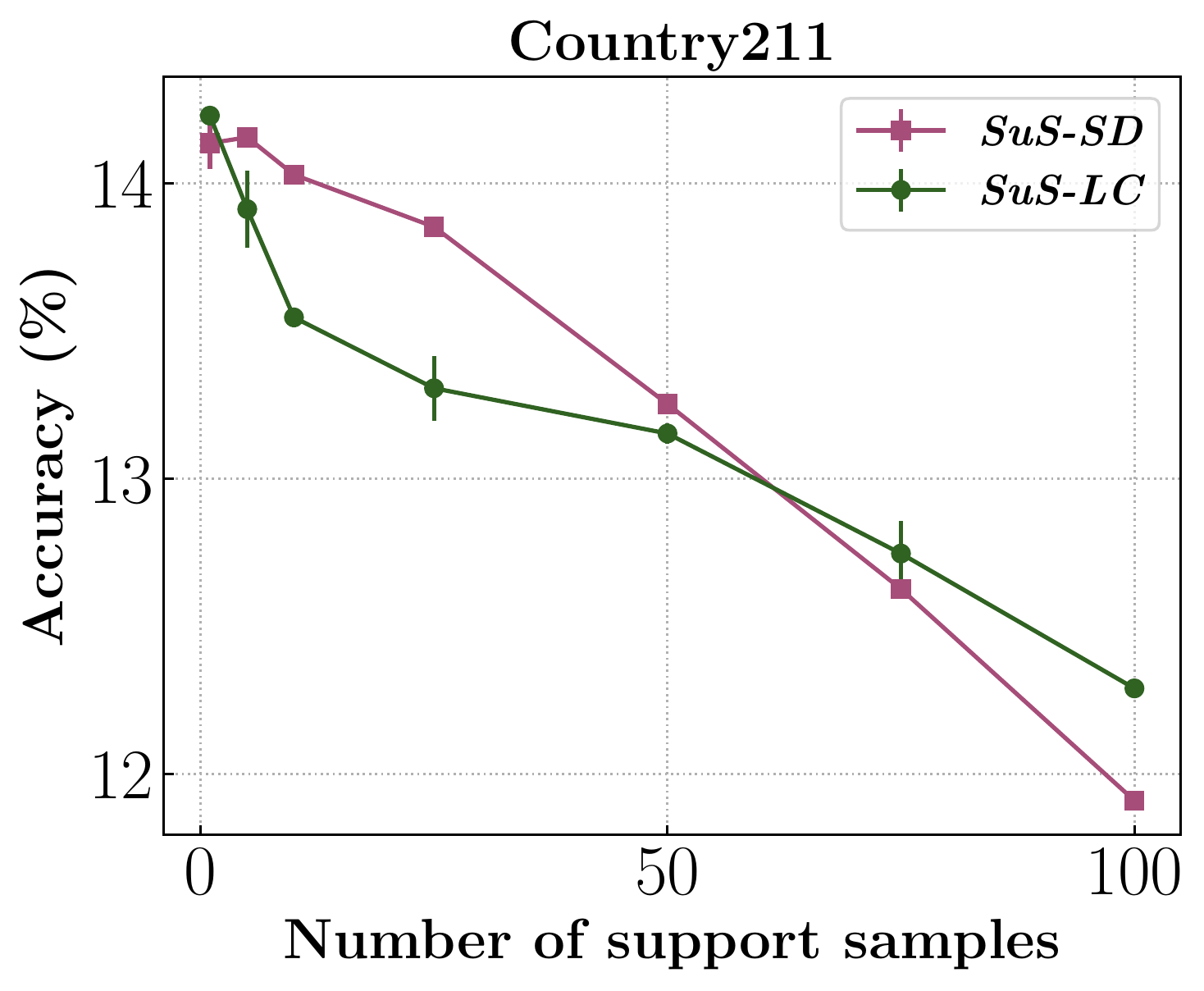}
  \hspace*{-0.2cm}
  \includegraphics[width=0.5\textwidth]{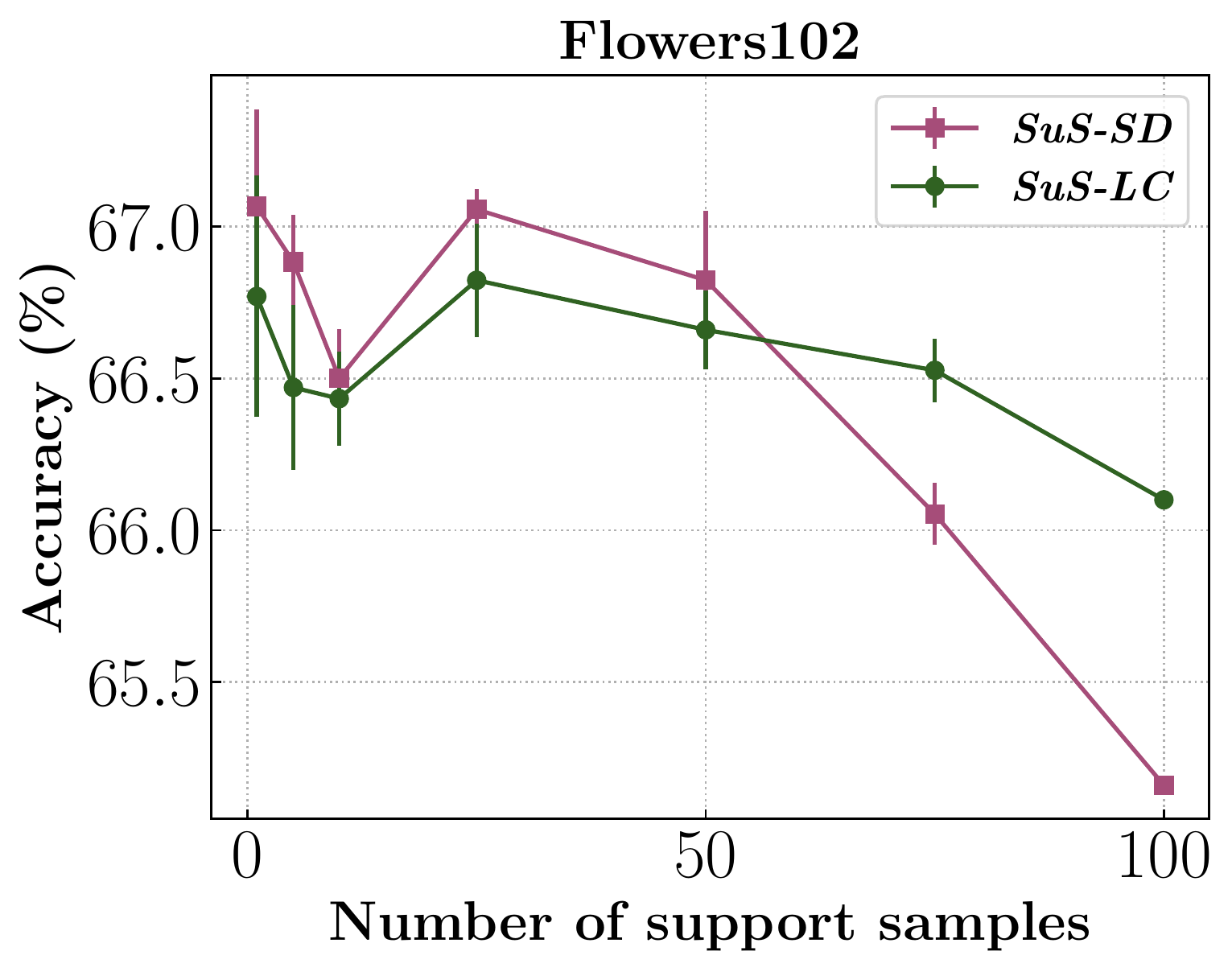}
  
  \end{minipage}%
  }
  \vspace{-1mm}
  \caption{\textbf{Effect of support size.}}
  \vspace{-3mm}
  \label{fig: sus-size-effects}
\end{figure}

\noindent \textbf{\textit{SuS} size.}\label{support-size-ablation} 
We study the effect of varying \textit{support set} size for \textit{SuS-LC} and \textit{SuS-SD}---we generate three different \textit{support sets} with random seeds for support sizes of 1, 5, 10, 25, 50, 75 and 100 samples.
From~\cref{fig: sus-size-effects}, we observe two broad trends---some tasks benefit (ImageNet-R, DTD) from having more \textit{support set} samples while others do not (Country211, Flowers102). 
We suggest that this is connected to the domain gap between the true data distribution and \textit{support set} samples---if the domain gap is large, it is inimical to provide
a large \textit{support set}, whereas if the domains are similar, providing more support samples always helps. 

\noindent \textbf{\textit{SuS} visualisation.} We visualise samples from both \textit{support set} construction methods on ImageNet in~\cref{fig: sus-examples}. 
It is hard to distinguish between the true ImageNet samples and the \textit{SuS} samples---we can therefore construct \textit{support sets} to mimic the true data distribution, with access to only the category names. 
A caveat is that the \textit{support set} does not always capture the domain characteristics of the true distribution, leading to a domain gap (lighting conditions, diverse scene backgrounds, confounding objects etc). 
To fully close the gap to using true few-shot datasets as \textit{support sets}~\cite{gao2021clip, zhang2022tip}, further research into exact unsupervised domain matching of \textit{support sets} and few-shot datasets is required. 

\noindent \textbf{Prompting strategies for \textit{SuS} construction.}\label{prompting-strategies} 
\cref{tab:photo-vs-cupl-strategies} depicts the performance of \textit{Photo} and \textit{CuPL} prompting---best results are achieved with the \textit{LC-Photo} and \textit{SD-CuPL} strategies. We further compare the diversity of images produced by the two strategies on ImageNet\footnote{We compute diversity as 1 minus the mean of the average pairwise image cosine-similarities within a class. A larger value implies low cosine similarities across images within a class, implying more diverse images. Alternatively, a smaller value implies less diverse images.}---from~\cref{tab:photo-vs-cupl-strategies}, it is evident that \textit{CuPL} prompting leads to more diverse support sets as compared to \textit{Photo} prompting.

\setlength\tabcolsep{1.5pt}
\begin{table}
    \footnotesize
    \centering
    \caption{\textbf{Component Analysis of \textbf{\textit{SuS-X}}.} 
    }
    \vspace{-3mm}
    \begin{tabular}{c|c|cc|c}
    \toprule
     \textbf{Text Prompts} & \textbf{Method} & \textbf{\textit{SuS}} & \textbf{\textit{TIP-X}} & \textbf{Average Accuracy}\\
    \midrule
    \multirow{5}{*}{\textit{Default}} & Zero-shot CLIP & \ding{55} & \ding{55} & 52.27 \\
 & SuS-TIP-SD & \ding{51} & \ding{55} & {53.49} \textcolor{ForestGreen}{(+1.22\%)} \\
& \textit{SuS-X-SD} & \ding{51} & \ding{51} & {53.69} \textcolor{ForestGreen}{(+1.42\%)} \\
& SuS-TIP-LC & \ding{51} & \ding{55} & {53.83} 
 \textcolor{ForestGreen}{(+1.56\%)} \\
& \textit{SuS-X-LC} & \ding{51} & \ding{51} & {54.20} \textcolor{ForestGreen}{ (+1.93\%)} \\
\midrule
\multirow{5}{*}{\textit{CuPL+e}} & CuPL+e & \ding{55} & \ding{55} & {55.76} \textcolor{ForestGreen}{ (+3.49\%)} \\
 & SuS-TIP-SD & \ding{51} & \ding{55} & {56.63} \textcolor{ForestGreen}{(+4.36\%)} \\
& \textit{SuS-X-SD} & \ding{51} & \ding{51} & \underline{56.73} \textcolor{ForestGreen}{(+4.46\%)} \\
& SuS-TIP-LC & \ding{51} & \ding{55} & {56.72} 
 \textcolor{ForestGreen}{(+4.45\%)} \\
& \textit{SuS-X-LC} & \ding{51} & \ding{51} & \textbf{56.87} \textcolor{ForestGreen}{ (+4.60\%)} \\
\bottomrule
    \end{tabular}
    \label{tab:component_analysis}
    \vspace{-2mm}
\end{table}

\setlength\tabcolsep{1.5pt}
\begin{table}
    \footnotesize
    \centering
    \caption{\textbf{Prompting strategies for \textit{SuS} construction.}}
    \vspace{-3mm}
    \begin{tabular}{c|cccc|cc}
    \toprule
    \multirow{2}{*}{
    \begin{minipage}[t]{0.07\textwidth}
          \centering
         \textit{\textbf{SuS}}\\
          \textbf{method}
        \end{minipage}
    } & \multicolumn{2}{c}{\textbf{Average Acc.}} & \multicolumn{2}{c|}{\textbf{ImageNet Acc.}} & \multicolumn{2}{c}{\textbf{Diversity}}
    \\
    \cmidrule{2-7}
    &
    {\textit{Photo}} & {\textit{CuPL}} &
    {\textit{Photo}} & {\textit{CuPL}} &
    {\textit{Photo}} & {\textit{CuPL}}
    \\
    \midrule
\textit{LC} & \textbf{56.87} & 56.20 & \textbf{61.89} & 61.79 & 0.28 & \textbf{0.32} \\
\textit{SD} & 56.32 &  \underline{56.73} & 61.79 & \underline{61.84} & 0.17 & \textbf{0.20}\\
\bottomrule
\end{tabular}
\vspace{-3mm}
\label{tab:photo-vs-cupl-strategies}
\end{table}

\noindent \textbf{Hyperparameter Sensitivity.} We perform a sensitivity test for our $\gamma$ hyperparameter (refer Eq.~\ref{txl-logits}) on ImageNet-R, OxfordPets, and DTD. We fix $\alpha$ and $\beta$ to be 1, and run a sweep over $\gamma\in[0, 1]$. From~\cref{tab:hparam-sensitivity}, we observe that moderate values of $\gamma$ are typically preferred, and the variance of the accuracy values is small. However, note that for DTD, the optimal $\gamma$ is slightly larger (0.75)---this is due to its specialised nature which requires more guidance from the specialised \textit{support set} to inform pre-trained CLIP. Previous few-shot adaptation works~\cite{gao2021clip, zhang2022tip} observed similar results. For more hyperparameter ablations, see the supp. mat.

\setlength\tabcolsep{3.2pt}
\begin{table}
    \footnotesize
    \centering
    \caption{\textbf{Hyperparameter sensitivity for} $\boldsymbol{\gamma}$}
    \vspace{-3mm}
    \begin{tabular}{c|ccccccc}
    \toprule
    \multirow{2}{*}{\textbf{Dataset}} & \multicolumn{7}{c}{$\boldsymbol{\gamma}$ \textbf{value}} \\
    \cmidrule{2-8}
    & $\boldsymbol{0}$ & $\boldsymbol{0.1}$ & $\boldsymbol{0.2}$ & $\boldsymbol{0.3}$ & $\boldsymbol{0.5}$ & $\boldsymbol{0.75}$ & $\boldsymbol{1}$ \\ 
    \midrule
    ImageNet-R
    & 60.87 & 60.98 & 61.03 & \textbf{61.05} & 61.00 & 60.89 & 60.65 \\
    OxfordPets
    & 76.76 & 77.17 & \textbf{77.58} & 77.44 & 77.17 & 77.17 & 76.90 \\
    DTD
    & 47.16 & 47.16 & 47.51 & 47.69 & 47.87 & \textbf{47.96} & 47.60 \\
    \bottomrule
    \end{tabular}
    \vspace{-3mm}
    \label{tab:hparam-sensitivity}
\end{table}

\footnotetext{
\rotatebox{180}{
 (a)LC,SD,Train,(b)SD,Train,LC,(c)Train,LC,SD,(d)SD,Train,LC}    
}

\subsection{Limitations and broader impact}

While demonstrating promising results, we note several limitations of our approach.
(1) To perform \textit{name-only} transfer, we rely on CLIP to have seen related concepts during pre-training. For concepts that are so rare that they do not appear during pre-training, transfer will not be feasible.
(2) We employ LAION-5B~\cite{schuhmann2022laion} as a source of knowledge. 
While reasonable for a proof of concept, this data is relatively uncurated and may contain harmful content. 
As such, our approach is not suitable for real-world deployment without careful mitigation strategies to address this concern.
Similar arguments apply to Stable Diffusion~\cite{stablediffusion}.

\section{Conclusion}
\label{conclusion}

In this paper, we studied the training-free name-only transfer paradigm for classification tasks. 
We systematically curated \textit{support sets} with no access to samples from the target distribution and showed that they help improve CLIP's zero-shot predictions by providing rich, task-specific knowledge. 
We further motivated the \textit{TIP-X} framework through the observation that CLIP's intra-modal embedding spaces are not optimal for computing similarities.
With these two building blocks, we demonstrated superior performance to prior state-of-the-art.

\noindent \textbf{Acknowledgements.}
This work was supported by the Isaac Newton Trust and an EPSRC access-to-HPC grant.
SA would like to acknowledge the support of Z. Novak and N. Novak in enabling his contribution. VU would like to thank Gyungin Shin, Surabhi S. Nath, Jonathan Roberts, Vlad Bogolin, Kaiqu Liang and Anchit Jain for helpful discussions and feedback.

{\small
\bibliographystyle{ieee_fullname}
\bibliography{references}
}

\newpage
\appendix
\onecolumn

\section{Dataset Details}

We enumerate the validation and testing split sizes of all datasets in~\cref{tab:datasets}. We make two small modifications to the standard datasets as described in CoOp~\cite{zhou2021learning}: (1) We discard the ``BACKGROUND Google'' and ``Faces easy classes'' from the Caltech101 dataset, and (2) For the UCF101 dataset, we consider the middle frame of each video as our image sample.

\begin{table}[ht]
    \centering
    \caption{\textbf{Dataset details for the 19 datasets used in this study.}}
    \label{tab:datasets}
    \begin{tabular}{ccccc}
    \toprule
     \textbf{Dataset} & \textbf{Classes} &
     \textbf{Val} &
     \textbf{Test}\\
     \midrule
     UCF-101 & 101 & 1898 & 3783 \\
     CIFAR-10 & 10 & 10000 & 10000 \\
     CIFAR-100 & 100 & 10000 & 10000 \\
     Caltech101 & 100 & 1649 & 2465 \\
     Caltech256 & 257 & 6027 & 9076 \\
     ImageNet & 1000 & 50000 & 50000 \\
     SUN397 & 397 & 3970 & 19850 \\
     FGVCAircraft & 100 & 3333 & 3333 \\
     Birdsnap & 500 & 7774 & 11747 \\
     StanfordCars & 196 & 1635 & 8041 \\
     CUB & 200 & 1194 & 5794 \\
     Flowers102 & 102 & 1633 & 2463 \\
     Food101 & 101 & 20200 & 30300 \\
     OxfordPets & 37 & 736 & 3669 \\
     DTD & 47 & 1128 & 1692 \\
     EuroSAT & 10 & 5400 & 8100 \\
     ImageNet-Sketch & 1000 & 50889 & 50889 \\
     ImageNet-R & 200 & 30000 & 30000 \\
     Country211 & 211 & 10550 & 21100 \\
     \bottomrule
    \end{tabular}
    \end{table}

\section{Details about Support Set Curation Strategies}
We include further technical details about our two support set curation strategies---Stable Diffusion Generation and LAION-5B Retrieval.

\noindent\textbf{Stable Diffusion Generation.} For all our experiments with the Stable Diffusion model, we use the \href{https://huggingface.co/CompVis/stable-diffusion-v1-4}{stable-diffusion-v1-4} checkpoint with a 9.5 guidance scale~\cite{ho2022classifier}, 85 diffusion steps and $512{\times}512$ output %
resolution.
We then downscale these images to CLIP's input resolution of $224{\times}224$.

\noindent\textbf{LAION-5B Retrieval.} For all our experiments, we rank all images in the LAION-5B corpus based on their image-text similarity with the given class textual prompt. We use the LAION-5B \href{https://huggingface.co/datasets/laion/laion5B-index}{pre-constructed index} that leverages the \href{https://openaipublic.azureedge.net/clip/models/b8cca3fd41ae0c99ba7e8951adf17d267cdb84cd88be6f7c2e0eca1737a03836/ViT-L-14.pt}{CLIP-ViT-L/14 model}. Finally, since the images might be of varying resolutions, we pre-process them to CLIP's input resolution of $224{\times}224$.

\section{Few-shot Learning with \textit{TIP-X}}

\begin{figure*}[h]
    \centering
    \subfloat[Average]{\includegraphics[width=0.3\textwidth]{images/cvpr-tip-x-few-shot-result-average.pdf}}\hfill
    \subfloat[ImageNet]{\includegraphics[width=0.3\textwidth]{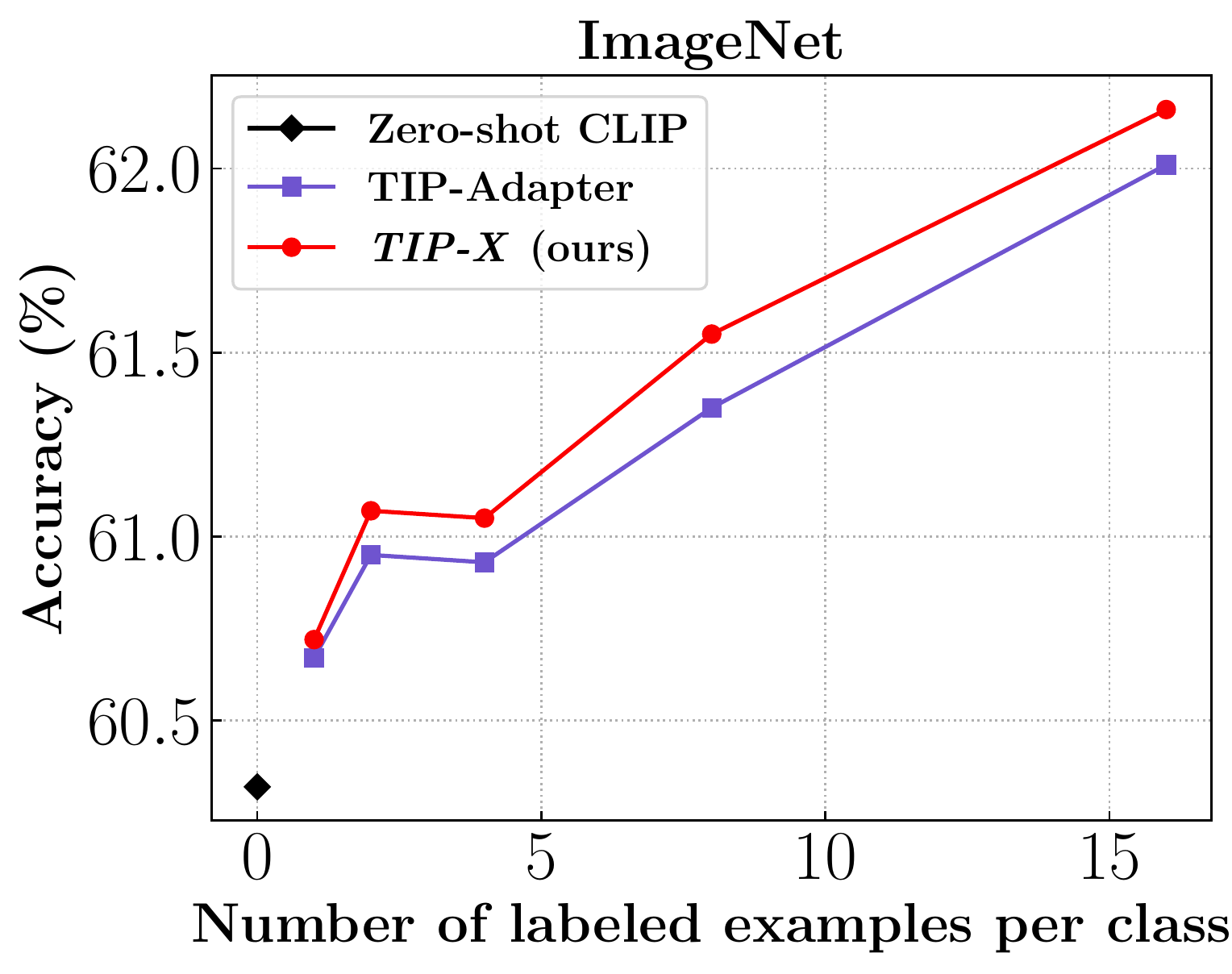}}\hfill
    \subfloat[Caltech101]{\includegraphics[width=0.3\textwidth]{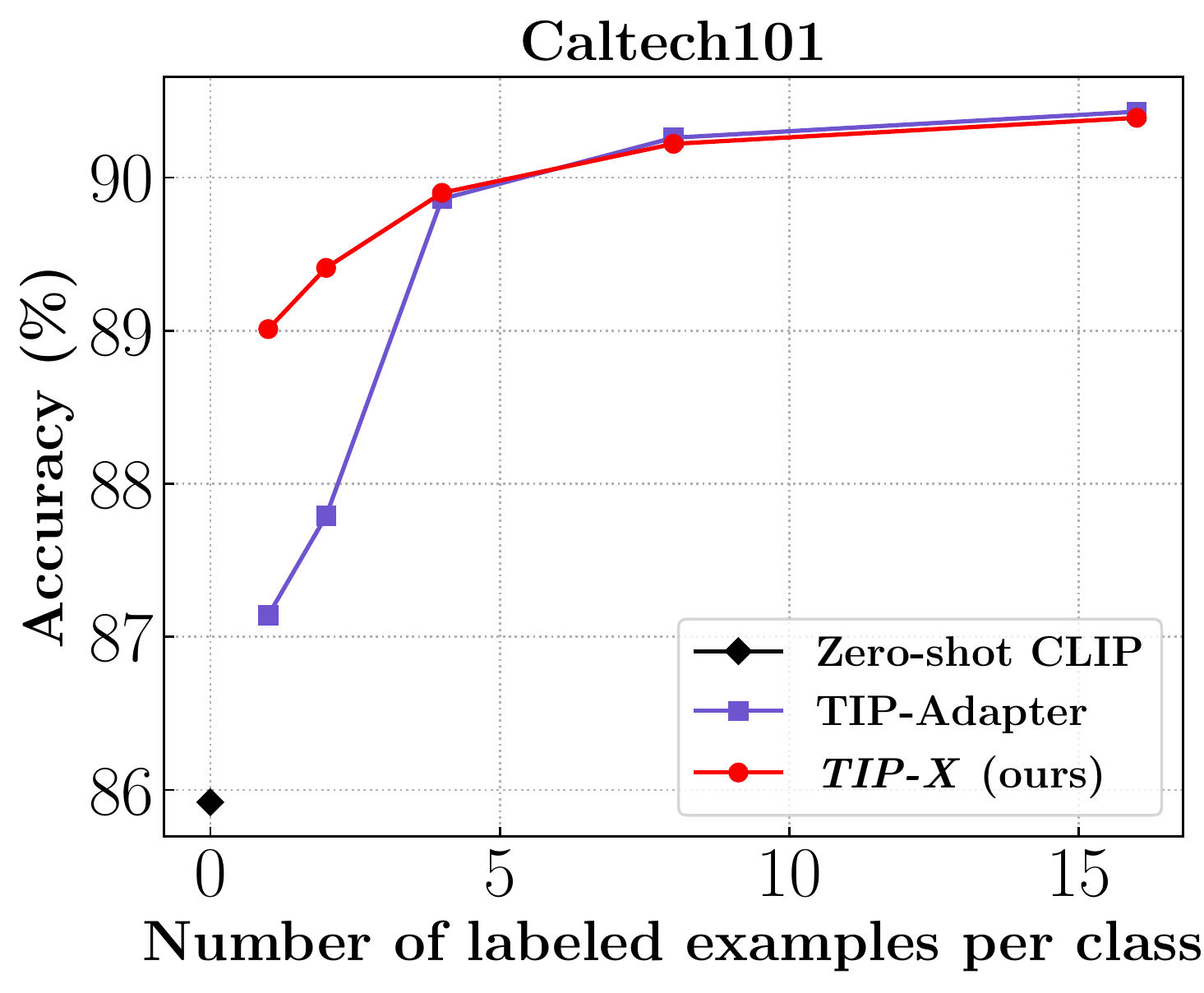}}\hfill\\

    \subfloat[OxfordPets]{\includegraphics[width=0.3\textwidth]{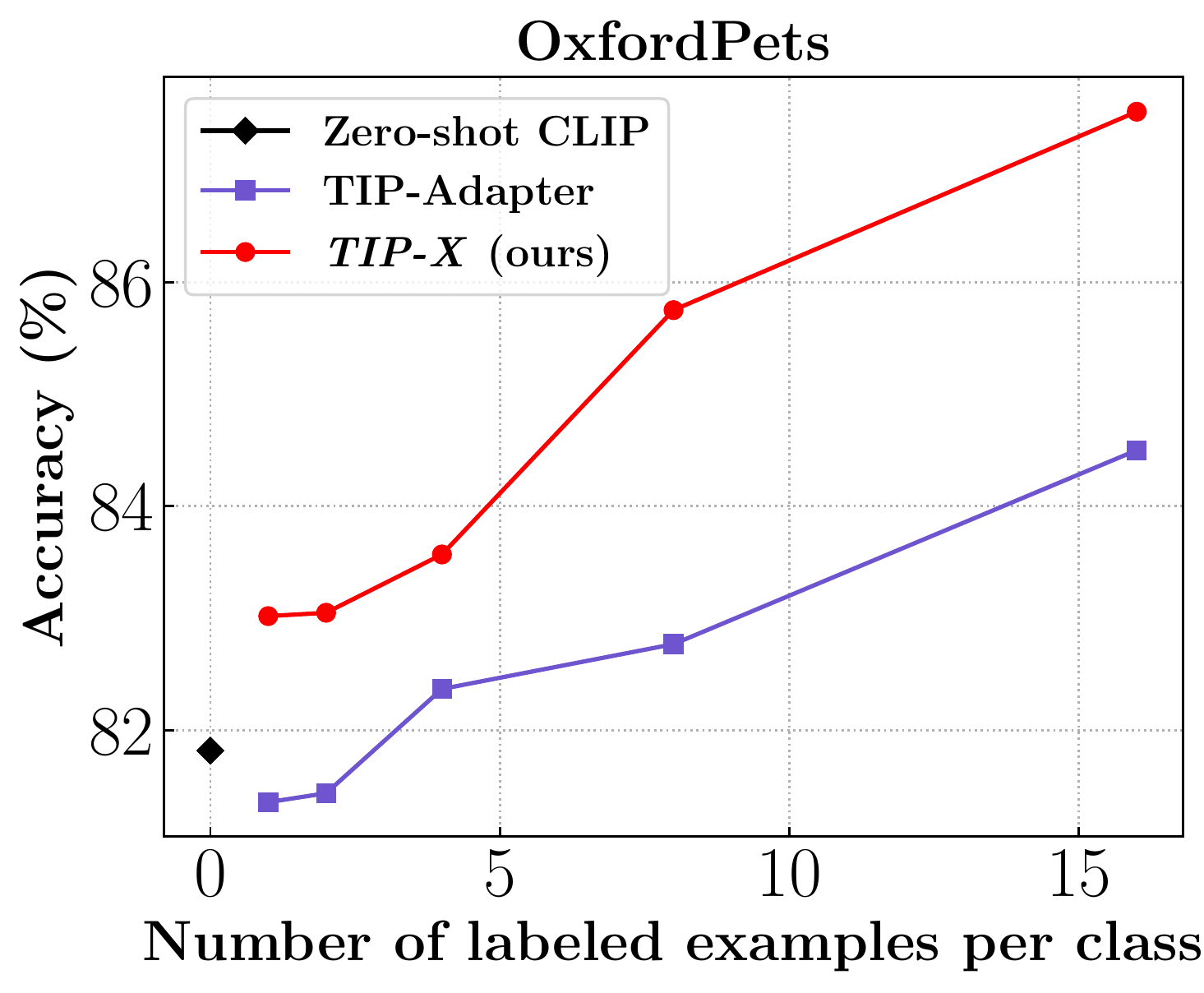}}\hfill
    \subfloat[StanfordCars]{\includegraphics[width=0.3\textwidth]{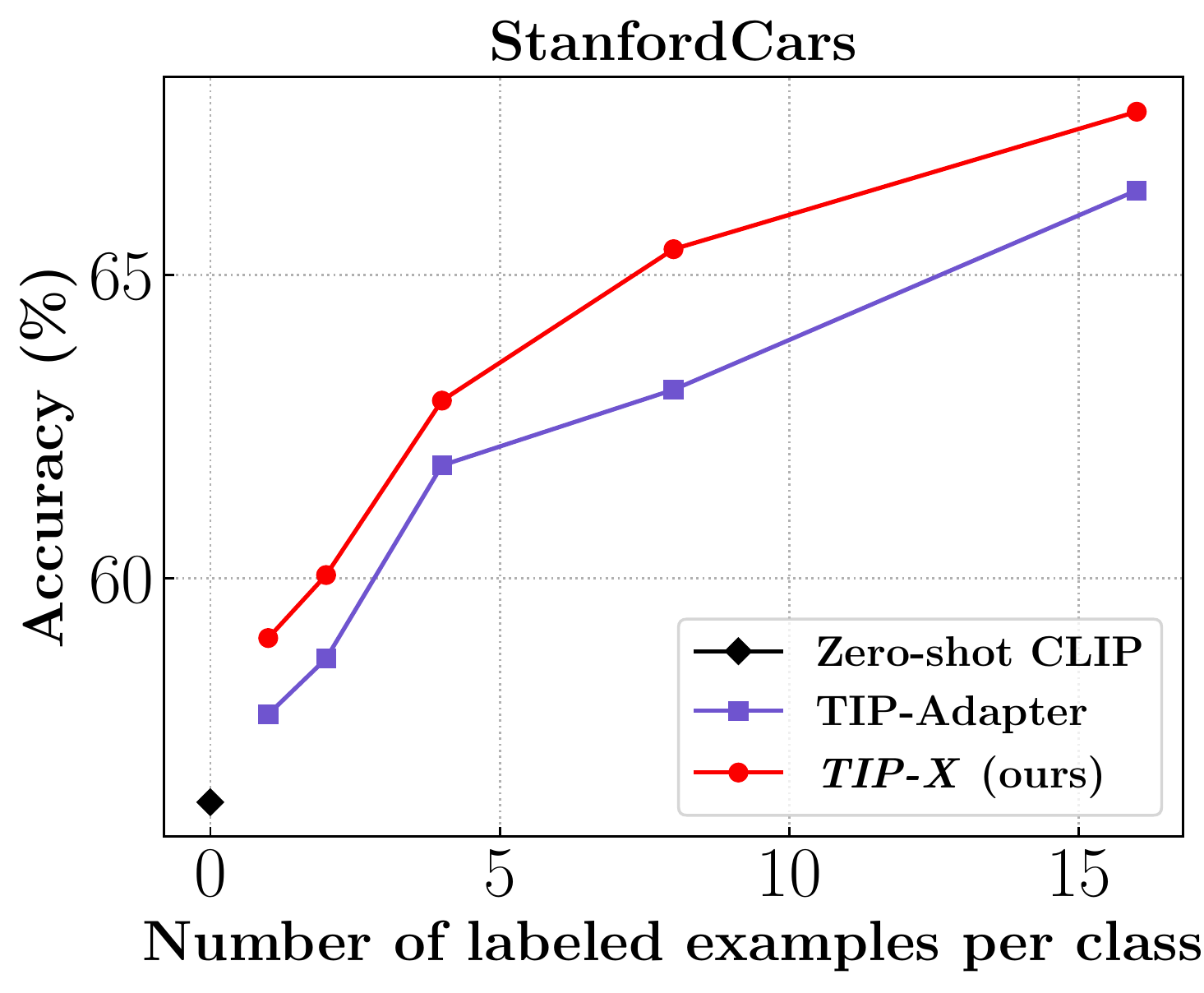}}\hfill
    \subfloat[Flowers102]{\includegraphics[width=0.3\textwidth]{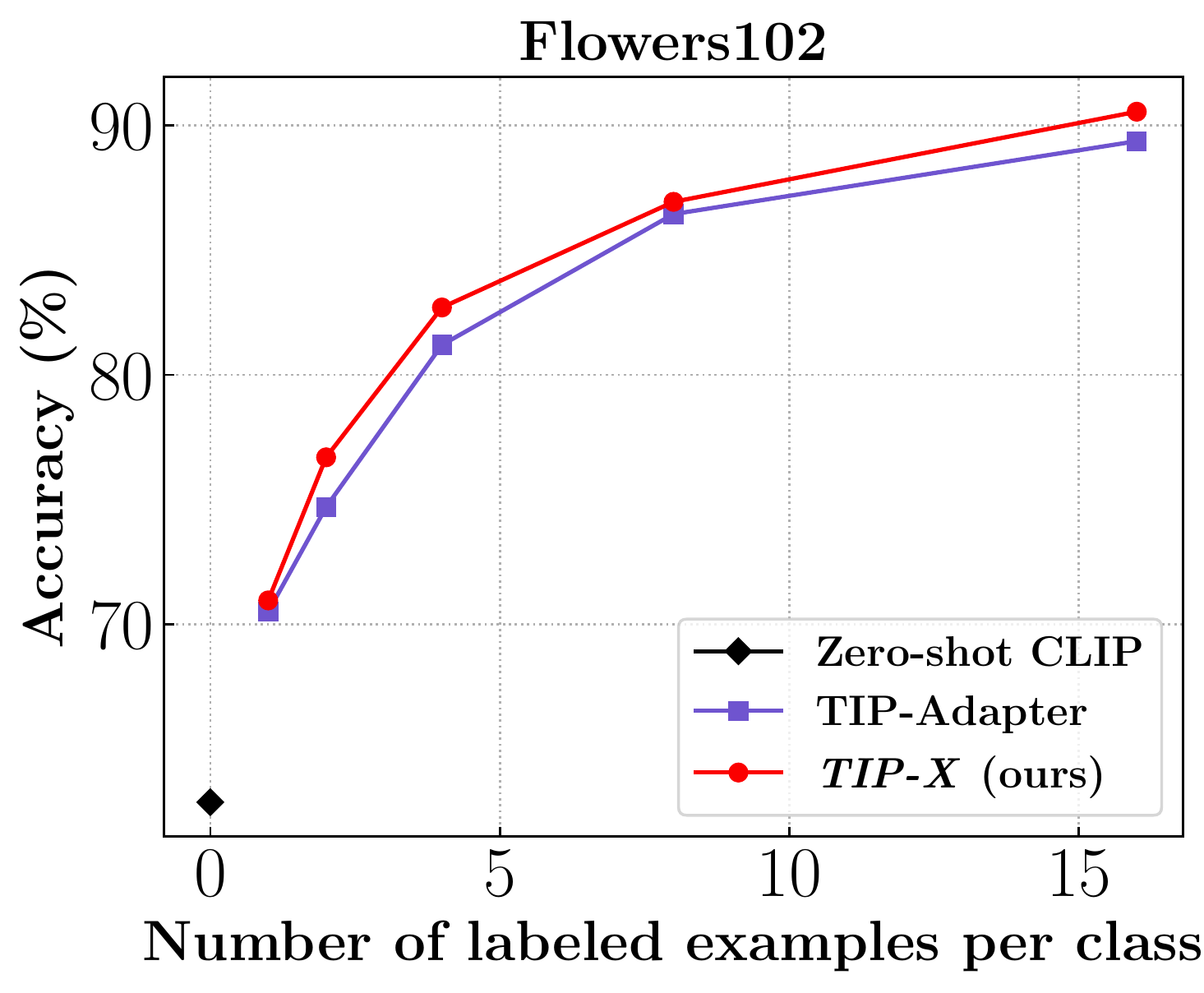}}\hfill\\

    \subfloat[Food101]{\includegraphics[width=0.3\textwidth]{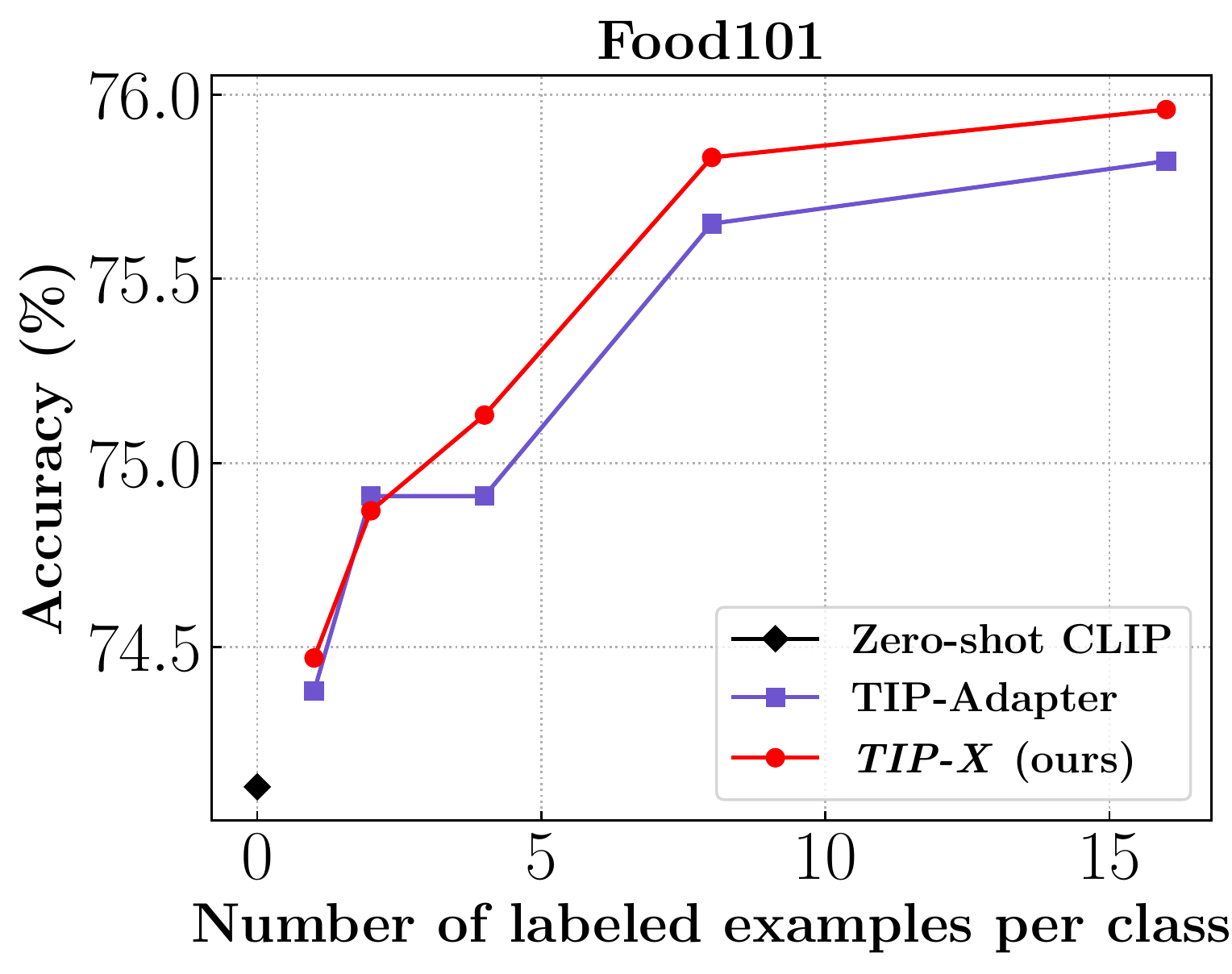}}\hfill
    \subfloat[FGVCAircraft]{\includegraphics[width=0.3\textwidth]{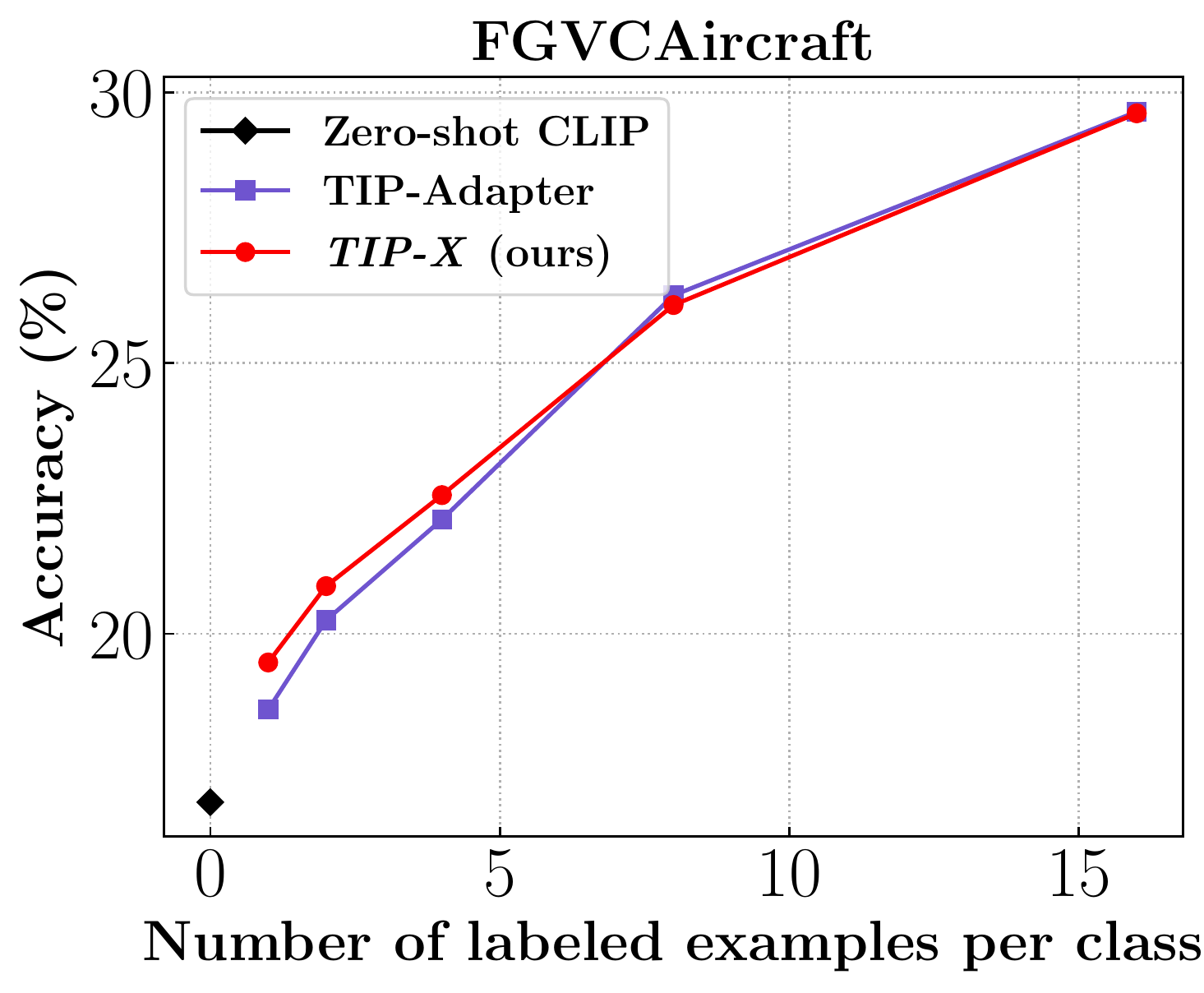}}\hfill
    \subfloat[SUN397]{\includegraphics[width=0.3\textwidth]{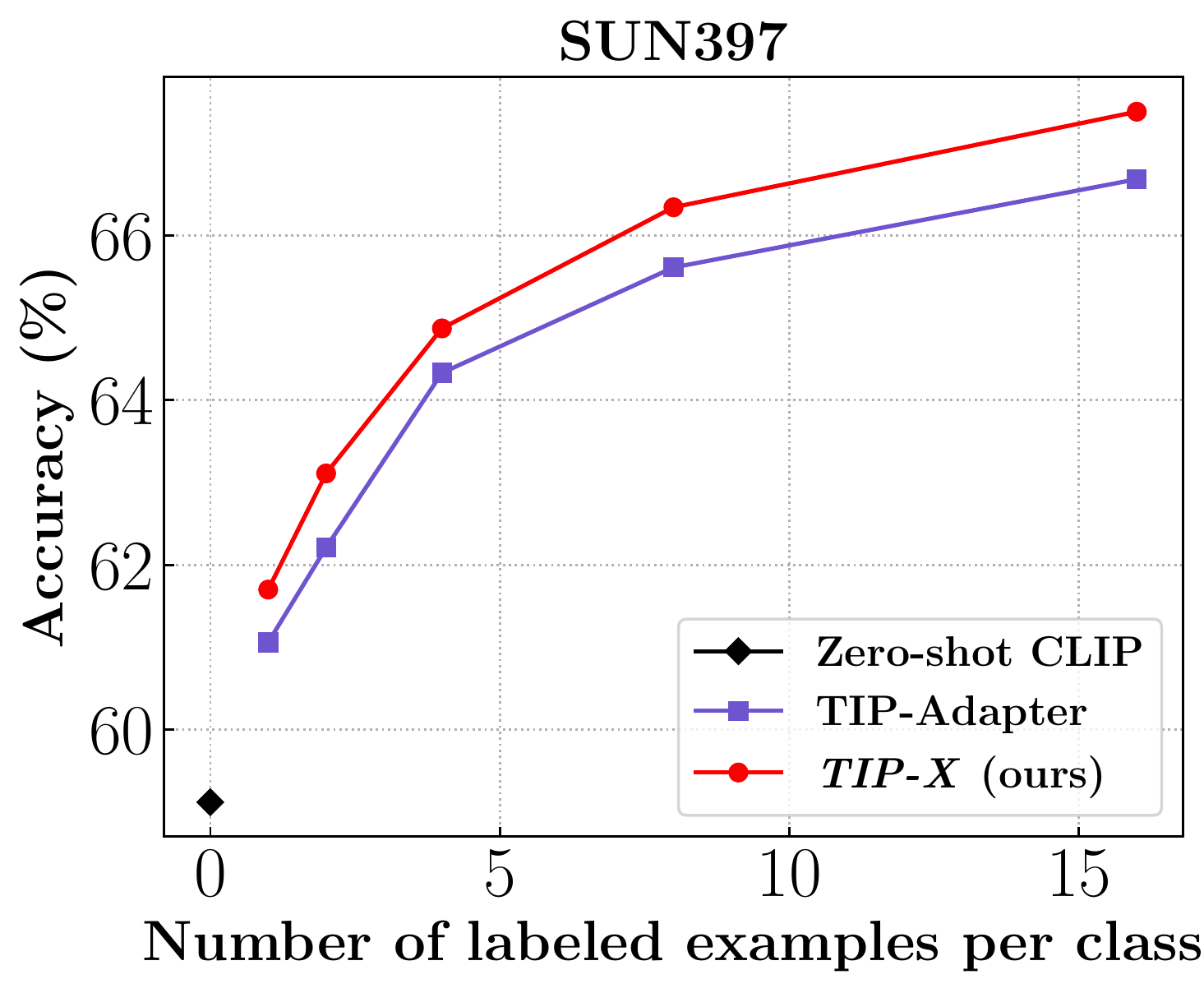}}\hfill\\

    \subfloat[DTD]{\includegraphics[width=0.3\textwidth]{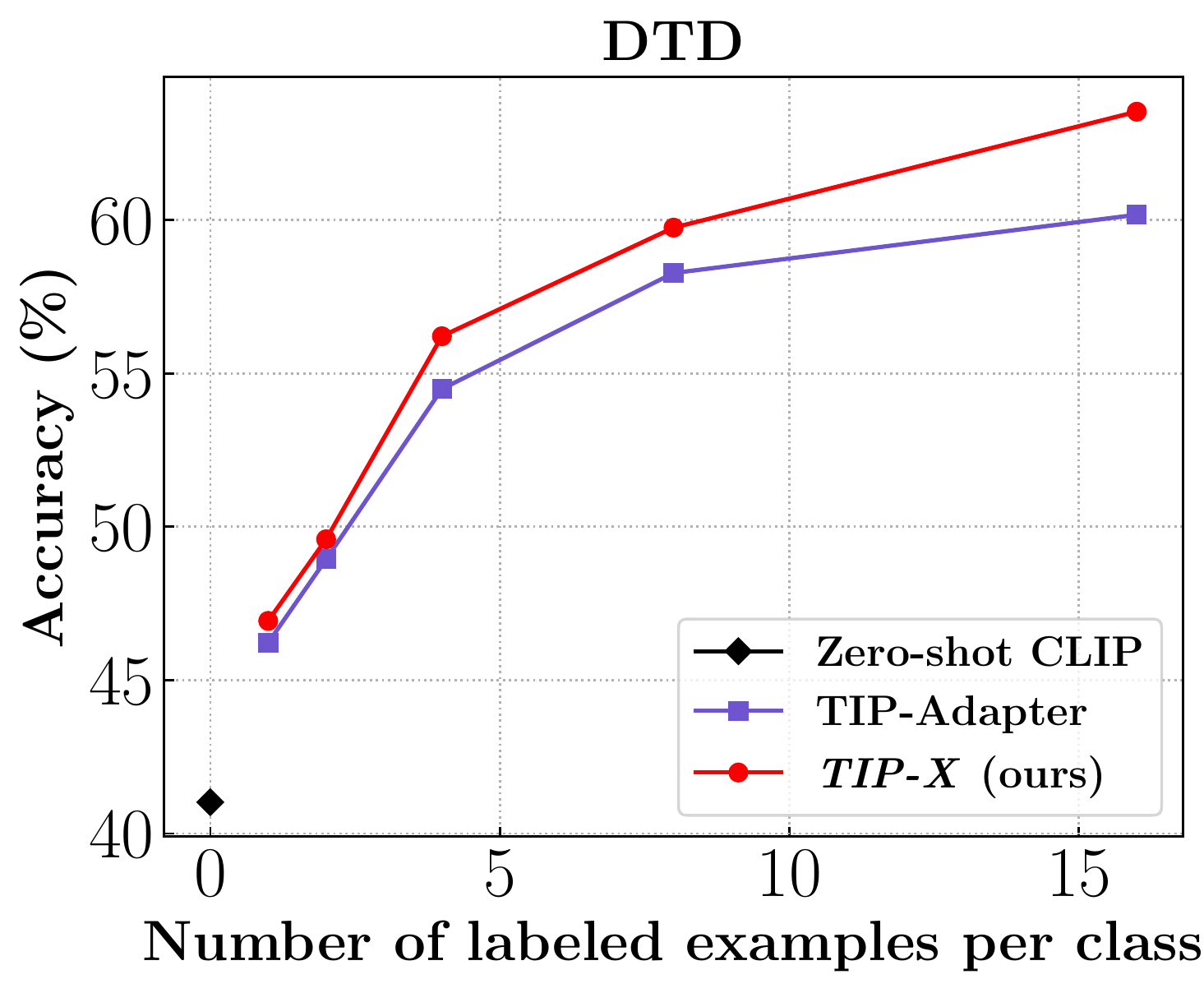}}\hfill
    \subfloat[EuroSAT]{\includegraphics[width=0.3\textwidth]{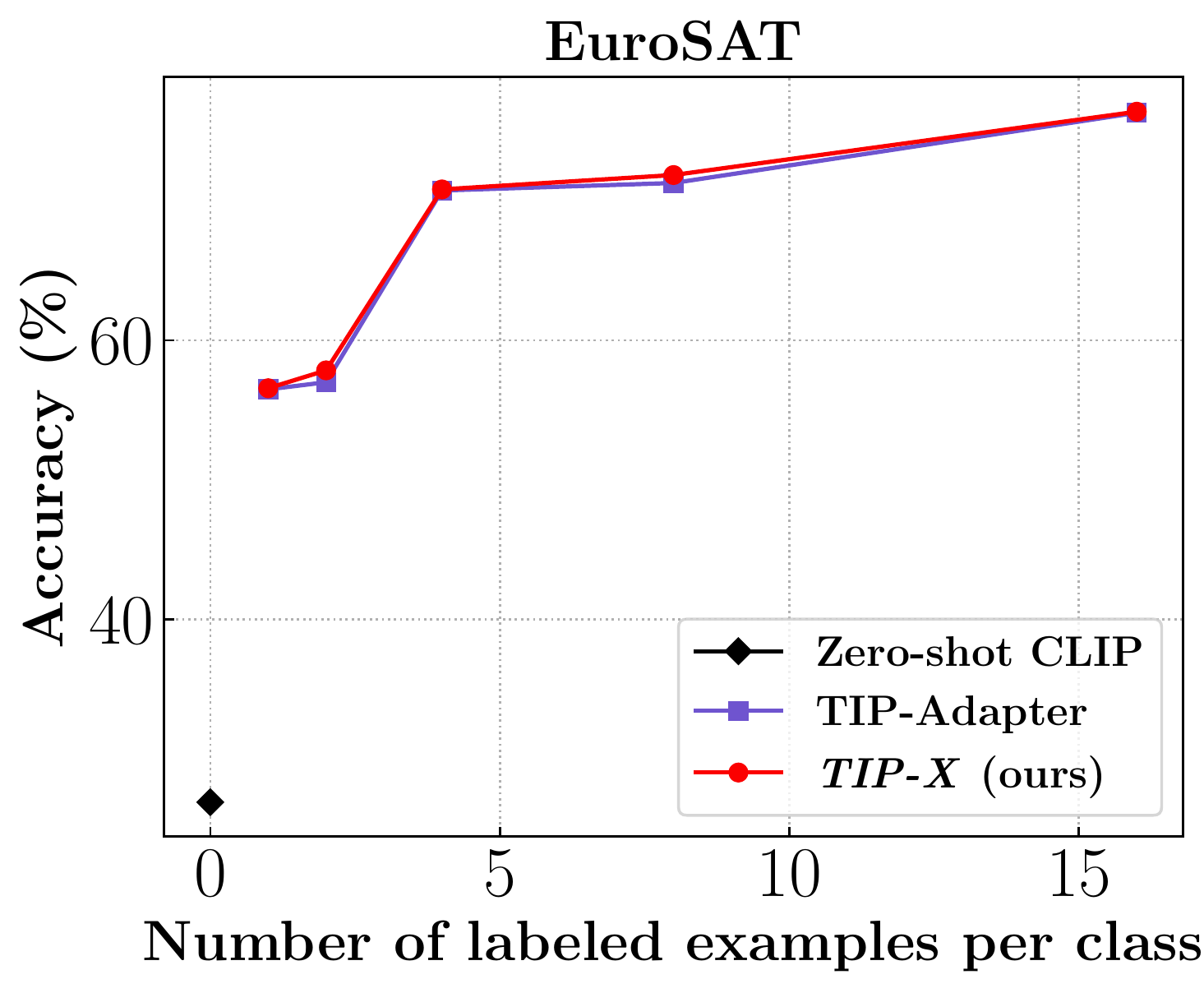}}\hfill
    \subfloat[UCF101]{\includegraphics[width=0.3\textwidth]{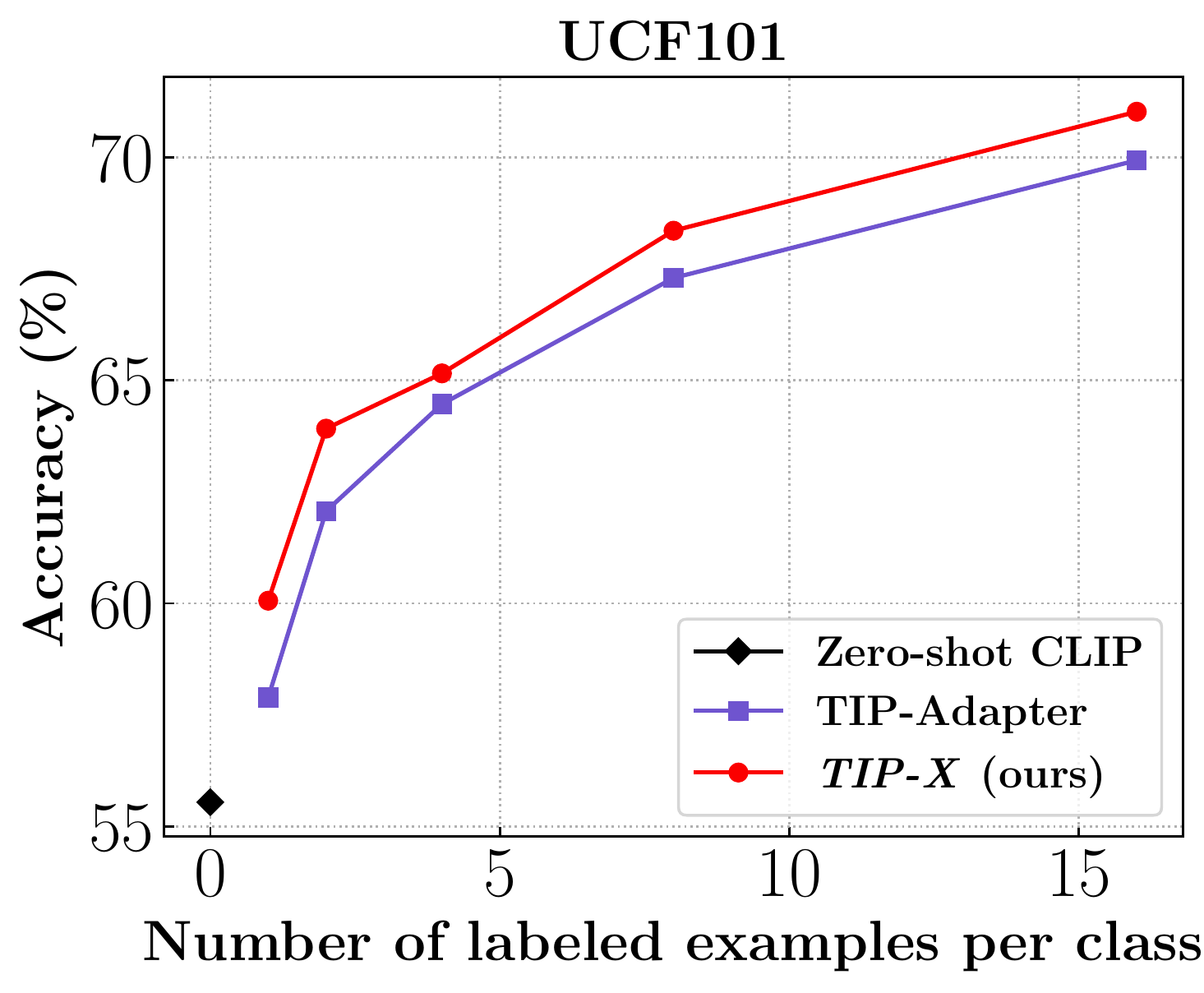}}\hfill\\

    \caption{\textbf{Results for the training-free few-shot regime across 11 datasets.}}
    \label{fig:tipx-vs-tip-few-shot}
\end{figure*}

In~\cref{adapt-tipx-few-shot}, we adapt \textit{TIP-X} to the few-shot training-free adaptation regime, and compare with the SoTA model TIP-Adapter. We now show the extended results on all 11 datasets in~\cref{fig:tipx-vs-tip-few-shot}. On average, we outperform TIP-Adapter by $0.91\%$ across all shots.

\newpage
\section{Details about Support Set Sizes}
For our main results in~\cref{training-free-section}, we use a fixed number of support set samples per dataset. In~\cref{tab:support-set-size}, we enumerate the number of support set samples used per dataset. As shown in~\cref{support-size-ablation}, the support set size can impact performance significantly---the nature of these impacts are dataset-specific.

\begin{table}[ht]
    \centering
    \caption{\textbf{Support Set Sizes}}    
    \begin{tabular}{cc}
    \toprule    
    \textbf{Dataset} & \textbf{Support Set Size}  \\
    \midrule
        UCF-101 & 5858 \\
        CIFAR-10 & 50 \\
        CIFAR-100 & 4700 \\
        Caltech101 & 101 \\
        Caltech256 & 3084  \\
        ImageNet & 36000  \\
        SUN397 & 397 \\
        FGVCAircraft & 7900 \\
        Birdsnap & 39000  \\
        StanfordCars & 980  \\
        CUB & 400 \\
        Flowers102 & 3162 \\
        Food101 & 3434  \\
        OxfordPets & 2627  \\
        DTD & 188  \\
        EuroSAT & 150 \\
        ImageNet-Sketch & 42000  \\
        ImageNet-R & 10200  \\
        Country211 & 844 \\
\bottomrule
    \end{tabular}
    \label{tab:support-set-size}
\end{table}

\newpage
\section{Details about Baselines}

For our main zero-shot/name-only training-free CLIP-based experiments, we use six main baselines---Zero-shot CLIP~\cite{radford2021learning}, CALIP~\cite{guo2022calip}, CLIP+DN~\cite{zhou2023distribution}, VisDesc~\cite{menon2022visual}, CuPL~\cite{pratt2022doescupl}
and CuPL+e. 

\noindent\textbf{Zero-shot CLIP.} For Zero-shot CLIP, we directly use the model weights released by OpenAI and the official repository for reproducing results on different datasets\footnote{\href{https://github.com/openai/CLIP}{https://github.com/openai/CLIP}}. For benchmarking all our results, we use the 7-prompt ensemble set used by TIP-Adapter~\cite{zhang2022tip} for all datasets. The 7 prompt templates in the ensemble are: ``itap of a $<$\texttt{class}$>$.'', ``a origami $<$\texttt{class}$>$.'', ``a bad photo of the $<$\texttt{class}$>$.'', ``a photo of the large $<$\texttt{class}$>$.'', ``a $<$\texttt{class}$>$ in a video game.'', ``art of the $<$\texttt{class}$>$.'', and ``a photo of the small $<$\texttt{class}$>$.''.

\noindent \textbf{CALIP details.} Due to the unavailability of publicly released code at the time of writing this paper, we re-implement the CALIP baseline, following the description in~\cite{guo2022calip}.
We provide access to our re-implementation as part of our released codebase.

\noindent \textbf{CLIP+DN details.} For CLIP+DN, we use the official code\footnote{\href{https://github.com/fengyuli2002/distribution-normalization}{https://github.com/fengyuli2002/distribution-normalization}} released by the authors on all datasets. As specified in the paper, we (i) use 100 random unlabeled validation samples for the mean estimation for DN, and (ii) report the average accuracy across 5 different random seeds.

\noindent \textbf{VisDesc details.} For VisDesc, we use the official code\footnote{\href{https://github.com/sachit-menon/classify_by_description_release}{https://github.com/sachit-menon/classify\_by\_description\_release}} released by the authors on all datasets. We use their default prompt settings for generating the GPT-3 descriptors.

\noindent \textbf{CuPL details.} For CuPL, we use the official code\footnote{\href{https://github.com/sarahpratt/CuPL}{https://github.com/sarahpratt/CuPL}} released by the authors on all datasets. The list of pre-prompts used as inputs to GPT-3 for different datasets are listed in~\cref{tab:cupl-prompts-1} and~\cref{tab:cupl-prompts-2}.

\noindent \textbf{CuPL+e details.} For CuPL+e, we simply concatenate the 7-prompt ensemble embeddings of each class with the custom GPT-3 generated CuPL embeddings of that particular class. We then average all the embeddings within a class to generate the textual embedding for that class. Then, we proceed as standard to construct the classifier weight matrix by stacking all class text embeddings.

\subsection{Transfer to other VLMs}
We can transfer all the aforementioned baselines to different VLMs by simply swapping out CLIP's frozen image and text encoders with those of TCL~\cite{yang2022visiontcl} and BLIP~\cite{li2022blip}. For the TCL\footnote{\href{https://github.com/uta-smile/TCL}{https://github.com/uta-smile/TCL}} experiments, we use the standard ViT-B/16 base model that is fine-tuned for retrieval on MS-COCO, released by the authors \href{https://drive.google.com/file/d/1PtcZF_XzJgIceg4rXLWqGQiXjizvxxS6/view}{here}. For the BLIP\footnote{\href{https://github.com/salesforce/BLIP}{https://github.com/salesforce/BLIP}} experiments, we use the standard ViT-B/16 base model fine-tuned for retrieval on MS-COCO, released by the authors \href{https://storage.googleapis.com/sfr-vision-language-research/BLIP/models/model_base_retrieval_coco.pth}{here}.

\newpage

\begin{table*}[ht]
    \centering
    \caption{\textbf{CuPL hand-written prompts (1/2)}}    
    \begin{tabular}{ll}
    \toprule
        \textbf{Dataset} & \textbf{GPT-3 prompts} \\
        \midrule
        
        \multirow{3}{*}{}{UCF101} & ``What does a person doing \{\} look like''\\
        & ``Describe the process of \{\}''\\
		& ``How does a person \{\}'' \\
  \midrule

        \multirow{5}{*}{}{CIFAR10} & ``Describe what a \{\} looks like''\\
        & ``How can you identify \{\}?''\\
	&	``What does \{\} look like?''\\
	&	``Describe an image from the internet of a \{\}''\\
	&	``A caption of an image of \{\}: '' \\
  \midrule

        \multirow{5}{*}{}{CIFAR100} & ``Describe what a \{\} looks like''\\
        & ``How can you identify \{\}?''\\
	&	``What does \{\} look like?''\\
	&	``Describe an image from the internet of a \{\}''\\
	&	``A caption of an image of \{\}: '' \\
  \midrule

        \multirow{3}{*}{}{Caltech101} & ``Describe what a \{\} looks like''\\
		& ``What does a \{\} look like''\\
	&	``Describe a photo of a \{\}''\\
  \midrule

        \multirow{3}{*}{}{Caltech256} & ``Describe what a \{\} looks like''\\
		& ``What does a \{\} look like''\\
	&	``Describe a photo of a \{\}''\\
  \midrule

        \multirow{5}{*}{}{ImageNet} & ``Describe what a \{\} looks like''\\
	&	``How can you identify \{\}?''\\
       & ``What does \{\} look like?''\\
	&	``Describe an image from the internet of a \{\}''\\
	& ``A caption of an image of \{\}: ''\\
  \midrule

        \multirow{3}{*}{}{SUN397} & ``Describe what a \{\} looks like''\\
		& ``How can you identify a \{\}?''\\
		& ``Describe a photo of a \{\}''\\
  \midrule

        \multirow{1}{*}{}{FGVCAircraft} & ``Describe a \{\} aircraft''\\
  \midrule

        \multirow{6}{*}{}{Birdsnap} & ``Describe what a \{\}, a species of bird, looks like''\\
		& ``What does a \{\} look like''\\
		& ``Visually describe a \{\}, a type of bird''\\
		& ``A caption of an image of a \{\}, a type of bird''\\
		& ``Describe the appearance of a \{\}''\\
		& ``What are the prominent features to identify a \{\} bird''\\
  \midrule

        \multirow{9}{*}{}{StanfordCars} & ``How can you identify a \{\}''\\
		& ``Description of a \{\}, a type of car''\\
		& ``A caption of a photo of a \{\}:''\\
		& ``What are the primary characteristics of a \{\}?''\\
		& ``Description of the exterior of a \{\}''\\
		& ``What are the identifying characteristics of a \{\}, a type of car?''\\
		& ``Describe an image from the internet of a \{\}''\\
		& ``What does a \{\} look like?''\\
		& ``Describe what a \{\}, a type of car, looks like''\\
  \bottomrule
\end{tabular}
    \label{tab:cupl-prompts-1}
\end{table*}

\begin{table*}[ht]
    \centering
    \caption{\textbf{CuPL hand-written prompts (2/2)}}    
    \begin{tabular}{ll}
    \toprule
        \textbf{Dataset} & \textbf{GPT-3 prompts} \\

        \midrule
        \multirow{6}{*}{}{CUB} & ``Describe what a \{\}, a species of bird, looks like''\\
		& ``What does a \{\} look like''\\
		& ``Visually describe a \{\}, a type of bird''\\
		& ``A caption of an image of a \{\}, a type of bird''\\
		& ``Describe the appearance of a \{\}''\\
		& ``What are the prominent features to identify a \{\} bird''\\
  \midrule

        \multirow{4}{*}{}{Flowers102} & ``What does a \{\} flower look like''\\
		& ``Describe the appearance of a \{\}''\\
		& ``A caption of an image of \{\}''\\
		& ``Visually describe a \{\}, a type of flower''\\
  \midrule

  \multirow{3}{*}{}{Food101} & ``Describe what a \{\} looks like''\\
		& ``Visually describe a \{\}''\\
	& ``How can you tell that the food in this photo is a \{\}?''\\
 \midrule

  \multirow{2}{*}{}{OxfordPets} & ``Describe what a \{\} pet looks like''\\
 & ``Visually describe a \{\}, a type of pet''\\
 \midrule

  \multirow{6}{*}{}{DTD} & ``What does a \{\} material look like?''\\
	&	``What does a \{\} surface look like?''\\
	&	``What does a \{\} texture look like?''\\
	&	``What does a \{\} object look like?''\\
	&	``What does a \{\} thing look like?''\\
	&	``What does a \{\} pattern look like?''\\
 \midrule

  \multirow{3}{*}{}{EuroSAT} & ``Describe an aerial satellite view of \{\}''\\
	&	``How does a satellite photo of a \{\} look like''\\
	&	``Visually describe a centered satellite view of a \{\}''\\
 \midrule

  \multirow{3}{*}{}{ImageNet-Sketch} & ``Describe how a black and white sketch of a \{\} looks like''\\
	&	``A black and white sketch of a \{\}''\\
	&	``Describe a black and white sketch from the internet of a \{\}''\\
 \midrule

  \multirow{12}{*}{}{ImageNet-R} & ``An art drawing of a \{\}''\\
	&	``Artwork showing a \{\}''\\
	&	``A cartoon a \{\}''\\
	&	``An origami of a \{\}''\\
	&	``A deviant art photo depicting a \{\}''\\
	&	``An embroidery of a {}''\\
	&	``A graffiti art showing a \{\}''\\
	&	``A painting of a \{\}''\\
	&	``A sculpture of a \{\}''\\
	&	``A black and white sketch of \{\}''\\
	&	``A toy of a \{\}''\\
	&	``A videogame of a \{\}''\\
 \midrule

  \multirow{12}{*}{}{Country211} & ``Visually describe what \{\} looks like''\\
	&	``What does the landscape of \{\} look like''\\
	&	``Describe a photo taken in \{\}''\\
	&	``How does a typical photo taken in \{\} look like''\\
 \bottomrule 
\end{tabular}    
    \label{tab:cupl-prompts-2}
\end{table*}

\section{More \textit{SuS} Visualisations}

\begin{figure*}[ht]    
  \subfloat[Birdsnap, Acadian Flycatcher]{%
  \begin{minipage}{0.48\linewidth}
  \centering
  \includegraphics[width=0.3\textwidth]{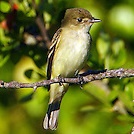}
  \includegraphics[width=0.3\textwidth]{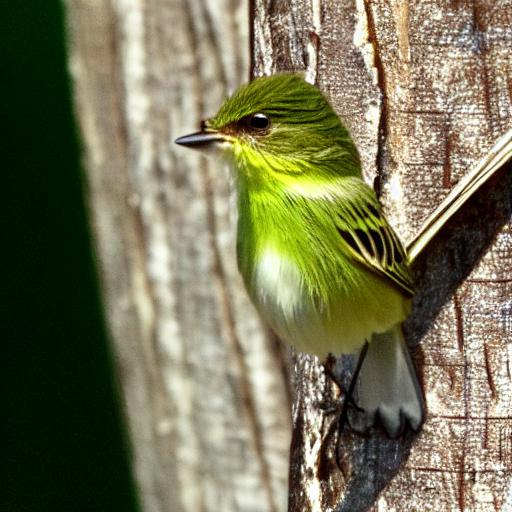}
  \includegraphics[width=0.3\textwidth]{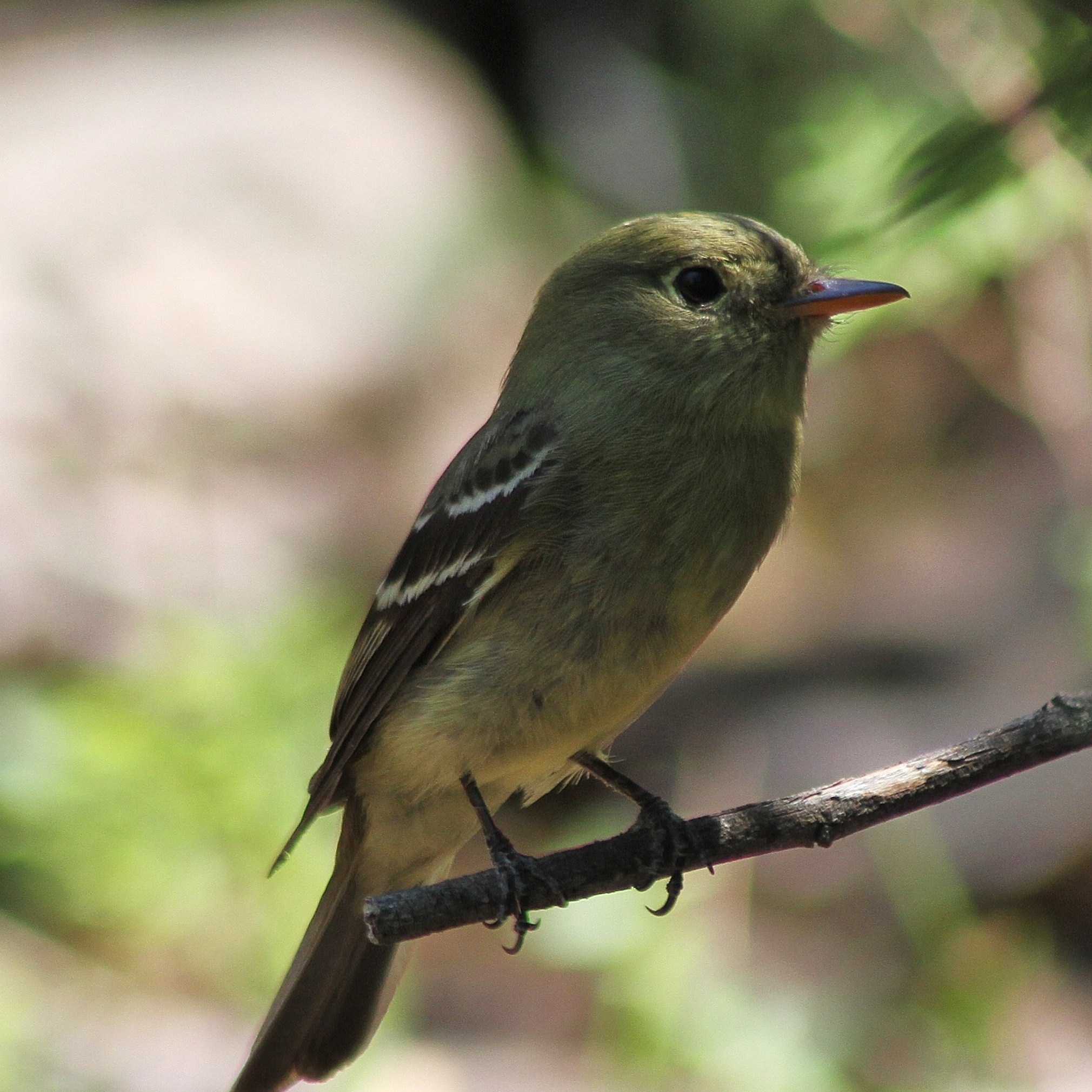}
  \end{minipage}%
  }
  \subfloat[Caltech101, Soccer Ball]{%
  \begin{minipage}{0.48\linewidth}
  \centering
  \includegraphics[width=0.3\textwidth]{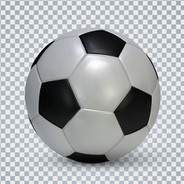}
  \includegraphics[width=0.3\textwidth]{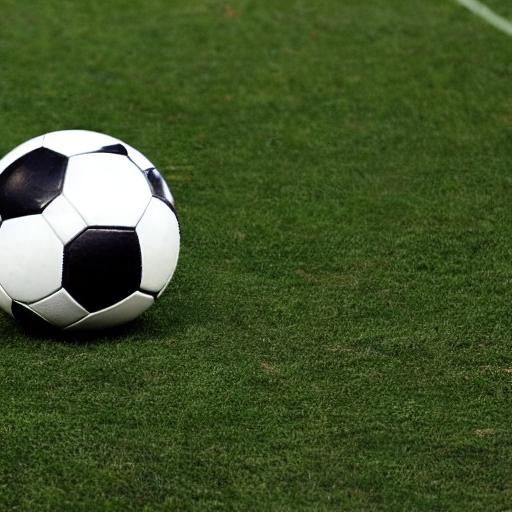}
  \includegraphics[width=0.3\textwidth]{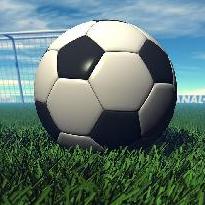}
  \end{minipage}%
  }
  \par
    \subfloat[DTD, Chequered]{%
  \begin{minipage}{0.48\linewidth}
  \centering
  \includegraphics[width=0.3\textwidth]{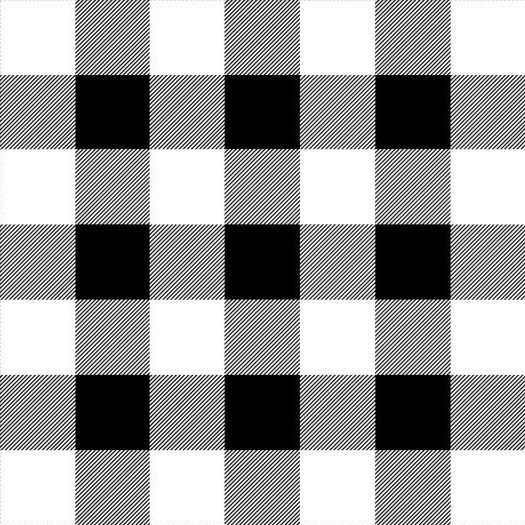}
  \includegraphics[width=0.3\textwidth]{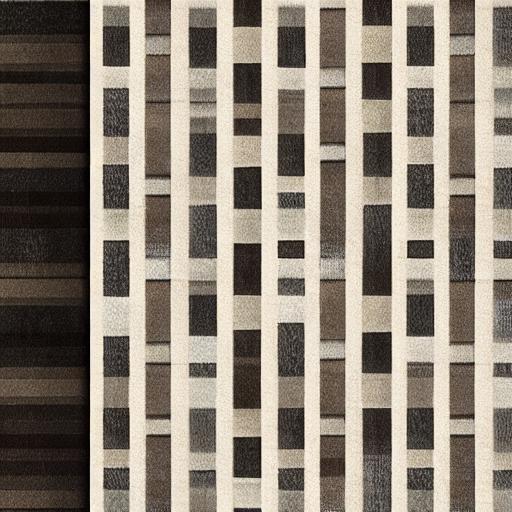}
  \includegraphics[width=0.3\textwidth]{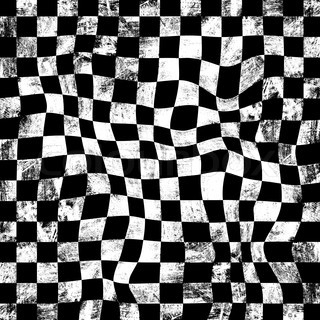}
  \end{minipage}%
  }
  \subfloat[EuroSAT, Residential]{%
  \begin{minipage}{0.48\linewidth}
  \centering
  \includegraphics[width=0.3\textwidth]{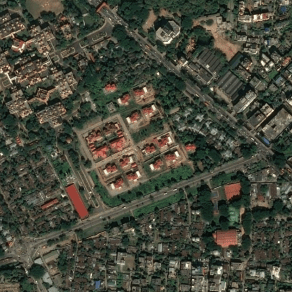}
  \includegraphics[width=0.3\textwidth]{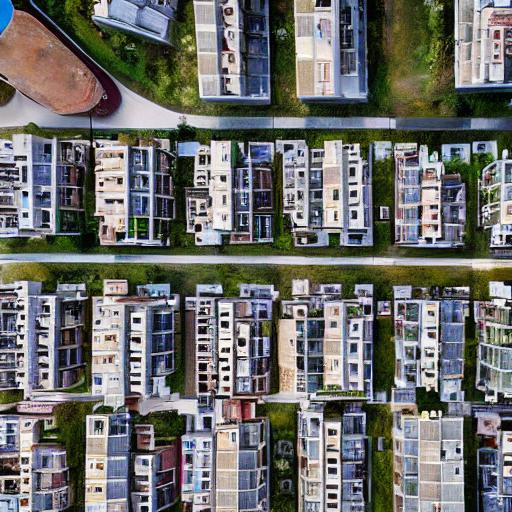}
    \includegraphics[width=0.3\textwidth]{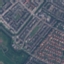}
  \end{minipage}%
  }

  \subfloat[FGVCAircraft, A320]{%
  \begin{minipage}{0.48\linewidth}
  \centering
  \includegraphics[width=0.3\textwidth]{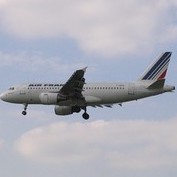}
  \includegraphics[width=0.3\textwidth]{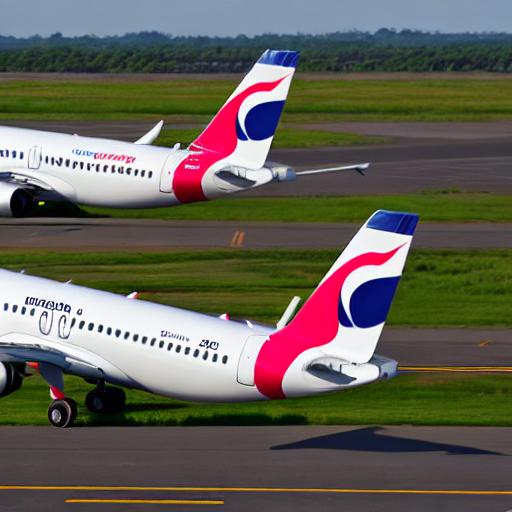}
  \includegraphics[width=0.3\textwidth]{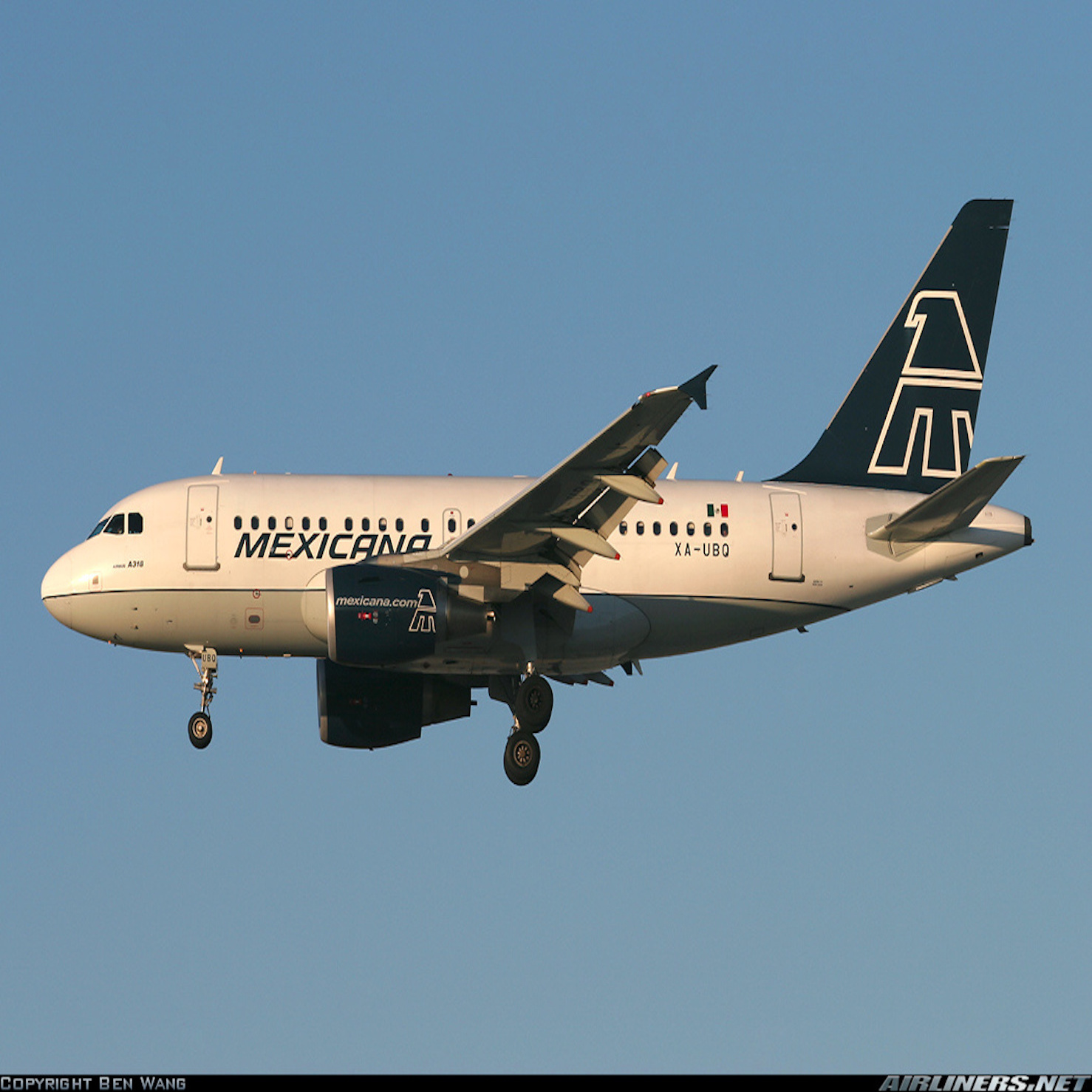}
  \end{minipage}%
  }
  \subfloat[Food101, Breakfast Burrito]{%
  \begin{minipage}{0.48\linewidth}
  \centering
  \includegraphics[width=0.3\textwidth]{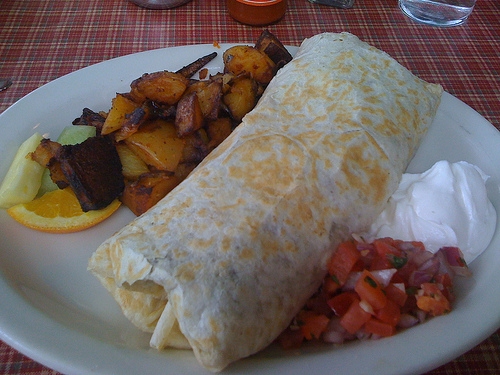}
  \includegraphics[width=0.3\textwidth]{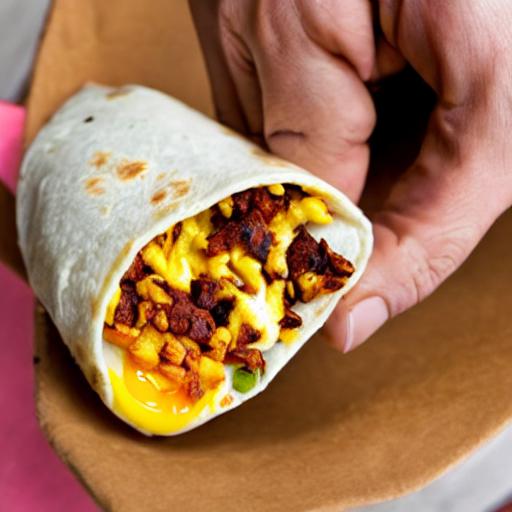}
  \includegraphics[width=0.3\textwidth]{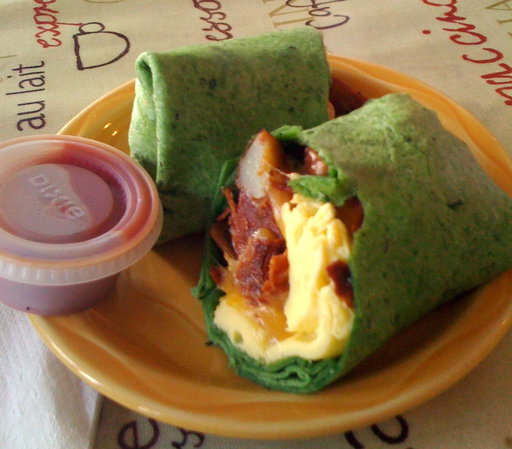}
  \end{minipage}%
  }
  \par
    \subfloat[Flowers102, Water Lily]{%
  \begin{minipage}{0.48\linewidth}
  \centering
  \includegraphics[width=0.3\textwidth]{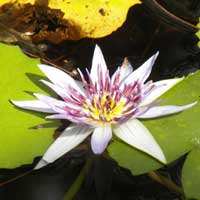}
  \includegraphics[width=0.3\textwidth]{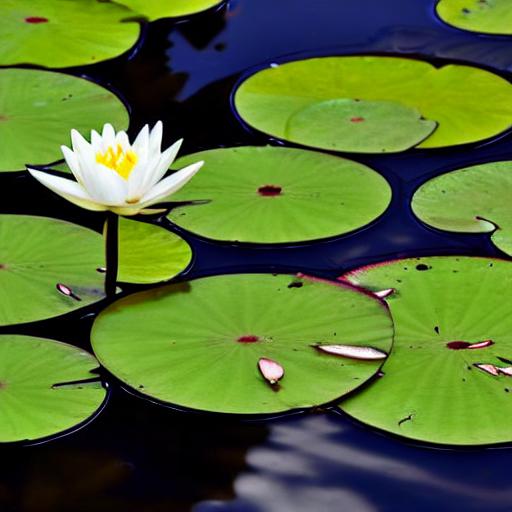}
  \includegraphics[width=0.3\textwidth]{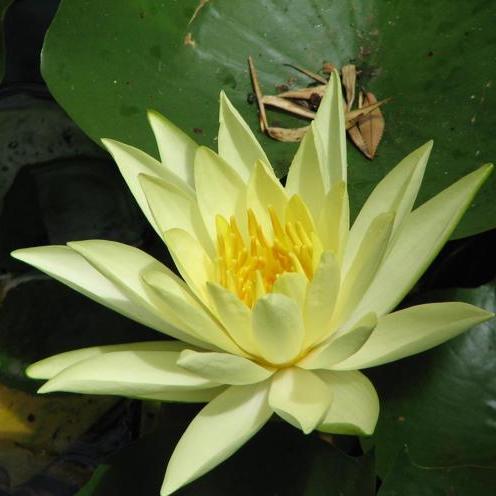}
  \end{minipage}%
  }
  \subfloat[OxfordPets, Persian Cat]{%
  \begin{minipage}{0.48\linewidth}
  \centering
  \includegraphics[width=0.3\textwidth]{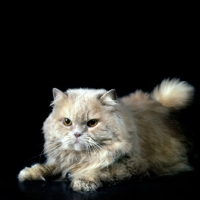}
  \includegraphics[width=0.3\textwidth]{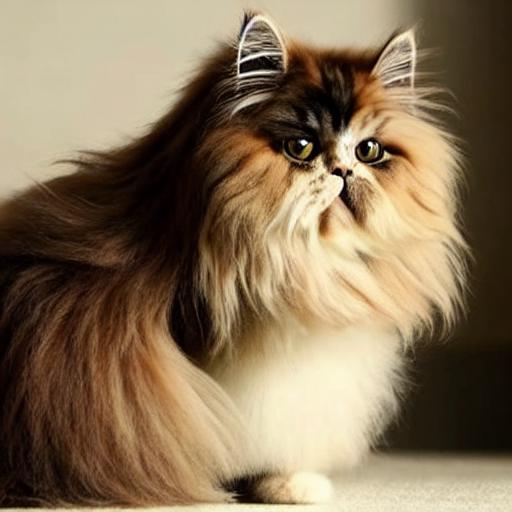}
    \includegraphics[width=0.3\textwidth]{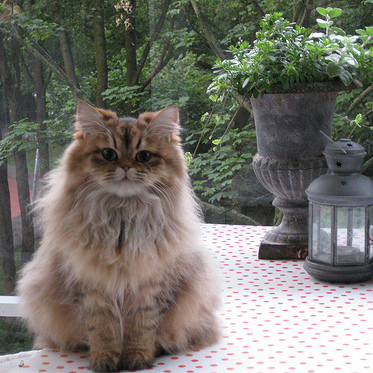}
  \end{minipage}%
  }
  \par
    \subfloat[StanfordCars, Rolls Royce Ghost]{%
  \begin{minipage}{0.48\linewidth}
  \centering
  \includegraphics[width=0.3\textwidth]{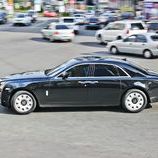}
  \includegraphics[width=0.3\textwidth]{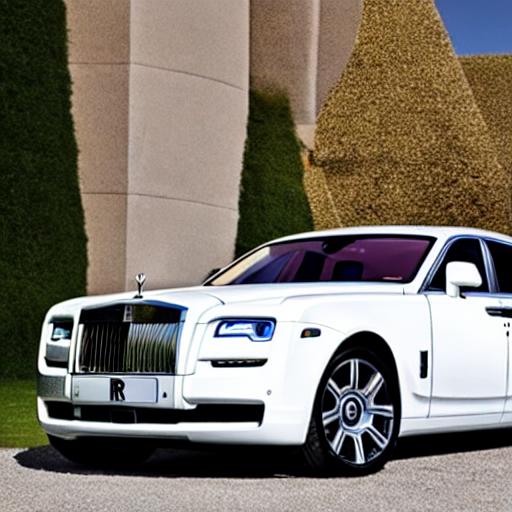}
  \includegraphics[width=0.3\textwidth]{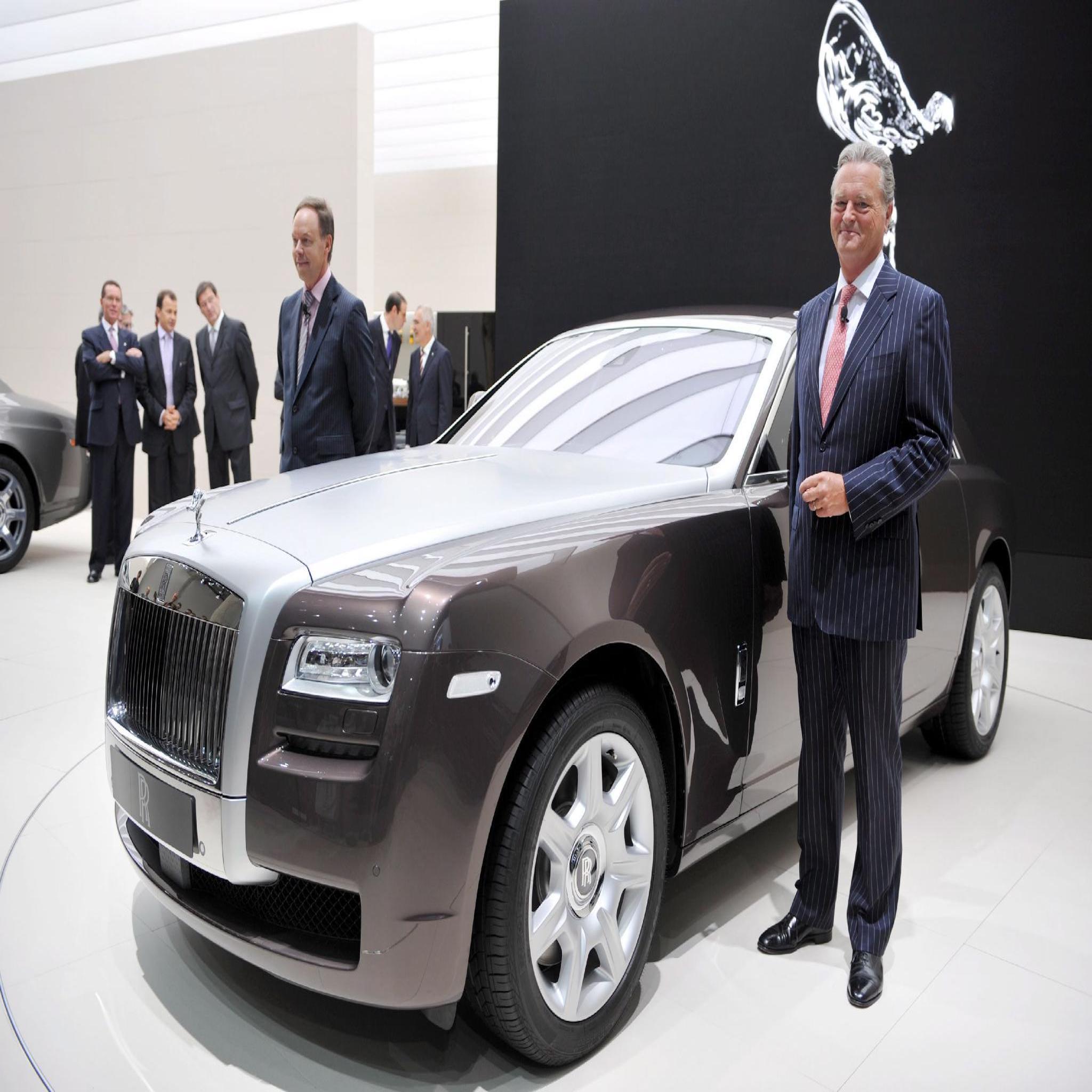}
  \end{minipage}%
  }
  \subfloat[UCF101, Cricket Shot]{%
  \begin{minipage}{0.48\linewidth}
  \centering
  \includegraphics[width=0.3\textwidth]{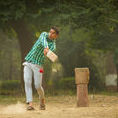}
  \includegraphics[width=0.3\textwidth]{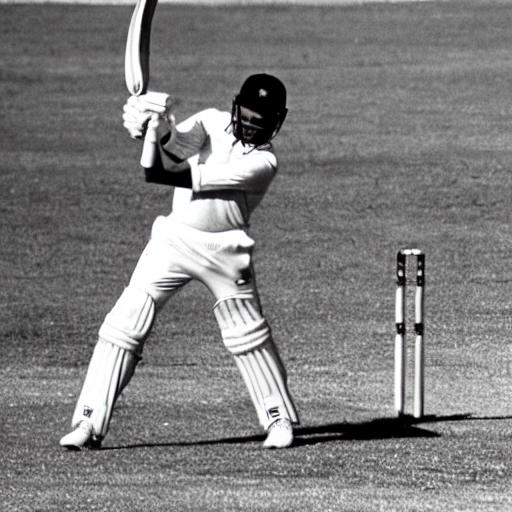}
    \includegraphics[width=0.3\textwidth]{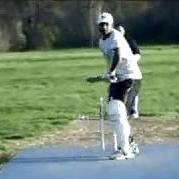}
  \end{minipage}%
  }  
  \caption{\textbf{Support samples from the generated \textit{SuS-SD}, retrieved \textit{SuS-LC} and true training distribution for different datasets.} 
  For each subfigure, the ordering of figures is---\textit{SuS-LC}, \textit{SuS-SD}, \textit{Train}. We label each figure with its source dataset and class name.}
  \label{fig: sus-examples-appendix}
\end{figure*}

In~\cref{fig: sus-examples-appendix}, we provide further support set samples across different datasets curated using both \textit{SuS-LC} and \textit{SuS-SD} methods.

\newpage

\section{Hyperparameter Settings}

We provide the hyperparameter settings for obtaining our main results from~\cref{main-results} in~\cref{tab:hparams}. For our hyperparameters, we conduct a search over $[0.1, 50]$ for $\alpha$, $[1, 50]$ for $\beta$ and $[0.1, 30]$ for $\gamma$. In the main paper, we have a hyperparameter sensitivity test which ensures that the variance in accuracy values is not too large as we vary our hyperparameters.

\begin{table}[h]
    \centering
    \caption{\textbf{Hyperparameter settings for the 19 datasets.}}
    \label{tab:hparams}    
    \begin{tabular}{cccc}
    \midrule
    \textbf{Dataset} & $\boldsymbol{\alpha}$ & $\boldsymbol{\beta}$ & $\boldsymbol{\gamma}$ \\
    \midrule
        UCF-101 & 0.10 & 8.59 & 0.10  \\
        CIFAR-10 & 5.09 & 5.41 & 0.10 \\
        CIFAR-100 & 0.10 & 1.49 & 0.10  \\
        Caltech101 & 0.10 & 1.27 & 0.10  \\
        Caltech256 & 0.10 & 12.76 & 0.10 \\
        ImageNet & 10.08 & 39.46 & 0.10 \\
        SUN397 & 2.60 & 8.35 & 0.10 \\
        FGVCAircraft & 2.60 & 24.52 & 0.69 \\
        Birdsnap & 48.53 & 22.55 & 0.69 \\
        StanfordCars & 0.10 & 1.58 & 0.10 \\
        CUB & 0.10 & 8.84 & 0.10 \\
        Flowers102 & 0.10 & 2.72 & 0.10  \\
        Food101 & 17.56 & 49.02 & 0.10 \\
        OxfordPets & 10.08 & 41.91 & 1.29 \\
        DTD & 5.09 & 23.79 & 0.70 \\
        EuroSAT & 2.60 & 1.00 & 0.10 \\
        ImageNet-Sketch & 30.04 & 38.48 & 0.69 \\
        ImageNet-R & 2.60 & 30.65 & 0.70 \\
        Country211 & 12.57 & 22.31 & 0.10 \\
    \bottomrule
    \end{tabular}
\end{table}

\noindent\textbf{Results without tuning.} We also report the results on all 19 datasets {without} {tuning} our hyperparameters in~\cref{tab:fixed-hparams}. For this, we fix the hyperparameters to be ${\alpha}{=}{0.1}$, ${\beta}{=}{1.0}$, ${\gamma}{=}{0.1}$. Even without hyperparameter tuning, we see large gains over Zero-shot CLIP.

\begin{table*}[h]
    \footnotesize
    \centering
    \setlength\tabcolsep{2pt}
    \caption{\textbf{Zero-shot/name-only results with fixed hyperparameters (no hyperparameter tuning)}}
    \begin{tabular}{c|ccccccccccccccccccc|cc}
    \toprule
         & \rotatebox{90}{\textbf{UCF101}} & \rotatebox{90}{\textbf{CIFAR-10}} & \rotatebox{90}{\textbf{CIFAR-100}} & \rotatebox{90}{\textbf{Caltech101}} & \rotatebox{90}{\textbf{Caltech256}} & \rotatebox{90}{\textbf{ImageNet}} & \rotatebox{90}{\textbf{SUN397}} & \rotatebox{90}{\textbf{FGVCAircraft}} & \rotatebox{90}{\textbf{Birdsnap}} & \rotatebox{90}{\textbf{StanfordCars}} & \rotatebox{90}{\textbf{CUB}} & \rotatebox{90}{\textbf{Flowers102}} & \rotatebox{90}{\textbf{Food101}} & \rotatebox{90}{\textbf{OxfordPets}} & \rotatebox{90}{\textbf{DTD}} & \rotatebox{90}{\textbf{EuroSAT}} & \rotatebox{90}{\textbf{ImageNet-Sketch}} & \rotatebox{90}{\textbf{ImageNet-R}} & \rotatebox{90}{\textbf{Country211}} & \rotatebox{90}{\textbf{Average (11 subset)}} & \rotatebox{90}{\textbf{Average (19 datasets})}\\
    \midrule
    \rotatebox{0}{\textbf{ZS-CLIP}} & 55.56 & 73.10 & 40.58 & 85.92 & 78.98 & 60.31 & 59.11 & 16.71 & 30.56 & 56.33 & 41.31 & 62.89 & 74.11 & 81.82 & 41.01 & 26.83 & 35.42 & 59.34 & 13.42 & 56.41 & 52.27 \\
    \midrule
    \rotatebox{0}{\textbf{\textit{SuS-X-SD-P}}} & 61.41 & 74.68 & 43.45 & 89.57 & 80.46 & 61.64 & 62.96 & 18.84 & 36.20 & 57.19 & 48.90 & 66.18 & 77.45 & 85.17 & 48.76 & 37.11 & 36.05 & 61.69 & 14.26 & 60.57 & 55.89 \\
    \midrule
    \rotatebox{0}{\textbf{\textit{SuS-X-SD-C}}} & 61.51 & 74.65 & 43.53 & 89.53 & 80.50 & 61.65 & 62.95 & 19.11 & 36.36 & 57.18 & 48.84 & 66.26 & 77.53 & 85.17 & 48.35 & 37.27 & 35.88 & 61.69 & 14.25 & 60.59 & 55.91 \\
    \midrule
    \rotatebox{0}{\textbf{\textit{SuS-X-LC-P}}} & 61.49  & 74.62 & 44.30 & 89.57 & 80.56 & 61.80 & 63.02 & 20.04 & 36.75 & 57.19 & 48.81 & 66.87 & 77.36 & 85.31 & 47.87 & 37.49 & 36.25 & 61.62 & 14.20 & 60.73 & 56.01 \\
    \midrule
    \rotatebox{0}{\textbf{\textit{SuS-X-LC-C}}} & 60.51 & 74.61 & 44.07 & 89.49 & 80.59 & 61.53 & 62.94 & 19.23 & 36.25 & 57.05 & 49.02 & 66.83 & 77.35 & 82.27 & 47.04 & 36.78 & 35.76 & 60.91 & 14.21 & 60.09 & 55.60 \\
    \bottomrule
    \end{tabular}
    \label{tab:fixed-hparams}
\end{table*}

\noindent\textbf{Analysis of hyperparameters.} From~\cref{tab:hparams}, we note that for some datasets, the weight for the inter-modal distance term $\gamma$ is dominated by the weight for the intra-modal distance term $\alpha$. We analyse this in depth, and show that despite this disparity, using inter-modal distances still brings gains.~\cref{tab:hparam-q} reports results on these datasets (for which ${\alpha}{>>}{\gamma}$) using their optimal hyperparameters (${\alpha}{>}{\gamma}$), fixed hyperparameters (${\alpha}{=}{\gamma}{=}{0.1}$), and removed inter-modal contributions (${\gamma}{=}{0}$). In most cases, it is beneficial to use small inter-modal distance contributions over neglecting them (see \textcolor{LimeGreen}{green rows}). Hence, we conclude that both these terms are important for bringing the large performance gains of our model.

\begin{table}[h]
\centering
\caption{\textbf{Analysis of $\alpha$ and $\gamma$ values.}}
\begin{tabular}{c|ccc}
\toprule
\multirow{2}{*}{\textbf{Dataset}} & \textbf{Optimal} & \textbf{Fixed\&Equal} & \textbf{Inter-modal only}\\
& ${\alpha}{>}{\gamma}$ & ${\alpha}{=}{\gamma}{=}{0.1}$ & ${\alpha}{=}{\text{optimal}},{\gamma}{=}{0}$ \\
\midrule    
\rowcolor{LimeGreen}ImageNet & 61.89 & 61.80 & 61.30 \\
\rowcolor{LimeGreen}ImageNet-Sketch & 37.83 & 36.25 & 36.10 \\
\rowcolor{LimeGreen}ImageNet-R & 62.10 & 61.62 & 61.30 \\
\rowcolor{LimeGreen}OxfordPets & 86.59 & 85.31 & 85.00 \\
Birdsnap & 38.50 & 36.75 & 37.70 \\
Food101 & 77.62 & 77.53 & 77.55 \\
\bottomrule
\end{tabular}
\label{tab:hparam-q}
\end{table}

\section{Discussion on \textit{SuS} vs CuPL/VisDesc}
As discussed in the main paper, CuPL and VisDesc are two \textit{name-only} transfer methods that leverage a large pre-trained language model (GPT-3) to enhance the textual prompts used for zero-shot classification. On the other hand, our \textit{SuS} construction strategies endow the zero-shot model with rich visual information to discriminate between different categories.

\begin{figure}[h]
    \centering
    \includegraphics[scale=0.4]{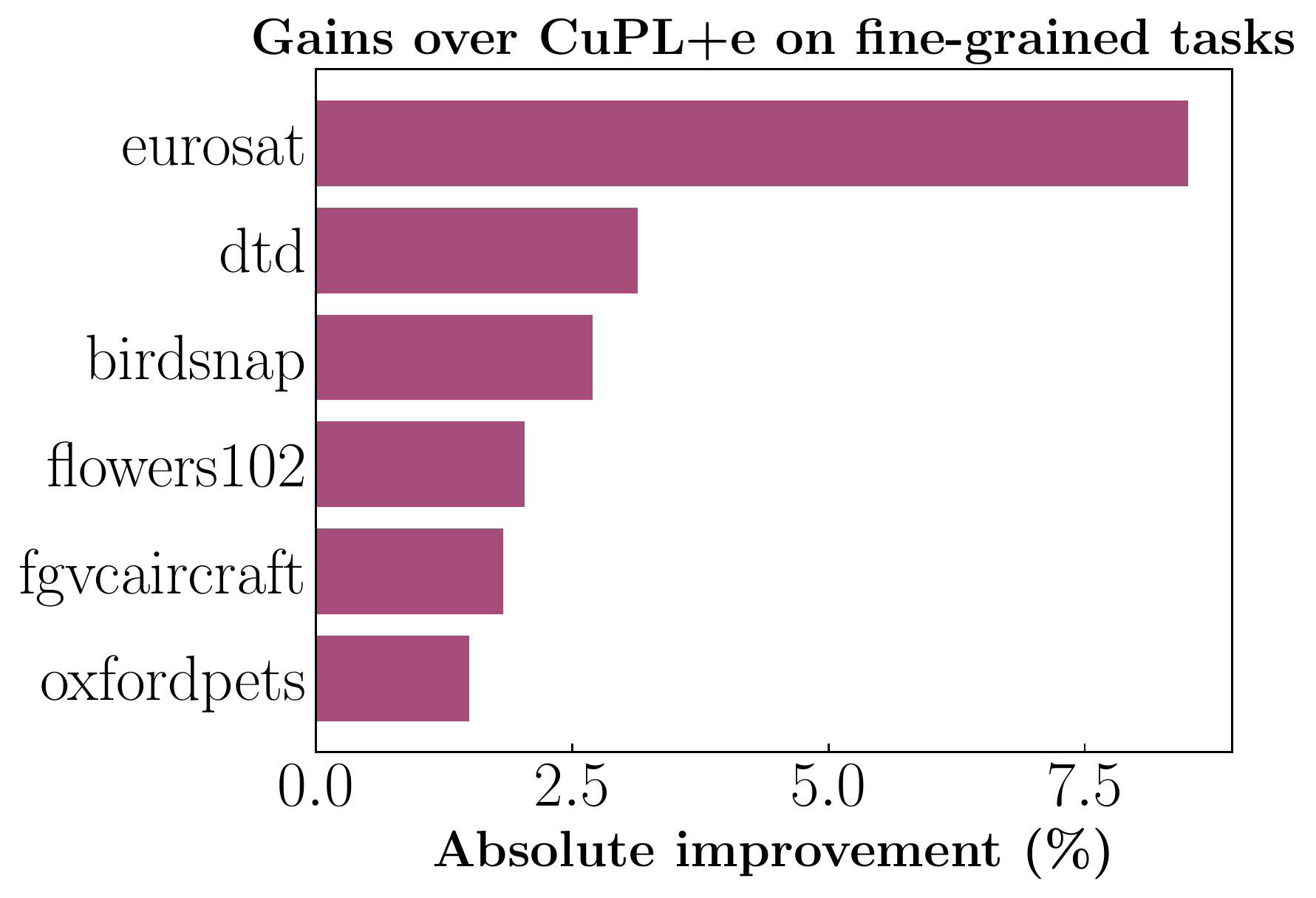}
    \caption{ \textbf{Improvement for fine-grained tasks using \textit{SuS}.}}
    \label{fig:susx-fine-grained}
\end{figure}

We note that text alone cannot model the rich information in the world~\cite{collier2022reality,browning2022ai}. Consider a task of classifying between two bird species---``Florida Scrub Jay'' and ``Blue Jay''. The difference is all in the subtle visual details---blue jays have a crest and distinct black markings on their necks. This level of rich visual information is hard to extract from textual descriptions of class names. Hence, the main advantage of \textit{SuS} is in imparting this expressive visual information for discriminating between fine-grained categories. We verify this empirically in~\cref{fig:susx-fine-grained} depicting large gains in fine-grained datasets like Birdsnap, Flowers102, OxfordPets etc (Full results in~\cref{tab:full-results-dump} below.).

\newpage

\section{Compute Cost Comparison}

We compare the computational requirements of our \textbf{\textit{SuS-X}} and the baselines in~\cref{tab:inference-times}---for each method, we measure the time and memory requirements for one ImageNet class \ie on 50 test images. For CuPL, VisDesc and \textbf{\textit{SuS-X}}, we measure the construction time required for curating the enhanced textual prompts and \textit{support sets}. Note that in practical applications, it is typical to cache the curated \textit{support sets}/prompts for each class, thereby amortising costs across queries. We note that our \textbf{\textit{SuS-X}} models offer the most competetive performance-efficiency tradeoff when comparing the compute requirements and accuracy values.

    \begin{table}[h]
    \centering
    \caption{\textbf{Compute requirements of different methods.} These numbers are with our tests on a single Nvidia A100-80GB GPU with one ImageNet class \ie 50 test images.}
    \begin{tabular}{c|ccc|c}
    \toprule
     \multirow{2}{*}{\textbf{Method}} & \textbf{Construction} & \textbf{Inference} & 
    \textbf{GPU} &
    \textbf{ImageNet}\\
    & \textbf{Time} & \textbf{Time} & \textbf{Memory} & \textbf{Accuracy} \\
    \midrule
    Zero-shot  & -- & 10.22ms & 2.2GB & 60.32 \\
    CALIP & -- & 121.26ms & 24GB & 60.57  \\
    CLIP+DN & -- & 10.22ms & 2.2GB & 60.16  \\
    VisDesc & $\sim$3s & 10.22ms & 2.2GB & 59.68 \\    
    CuPL+e & $\sim$3s & 10.22ms & 2.2GB &  61.64 \\
    \midrule
    \textit{SuS-X-SD} & $\sim$60s & 10.50ms & 3.2GB &  61.84 \\
    \textit{SuS-X-LC} & $\sim$2s & 10.50ms & 3.2GB & 61.89 \\
\bottomrule
    \end{tabular}
    \label{tab:inference-times}            
\end{table}

\section{Diversity of \textit{CuPL} and \textit{Photo} prompting strategies}
In this section, we describe in detail the computation of the diversity metric used in~\cref{prompting-strategies}. 

We assume access to a support set $S$ of size $NC$, where there are $C$ classes and $N$ support samples per class. We denote the support subset of a given class $i$ as $S_{i} = \{s_{i, 1}, s_{i, 2}, \dots, s_{i, N}\}$, where $s_{i, j}$ denotes the $j^{th}$ support sample for class $i$. Corresponding to these support subsets, we denote the features of $S_{i}$ as $F_{i}$ (using CLIP's image encoder):
\[
\begin{gathered}
F_{i, j} = \texttt{CLIPImageEncoder}(s_{i, j}), F_{i, j}\in\mathbb{R}^d,i\in[1, C], j\in[1, N]\\
F_{i} = \texttt{Concat}([F_{i, 1}, F_{i, 2}, \dots, F_{i, N}]), F_{i}\in\mathbb{R}^{N\times d}
\end{gathered}
\]

We now compute the mean pairwise cosine similarity between all support samples within a class \textit{i.e.} for class $i$, we compute:
\[
\texttt{PCS}_{i} = \frac{\sum_{j=1}^{N}\sum_{k=1}^{N}F_{i, j} F_{i, k}^T}{N^2}
\]
The intuition is that if all the support samples within a class are similar to each other, then the support set is less diverse. Hence, a higher value of $\texttt{PCS}_i$ implies a lower diversity. We then compute the mean $\texttt{PCS}$ over all classes as:
\[
\texttt{MPCS} = \frac{\sum_{i=1}^{C}\texttt{PCS}_i}{C}
\]
Finally, we define diversity to be:
\[
\texttt{Diversity} = 1 - \texttt{MPCS}
\]
\newpage

\section{Further Analyses}

We conduct some further ablation studies to analyse our novel {\textit{TIP-X}} method with more rigour. Due to lack of space in the main paper, we include these ablations here, however these are vital analyses which delineate important properties of our method.

\subsection{Contribution of intra-model and inter-modal distances}

In~\cref{method:subsec:tip-x}, we describe our \textit{TIP-X} method that utilises image-text distances as a bridge for computing image-image intra-modal similarities. We refer to the main equation for computing \textit{TIP-X} logits again, highlighting the importance of each term:
\begin{equation*}
\texttt{TXL} = \underbrace{fW^{T}}_{\text{1. zero-shot component}} + \underbrace{\alpha AL}_{\text{2. intra-modal distance component}} + \underbrace{\gamma\psi(-M)L}_{\text{3. inter-modal distance component}}
\end{equation*}

Zero-shot CLIP utilises only the zero-shot term (\textbf{1}) above. TIP-Adapter utilises the zero-shot and intra-modal distance terms (\textbf{1+2}). Our method uses all three terms (\textbf{1+2+3}). We further ablate this design choice to break down the gains brought forth from each individual term. In~\cref{tab:tip-kltip-tipx}, we show the performance gains from each of these terms with our best performing \textbf{\textit{SuS-X-LC}} model across 19 datasets. We observe large gains from inter-modal and intra-modal distances independently over just using the zero-shot term. Further, both these distances provide complementary information to each other, and hence can be productively combined leading to the best results.

\begin{table}[h]
\centering
\caption{\textbf{Contribution of intra-modal and inter-modal distances.}}
\begin{tabular}{c|cccc}
\toprule    
 \multirow{2}{*}{\textbf{Dist. terms used}} & {\textbf{1}} & {\textbf{1+3}} & 
{\textbf{1+2}} &
{\textbf{1+2+3}}\\
& (Zero-shot) & (Inter-modal) & (Intra-modal) & (Both) \\
\midrule
\textbf{Average Acc.} & 52.27 & 56.30 & 56.56 & 56.87 \\
\textbf{Gain} & \textcolor{ForestGreen}{0} & \textcolor{ForestGreen}{+4.03} & \textcolor{ForestGreen}{+4.29} & \textcolor{ForestGreen}{+4.60} \\    
\bottomrule
\end{tabular}
\label{tab:tip-kltip-tipx}
\end{table}

\subsection{Comparing \textit{\textbf{name-only}} \textbf{\textit{SuS-X }} to few-shot methods}

In~\cref{main-results} of the main paper, we showcased SoTA results with our \textbf{\textit{SuS-X}} model in the \textit{name-only} setting. Recollect that in this setting, we use no images from the true target distribution. Here, we evaluate how well our \textbf{\textit{SuS-X}} model fares against methods that use image samples from the true target distribution. We compare our best performing \textbf{\textit{SuS-X-LC}} method (uses no images from target distribution) with 16-shot TIP-Adapter and 16-shot \textit{TIP-X} (both using 16 labelled images per class). From~\cref{tab:susx-vs-tip}, we see that \textbf{\textit{SuS-X-LC}} is competitive (\textcolor{LimeGreen}{in green}) against these few-shot adaptation methods, despite using no target task images. There are however cases where \textbf{\textit{SuS-X-LC}} severely underperforms the few-shot methods---this is due to the domain gap between the \textit{SuS} images and the true labelled images (refer~\cref{analysis}).

\begin{table}[h]
\centering
\caption{\textbf{\textit{SuS-X} against few-shot labelled methods.}}
\begin{tabular}{c|cccc}
\toprule
\multirow{2}{*}{{\textbf{Dataset}}} & \multirow{2}{*}{\textbf{Zero-shot}} & \textbf{\textit{SuS-X-LC}} & \textbf{TIP-Adapter} & \textbf{\textit{TIP-X}}\\
 &  & \footnotesize{(name-only, ours)} & \footnotesize{(few-shot)} & \footnotesize{(few-shot, ours)} \\
\midrule    
\rowcolor{LimeGreen}ImageNet & 60.31 & 61.89 & 62.01 & 62.16 \\
\rowcolor{LimeGreen}Food101 & 74.11 & 77.62 & 75.82 & 75.96 \\
\rowcolor{LimeGreen}OxfordPets & 81.82 & 86.59 & 84.50 & 87.52 \\
\rowcolor{LimeGreen}Caltech101 & 85.92 & 89.65 & 90.43 & 90.39 \\
Flowers102 & 62.89 & 67.97 & 89.36 & 90.54 \\
FGVCAircraft & 16.71 & 21.09 & 29.64 & 29.61 \\
\bottomrule
\end{tabular}
\label{tab:susx-vs-tip}
\end{table}

\subsection{Intuitions for best performing configurations}

From~\cref{tab:photo-vs-cupl-strategies} of the main paper, we note that our best \textit{name-only} results are achieved with the LC-Photo and SD-CuPL \textit{SuS} construction strategies. A natural question arises: ``\textit{Why do the two \textit{SuS} construction methods require different prompting strategies for achieving their best results?}''. We attempt to answer this question via careful inspection of the \textit{support sets} curated from these strategies. For this case study, we inspect the \textit{support sets} for the CIFAR-10 dataset.

\begin{figure*}[h]    
  \subfloat[\textit{SuS-LC}, \textit{Photo}, Airplane]{%
  \begin{minipage}{0.5\linewidth}
  \centering
  \includegraphics[width=0.3\textwidth]{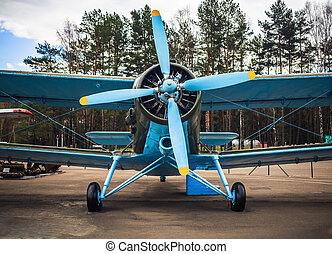}
  \includegraphics[width=0.3\textwidth]{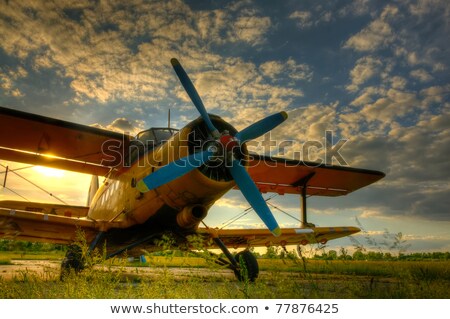}
  \includegraphics[width=0.3\textwidth]{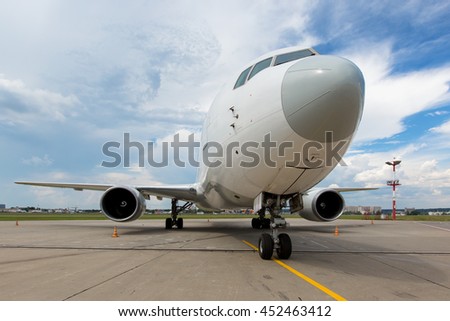}
  \end{minipage}%
  }
  \subfloat[\textit{SuS-LC}, \textit{CuPL}, Airplane]{%
  \begin{minipage}{0.5\linewidth}
  \centering
  \includegraphics[width=0.3\textwidth]{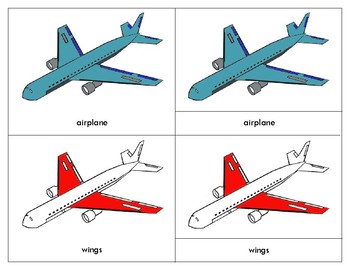}
  \includegraphics[width=0.3\textwidth]{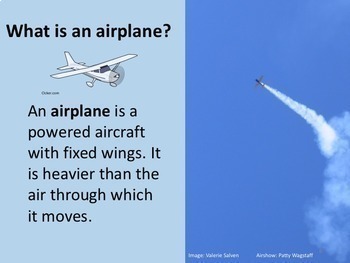}
  \includegraphics[width=0.3\textwidth]{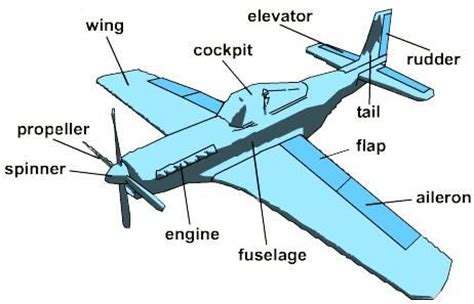}
  \end{minipage}%
  }
  \par
    \subfloat[\textit{SuS-LC}, \textit{Photo}, Bird]{%
  \begin{minipage}{0.5\linewidth}
  \centering
  \includegraphics[width=0.3\textwidth]{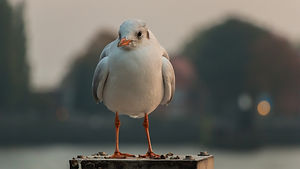}
  \includegraphics[width=0.3\textwidth]{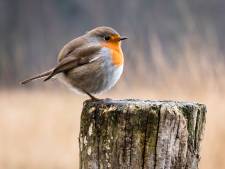}
  \includegraphics[width=0.3\textwidth]{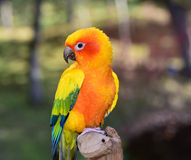}
  \end{minipage}%
  }
  \subfloat[\textit{SuS-LC}, \textit{CuPL}, Bird]{%
  \begin{minipage}{0.5\linewidth}
  \centering
  \includegraphics[width=0.3\textwidth]{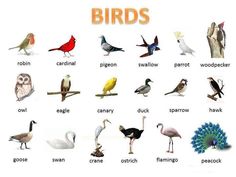}
  \includegraphics[width=0.3\textwidth]{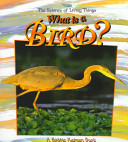}
    \includegraphics[width=0.3\textwidth]{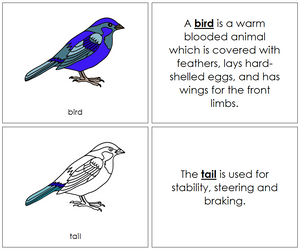}
  \end{minipage}%
  }

  \subfloat[\textit{SuS-SD}, \textit{Photo}, Airplane]{%
  \begin{minipage}{0.5\linewidth}
  \centering
  \includegraphics[width=0.3\textwidth]{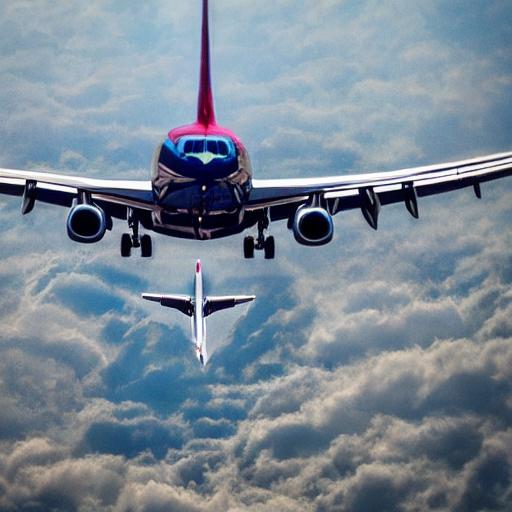}
  \includegraphics[width=0.3\textwidth]{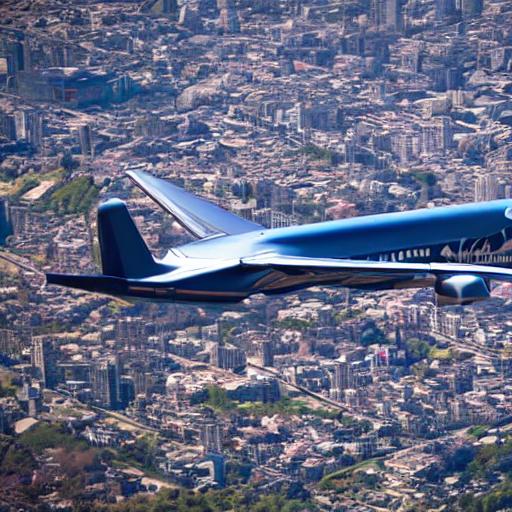}
  \includegraphics[width=0.3\textwidth]{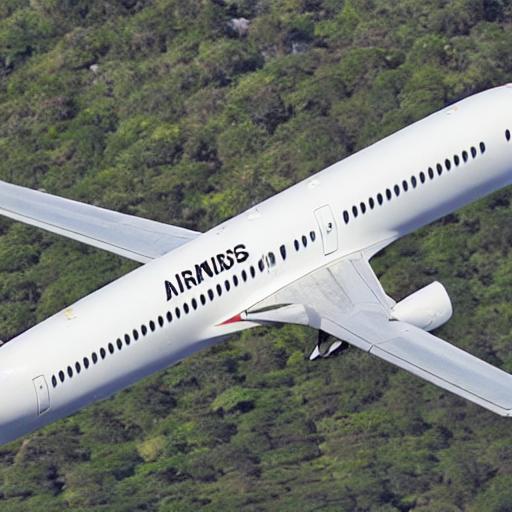}
  \end{minipage}%
  }
  \subfloat[\textit{SuS-SD}, \textit{CuPL}, Airplane]{%
  \begin{minipage}{0.5\linewidth}
  \centering
  \includegraphics[width=0.3\textwidth]{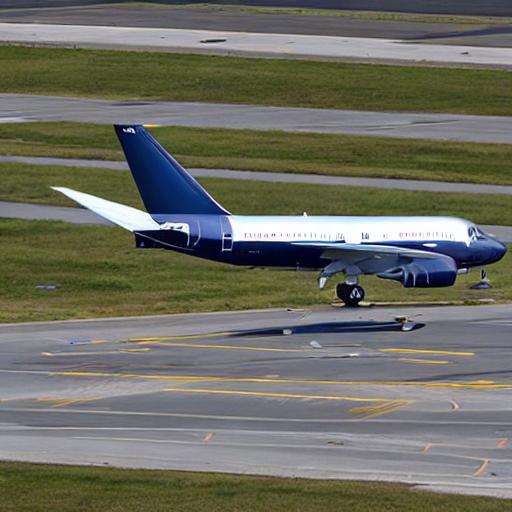}
  \includegraphics[width=0.3\textwidth]{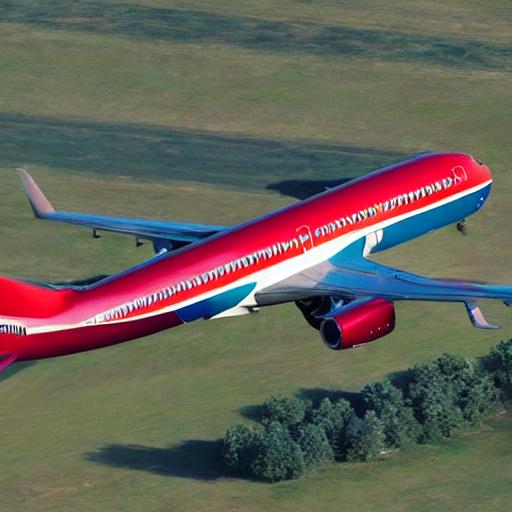}
  \includegraphics[width=0.3\textwidth]{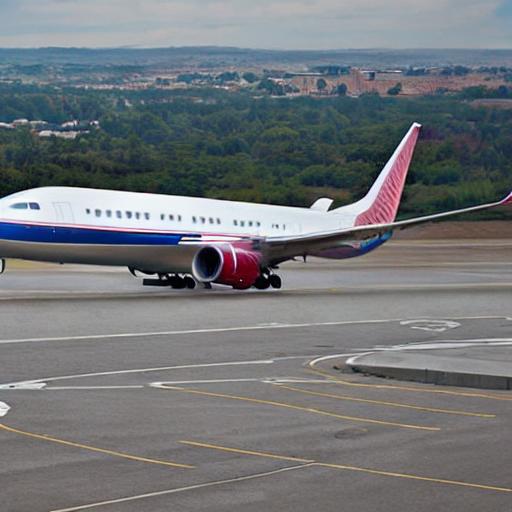}
  \end{minipage}%
  }
  \par
    \subfloat[\textit{SuS-SD}, \textit{Photo}, Bird]{%
  \begin{minipage}{0.5\linewidth}
  \centering
  \includegraphics[width=0.3\textwidth]{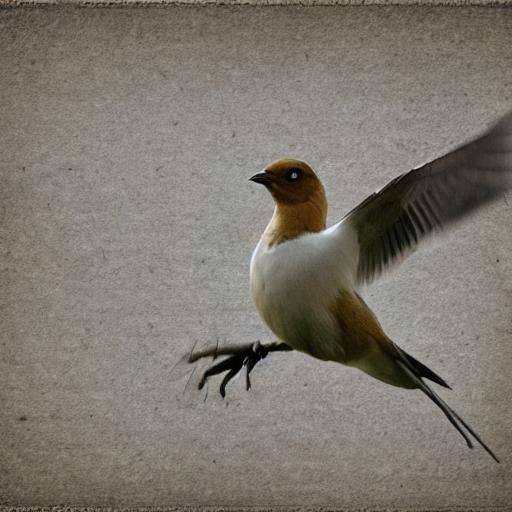}
  \includegraphics[width=0.3\textwidth]{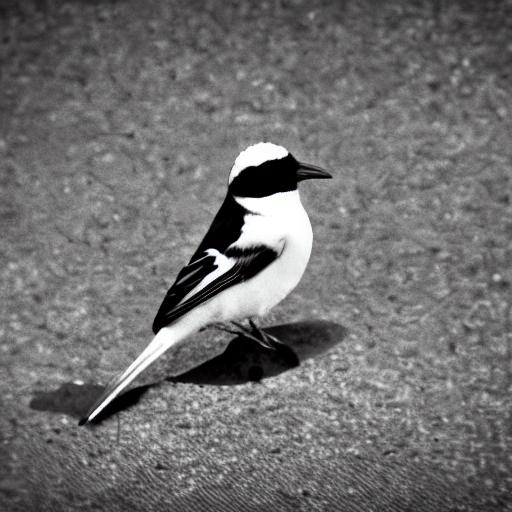}
  \includegraphics[width=0.3\textwidth]{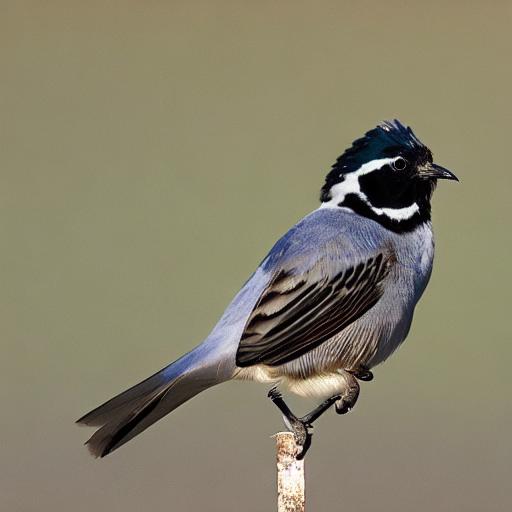}
  \end{minipage}%
  }
  \subfloat[\textit{SuS-SD}, \textit{CuPL}, Bird]{%
  \begin{minipage}{0.5\linewidth}
  \centering
  \includegraphics[width=0.3\textwidth]{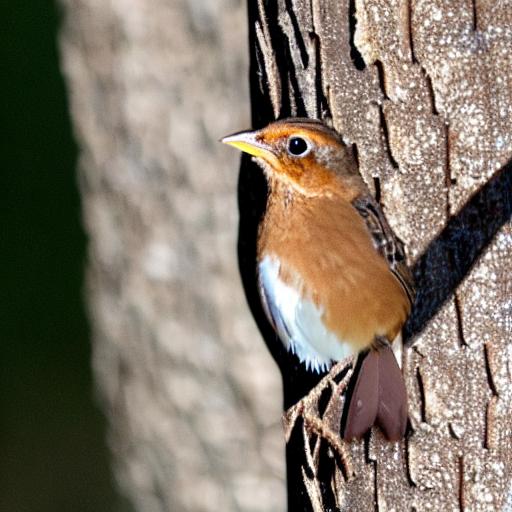}
  \includegraphics[width=0.3\textwidth]{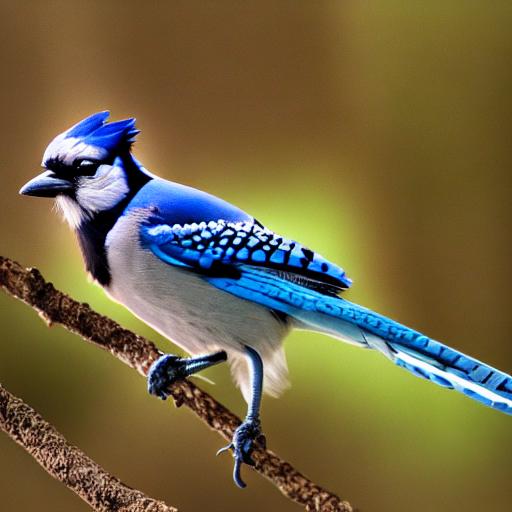}
    \includegraphics[width=0.3\textwidth]{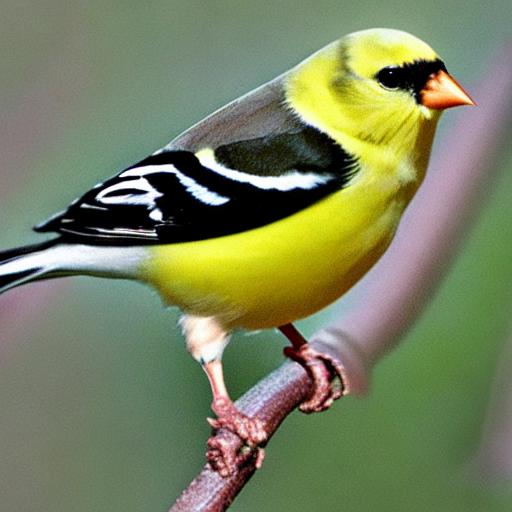}
  \end{minipage}%
  }  
  \caption{\textbf{Uncovering the intuitions for different prompting configurations.} We showcase some support samples using different prompting configurations for two CIFAR-10 classes---\textit{airplane} and \textit{bird}. The key takeaways upon inspecting these samples are enumerated below.}
  \label{fig: sus-intuitions}
\end{figure*}

From~\cref{fig: sus-intuitions}, we can draw two key takeaways regarding the best prompting strategies for the two \textit{SuS} curation methods:
\begin{enumerate}
    \item \textbf{LAION-5B retrieval.} The \textit{support sets} constructed with \textit{CuPL} prompts are largely divergent from the ``true'' distribution of natural semantic images of the target concepts/classes. This can be noted from the right panels of the first two rows in~\cref{fig: sus-intuitions}---this disparity in the retrieved \textit{support set} images leads to a large domain gap to the target distribution, hence resulting in poorer performance than the \textit{Photo} prompting strategy. Further, since the LAION-5B \textit{support sets} consist of natural images \ie images available on the web, the LAION-5B \textit{Photo} \textit{support set} images are closer to the true target distribution of images. 
    \item \textbf{Stable Diffusion Generation.} The \textit{support sets} generated using Stable Diffusion represent a synthetic data distribution \ie there is an innate distribution shift from the target distribution images owing to the target datasets (mostly) consisting of natural images. Hence, the Stable Diffusion \textit{support sets} are inherently at a disadvantage compared to the LAION-5B \textit{support sets}. However, within the constructed Stable Diffusion \textit{support sets}, the \textit{CuPL} prompting strategy mildly wins over the \textit{Photo} strategy since it helps generate a more diverse set of images (consisting of more expansive lighting conditions, background scenes etc.)---this diversity helps reduce the domain gap to the target dataset to a small extent. This phenomenon of added diversity in synthetic datasets aiding downstream performance has also been noted in previous works~\cite{he2022syntheticclipsyn}.
\end{enumerate}

\newpage

\section{Extended Results on all Datasets}

In~\cref{tab:full-results-dump}, we report the accuracies obtained on each of the 19 individual datasets for all our baselines, and our \textbf{\textit{SuS-X}} model variants with CLIP as te VLM. We also report the average accuracy obtained on the 11 dataset subset used in previous CLIP adaptation works~\cite{zhang2022tip, gao2021clip, zhou2021learning}. In~\cref{tab:tcl-full-results-dump}, we report all the results with the TCL model as the VLM, and in~\cref{tab:blip-full-results-dump}, we report the results with the BLIP model as the VLM.

\begin{table*}[h]
    \footnotesize
    \centering
    \setlength\tabcolsep{2pt}
    \caption{\textbf{Training-free zero-shot/name-only results across model configurations and datasets.} We report average results using both the 11 dataset subset used by previous works on few-shot adaptation~\cite{zhang2022tip, gao2021clip,zhou2021learning} and the entire 19 dataset suite. For the CALIP baseline, we report numbers from the original paper (denoted with a subscript o) as well as our re-implementation (denoted with a subscript (r)). We refer to the Zero-shot CLIP model as ZS-CLIP and CuPL+ensemble baseline as CuPL+e. We use the CuPL+ensemble prompts for CLIP's text classifier in this experiment. For both variants of our models, we append P or C to the name to distinguish between \textit{Photo} and \textit{CuPL} prompt strategies. For instance, \textit{SuS-X-LC-P} refers to the \textit{SuS-X} model with LC curation using the \textit{Photo} strategy. All models use the ResNet-50 visual backbone. The best results for each dataset are \textbf{bolded} and the second best are \underline{underlined}. 
    This table contains the full set of values used for generating~\cref{fig:sus-x-zs-comparison} and populating~\cref{tab:main-result-table-with-baselines} in the paper.}
    \begin{tabular}{c|ccccccccccccccccccc|cc}
    \toprule
         & \rotatebox{90}{\textbf{UCF101}} & \rotatebox{90}{\textbf{CIFAR-10}} & \rotatebox{90}{\textbf{CIFAR-100}} & \rotatebox{90}{\textbf{Caltech101}} & \rotatebox{90}{\textbf{Caltech256}} & \rotatebox{90}{\textbf{ImageNet}} & \rotatebox{90}{\textbf{SUN397}} & \rotatebox{90}{\textbf{FGVCAircraft}} & \rotatebox{90}{\textbf{Birdsnap}} & \rotatebox{90}{\textbf{StanfordCars}} & \rotatebox{90}{\textbf{CUB}} & \rotatebox{90}{\textbf{Flowers102}} & \rotatebox{90}{\textbf{Food101}} & \rotatebox{90}{\textbf{OxfordPets}} & \rotatebox{90}{\textbf{DTD}} & \rotatebox{90}{\textbf{EuroSAT}} & \rotatebox{90}{\textbf{ImageNet-Sketch}} & \rotatebox{90}{\textbf{ImageNet-R}} & \rotatebox{90}{\textbf{Country211}} & \rotatebox{90}{\textbf{Average (11 subset)}} & \rotatebox{90}{\textbf{Average (19 datasets})}\\
    \midrule
    \rotatebox{0}{\textbf{ZS-CLIP}} & 55.56 & 73.10 & 40.58 & 85.92 & 78.98 & 60.31 & 59.11 & 16.71 & 30.56 & 56.33 & 41.31 & 62.89 & 74.11 & 81.82 & 41.01 & 26.83 & 35.42 & 59.34 & 13.42 & 56.41 & 52.27 \\
    \midrule
    \rotatebox{0}{\textbf{CALIP$\boldsymbol{_o}$}} & \textbf{61.72} & -- & -- & 87.71 & -- & 60.57 & 58.59 & 17.76 & -- & 56.27 & -- & 66.38 & 77.42 & 86.21 & 42.39 & 38.90 & -- & -- & -- & 59.45 & -- \\
    \midrule
    \rotatebox{0}{\textbf{CALIP$\boldsymbol{_r}$}} & 55.61 & 73.15 & 40.62 & 86.20 & 79.08 & 60.31 & 59.10 & 16.71 & 30.68 & 56.32 & 41.40 & 63.01 & 74.13 & 81.84 & 41.01 & 26.96 & 36.10 & 59.32 & 13.45 & 56.47 & 52.37 \\
    \midrule
    \rotatebox{0}{\textbf{CLIP+DN}} & 55.60 & 74.49 & 43.73 & 87.25 & 79.24 & 60.16 & 59.11 & 17.43 & 31.23 & 56.55 & 43.03 & 63.32 & 74.64 & 81.92 & 41.21 & 28.31 & 35.95 & 60.37 & 13.76 & 56.86 & 53.02 \\
    \midrule
    \rotatebox{0}{\textbf{CuPL}} & 58.97 & 74.13 & 42.90 & 89.29 & 80.29 & 61.45 & 62.55 & 19.59 & 35.65 & \underline{57.28} & 48.84 & 65.44 & 76.94 & 84.84 & 48.64 & 38.38 & 35.13 & 61.02 & 13.27 & 60.30 & 55.50 \\
    \midrule
    \rotatebox{0}{\textbf{CuPL+e}} & 61.45 & 74.67 & 43.35 & 89.41 & 80.57 & 61.64 & \underline{62.99} & 19.26 & 35.80 & {57.23} & 48.77 & 65.93 & 77.52 & 85.09 & 47.45 & 37.06 & 35.85 & 61.17 & \textbf{14.27} & 60.45 & 55.76 \\
    \midrule
    \rotatebox{0}{\textbf{VisDesc}} & 58.47 & 73.22 & 39.69 & 88.11 & 79.94 & 59.68 & 59.84 & 16.26 & 35.65 & 54.76 & 48.31 & 65.37 & 76.80 & 82.39 & 41.96 & 37.60 & 33.78 & 57.16 & 12.42 & 58.30 & 53.76  \\
    \midrule
    \rotatebox{0}{\textbf{\textit{SuS-X-SD-P}}} & \textbf{61.72} & 74.71 & 44.14 & \textbf{89.65} & 80.62 & 61.79 & {62.96} & 19.17 & 36.59 & \textbf{57.37} & \textbf{49.12} & \textbf{67.97} & \underline{77.59} & \underline{86.24} & 49.35 & 38.11 & \underline{36.58} & \textbf{62.10} & \underline{14.26} & 61.08 & 56.32 \\
    \midrule
    \rotatebox{0}{\textbf{\textit{SuS-X-SD-C}}} & \underline{61.54} & 74.69 & \textbf{44.63} & 89.53 & \textbf{80.64} & \underline{61.84} & 62.95 & 19.47 & \underline{37.14} & 57.27 & \textbf{49.12} & \underline{67.72} & 77.58 & 85.34 & \textbf{50.59} &\textbf{ 45.57} & 36.30 & \underline{61.76} & \textbf{14.27} & \textbf{61.76} & \underline{56.73} \\
    \midrule
    \rotatebox{0}{\textbf{\textit{SuS-X-LC-P}}} & 61.49 & \textbf{74.95} & \underline{44.48} & 89.57 & 80.62 & \textbf{61.89} &\textbf{ 63.01} & \textbf{21.09} & \textbf{38.50} & 57.17 & \underline{48.86} & 67.07 & \textbf{77.62} & \textbf{86.59} & 49.23 & \underline{44.23} & \textbf{37.83} & \textbf{62.10} & 14.24 & \underline{61.72} & \textbf{56.87} \\
    \midrule
    \rotatebox{0}{\textbf{\textit{SuS-X-LC-C}}} & 61.43 & \underline{74.76} & 44.12 & \underline{89.61} & \underline{80.63} & 61.79 & 62.94 & \underline{20.34} & 37.07 & 57.06 & \underline{48.86} & 67.60 & 77.58 & 85.22 & \underline{49.47} & 37.16 & 36.45 & 61.39 & \underline{14.26} & 60.93 & 56.20 \\
    \bottomrule
    \end{tabular}
    \label{tab:full-results-dump}
\end{table*}

\begin{table*}[ht]
    \footnotesize
    \centering
    \setlength\tabcolsep{2pt}
    \caption{\textbf{Training-free zero-shot/name-only results across model configurations using the TCL~\cite{yang2022visiontcl} architecture.} For our \textit{SuS-X} models, we only use the two best configurations from the previous CLIP experiment \textit{i.e.} \textit{SuS-X-SD} with \textit{CuPL} strategy and \textit{SuS-X-LC} with \textit{Photo} strategy. This table contains the full set of values used for populating~\cref{tab:other-architecture-results} in the paper.}
    \begin{tabular}{c|ccccccccccccccccccc|cc}
    \toprule
         & \rotatebox{90}{\textbf{UCF101}} & \rotatebox{90}{\textbf{CIFAR-10}} & \rotatebox{90}{\textbf{CIFAR-100}} & \rotatebox{90}{\textbf{Caltech101}} & \rotatebox{90}{\textbf{Caltech256}} & \rotatebox{90}{\textbf{ImageNet}} & \rotatebox{90}{\textbf{SUN397}} & \rotatebox{90}{\textbf{FGVCAircraft}} & \rotatebox{90}{\textbf{Birdsnap}} & \rotatebox{90}{\textbf{StanfordCars}} & \rotatebox{90}{\textbf{CUB}} & \rotatebox{90}{\textbf{Flowers102}} & \rotatebox{90}{\textbf{Food101}} & \rotatebox{90}{\textbf{OxfordPets}} & \rotatebox{90}{\textbf{DTD}} & \rotatebox{90}{\textbf{EuroSAT}} & \rotatebox{90}{\textbf{ImageNet-Sketch}} & \rotatebox{90}{\textbf{ImageNet-R}} & \rotatebox{90}{\textbf{Country211}} & \rotatebox{90}{\textbf{Average (11 subset)}} & \rotatebox{90}{\textbf{Average (19 datasets})}\\
    \midrule
    \rotatebox{0}{\textbf{ZS-TCL}} & 35.29 & 82.33 & 50.86 & 77.65 & 61.90 & 35.55 & 42.12 & 2.25 & 4.51 & 1.53 & 7.63 & 28.30 & 24.71 & 20.63 & 28.55 & 20.80 & 24.24 & 46.05 & 1.42 & 28.84 & 31.38 \\
    \midrule
    \rotatebox{0}{\textbf{CuPL}} & 41.23 & 81.75 & 52.63 & \textbf{81.66} & 65.91 & 41.60 & 49.35 & 3.48 & 6.83 & 2.11 & \textbf{10.20} & 26.10 & 23.62 & 22.15 & 42.84 & 26.30 & 25.67 & 53.61 & \textbf{4.07} & 32.77 & 34.79  \\
    \midrule
    \rotatebox{0}{\textbf{CuPL+e}} & 41.63 & 82.07 & 52.66 & 81.29 & 66.46 & 41.36 & \underline{49.98} & 3.51 & 6.60 & 2.11 & \underline{9.80} & 26.91 & 24.84 & 21.17 & 41.96 & 25.88 & 26.36 & 53.36 & 3.68 & 34.82 & 32.79 \\
    \midrule
    \rotatebox{0}{\textbf{VisDesc}} & 42.53 & 82.30 & 51.89 & 77.00 & 66.51 & 40.40 & \textbf{51.18} & 3.21 & 5.69 & \textbf{2.91} & 8.96 & 25.13 & 27.16 & 24.58 & 34.28 & 21.27 & 27.05 & 49.26 & 3.57 & 31.77 & 33.94 \\
    \midrule
    \rotatebox{0}{\textbf{\textit{SuS-X-SD-C}}} & \underline{47.66} & \underline{82.92} & \underline{55.19} & \underline{81.38} & \underline{66.52} & \underline{52.29} & \underline{49.98} & \underline{9.21} & \underline{13.60} & {2.31} & 9.72 & \textbf{30.98} & \textbf{48.87} & \underline{65.96} & \textbf{48.17 }& \underline{28.75} & \underline{32.22} & \textbf{58.95} & 3.66 & \underline{42.32} & \underline{41.49} \\
    \midrule
    \rotatebox{0}{\textbf{\textit{SuS-X-LC-P}}} & \textbf{50.28} & \textbf{83.14} & \textbf{57.47} & \underline{81.38} & \textbf{66.80} & \textbf{52.77} & {49.97} & \textbf{10.98} & \textbf{17.93} & \underline{2.57} & 9.77 & \underline{30.04} & \underline{48.06} & \textbf{69.96} & \underline{46.63} & \textbf{36.90} & \textbf{36.28} & \underline{57.58} & \underline{3.72} & \textbf{43.59} & \textbf{42.75} \\
    \bottomrule
    \multicolumn{12}{l}{} &  \multicolumn{10}{l}{\footnotesize *We use the official TCL-base checkpoint from \href{https://drive.google.com/file/d/1PtcZF_XzJgIceg4rXLWqGQiXjizvxxS6/view?usp=sharing}{here} for these results.} \\
    \end{tabular}
    \label{tab:tcl-full-results-dump}
\end{table*}

\begin{table*}[ht]
    \footnotesize
    \centering
    \setlength\tabcolsep{2pt}
    \caption{\textbf{Training-free zero-shot/name-only results across model configurations using the BLIP~\cite{li2022blip} architecture.} For our \textit{SuS-X} models, we only use the two best configurations from the previous CLIP experiment \textit{i.e.} \textit{SuS-X-SD} with \textit{CuPL} strategy and \textit{SuS-X-LC} with \textit{Photo} strategy. This table contains the full set of values used for populating~\cref{tab:other-architecture-results} in the paper.}
    \begin{tabular}{c|ccccccccccccccccccc|cc}
    \toprule
         & \rotatebox{90}{\textbf{UCF101}} & \rotatebox{90}{\textbf{CIFAR-10}} & \rotatebox{90}{\textbf{CIFAR-100}} & \rotatebox{90}{\textbf{Caltech101}} & \rotatebox{90}{\textbf{Caltech256}} & \rotatebox{90}{\textbf{ImageNet}} & \rotatebox{90}{\textbf{SUN397}} & \rotatebox{90}{\textbf{FGVCAircraft}} & \rotatebox{90}{\textbf{Birdsnap}} & \rotatebox{90}{\textbf{StanfordCars}} & \rotatebox{90}{\textbf{CUB}} & \rotatebox{90}{\textbf{Flowers102}} & \rotatebox{90}{\textbf{Food101}} & \rotatebox{90}{\textbf{OxfordPets}} & \rotatebox{90}{\textbf{DTD}} & \rotatebox{90}{\textbf{EuroSAT}} & \rotatebox{90}{\textbf{ImageNet-Sketch}} & \rotatebox{90}{\textbf{ImageNet-R}} & \rotatebox{90}{\textbf{Country211}} & \rotatebox{90}{\textbf{Average (11 subset)}} & \rotatebox{90}{\textbf{Average (19 datasets})}\\
    \midrule
    \rotatebox{0}{\textbf{ZS-BLIP}} & 50.49 & 86.68 & 61.72 & 92.13 & 82.17 & 50.59 & 54.22 & 5.40 & 10.21 & 54.71 & 14.95 & 40.15 & 54.21 & 59.04 & 44.68 & 44.10 & 43.69 & 70.93 & 5.84 & 49.97 & 48.73 \\
    \midrule
    \rotatebox{0}{\textbf{CuPL}} & 56.09 & 86.06 & 61.99 & \textbf{92.41 }& 83.45 & 52.96 & 59.16 & 5.85 & 12.24 & 54.64 & 18.53 & \underline{43.97} & 56.14 & 72.00 & 52.95 & 39.37 & 44.83 & 72.27 & 6.26 & 53.23 & 51.11  \\
    \midrule
    \rotatebox{0}{\textbf{CuPL+e}} & 55.61 & 86.33 & 62.16 & 92.29 & 83.59 & 53.07 & 59.38 & 6.27 & 12.18 & \underline{54.89} & 18.63 & 43.72 & 57.10 & 71.73 & 53.30 & 41.48 & 45.34 & 72.40 & 6.42 &  53.53 & 51.36 \\
    \midrule
    \rotatebox{0}{\textbf{VisDesc}} & 53.42 & 86.78 & 60.47 & 92.04 & 81.53 & 50.94 & 55.85 & 6.30 & 11.69 & 54.64 & 16.65 & 42.71 & 58.50 & 69.22 & 47.45 & 42.25 & 43.30 & 68.62 & 6.01 & 52.12 & 49.91 \\
    \midrule
    \rotatebox{0}{\textbf{\textit{SuS-X-SD-C}}} & \underline{57.28} & \underline{87.56} & \underline{63.60} & \underline{92.33} & \textbf{83.66} & \underline{55.93} & \textbf{59.46} & \underline{10.14} & \underline{16.95} & \underline{54.89} & \underline{18.95} & \textbf{44.38} & \underline{62.75} & \underline{74.68} & \textbf{56.15} & \underline{45.36} & \underline{46.51} & \textbf{73.85} & \textbf{6.45} & \underline{55.76} & \underline{53.20} \\
    \midrule
    \rotatebox{0}{\textbf{\textit{SuS-X-LC-P}}} & \textbf{59.90} & \textbf{88.28} & \textbf{64.43 }& 92.29 & \underline{83.61} & \textbf{56.75} & \underline{59.39}  & \textbf{11.82} & \textbf{23.78} & \textbf{54.94}  & \textbf{19.24} & \underline{43.97} & \textbf{64.14} & \textbf{79.72} & \underline{55.91} & \textbf{51.62} & \textbf{48.53} & \underline{73.42} & \underline{6.44} &\textbf{ 57.31} & \textbf{54.64} \\
    \bottomrule
    \multicolumn{12}{l}{} &  \multicolumn{10}{l}{\footnotesize *We use the official BLIP-base checkpoint from \href{https://storage.googleapis.com/sfr-vision-language-research/BLIP/models/model_base_retrieval_coco.pth}{here} for these results.} \\    
    \end{tabular}
    \label{tab:blip-full-results-dump}
\end{table*}

\newpage
\section{Results with different Visual Backbones}

All our main results use the ResNet-50~\cite{he2016deep} visual backbone for CLIP's image encoder. In~\cref{tab:vi-backbones}, we compare the accuracies obtained on all 19 datasets using 2 different visual backbone model classes---ResNets~\cite{he2016deep} (ResNet-50, ResNet-101) and Vision Transformers~\cite{dosovitskiy2020image} (ViT-B/32, ViT-B/16). We observe that the accuracy values monotonically improve as we increase the model capacity.

\begin{table}[ht]
    \scriptsize
    \centering
    \setlength\tabcolsep{2pt}    
    \caption{\textbf{Training-free name-only results across visual backbones.} For this experiment, we use the default versions of our \textbf{\textit{SuS-X}} models: \textbf{\textit{SuS-X-LC}} with \textit{Photo} strategy and \textbf{\textit{SuS-X-SD}} with \textit{CuPL} strategy. This experiment uses the CuPL prompts for CLIP's text classifier. 
    This table contains the raw data for generating~\cref{fig:visual-backbone-ablation} of the paper.}
    \label{tab:vi-backbones}    
    \begin{tabular}{cc|ccccccccccccccccccc|cc}
    \toprule
         & & \rotatebox{90}{\textbf{UCF101}} & \rotatebox{90}{\textbf{CIFAR-10}} & \rotatebox{90}{\textbf{CIFAR-100}} & \rotatebox{90}{\textbf{Caltech101}} & \rotatebox{90}{\textbf{Caltech256}} & \rotatebox{90}{\textbf{ImageNet}} & \rotatebox{90}{\textbf{SUN397}} & \rotatebox{90}{\textbf{FGVCAircraft}} & \rotatebox{90}{\textbf{Birdsnap}} & \rotatebox{90}{\textbf{StanfordCars}} & \rotatebox{90}{\textbf{CUB}} & \rotatebox{90}{\textbf{Flowers102}} & \rotatebox{90}{\textbf{Food101}} & \rotatebox{90}{\textbf{OxfordPets}} & \rotatebox{90}{\textbf{DTD}} & \rotatebox{90}{\textbf{EuroSAT}} & \rotatebox{90}{\textbf{ImageNet-Sketch}} & \rotatebox{90}{\textbf{ImageNet-R}} & \rotatebox{90}{\textbf{Country211}} & \rotatebox{90}{\textbf{Average (11 subset)}} & \rotatebox{90}{\textbf{Average (19 datasets})}\\
    \midrule 
    \multirow{2}{*}{\textbf{RN50}} & \textit{SuS-X-LC} & 59.98 & 74.79 & 44.22 & 89.29 & 80.29 & 61.66 & 62.70 & 21.87 & 38.56 & 56.92 & 48.90 & 66.91 & 77.21 & 86.35 & 50.06 & 43.99 & 37.25 & 61.97 & 13.21 & 61.54 & 56.64 \\
    & \textit{SuS-X-SD} & 59.48 & 74.21 & 44.33 & 89.25 & 80.27 & 61.65 & 62.58 & 19.92 & 37.00 & 57.14 & 49.10 & 67.32 & 77.02 & 85.09 & 51.00 & 47.69 & 37.25 & 61.73 & 13.30 & 61.65 & 56.59 \\
    \midrule
    \multirow{2}{*}{\textbf{RN101}} & \textit{SuS-X-LC} & 60.03 & 77.51 & 46.72 & 92.09 & 81.96 & 62.11 & 61.50 & 22.92 & 39.87 & 61.20 & 45.82 & 59.28 & 78.52 & 88.44 & 51.18 & 39.23 & 40.05 & 69.07 & 11.45 & 61.50 & 57.31 \\
    & \textit{SuS-X-SD} & 57.84 & 76.97 & 46.01 & 92.09 & 81.96 & 62.18 & 61.61 & 21.66 & 35.60 & 61.05 & 45.93 & 60.90 & 78.41 & 86.56 & 51.95 & 39.23 & 40.47 & 68.94 & 11.41 & 61.23 & 56.88 \\
    \midrule
    \multirow{2}{*}{\textbf{ViT-B/32}} & \textit{SuS-X-LC} & 63.49 & 89.32 & 65.25 & 93.18 & 84.73 & 64.73 & 65.49 & 23.01 & 40.77 & 61.19 & 53.03 & 68.01 & 80.31 & 87.95 & 52.25 & 53.91 & 43.10 & 70.55 & 14.91 & 64.87 & 61.85 \\
    & \textit{SuS-X-SD} & 63.20 & 88.39 & 64.84 & 93.18 & 84.73 & 64.71 & 65.47 & 21.66 & 38.97 & 61.12 & 53.52 & 68.17 & 80.24 & 86.81 & 51.89 & 53.91 & 43.27 & 70.42 & 14.91 & 64.58 & 61.55 \\
    \midrule
    \multirow{2}{*}{\textbf{ViT-B/16}} & \textit{SuS-X-LC} & 66.72 & 90.94 & 68.66 & 93.91 & 87.41 & 70.00 & 67.85 & 30.51 & 47.71 & 65.90 & 56.96 & 73.08 & 86.08 & 91.58 & 55.32 & 58.06 & 49.34 & 78.20 & 19.19 & 69.00 & 66.18 \\
    & \textit{SuS-X-SD} & 66.59 & 89.88 & 68.47 & 93.96 & 87.45 & 69.88 & 67.73 & 28.68 & 45.53 & 66.13 & 57.11 & 73.81 & 86.08 & 90.57 & 54.55 & 57.49 & 49.51 & 78.22 & 19.28 & 68.68 & 65.84 \\
    \midrule    
\end{tabular}
\end{table}

\section{Results with different Text-to-Image Generation Models}
We also experiment with different text-to-image generation models for \textit{support set} generation to showcase the generalisability and robustness of our method's results.~\cref{tab:t2i} depicts \textbf{\textit{SuS-X-SD}} results by generating \textit{support sets} using different text-to-image generation models. The results presented in the main paper all use the Stable-Diffusion-v1.4 model, but we also note similar gains across three other generative models.

\begin{table}[h]
\centering
\caption{\textbf{\textit{SuS-X-SD} Results with additional T2I models.}}
\begin{tabular}{c|cccc|c}
\toprule
\textbf{T2I Model} & \textbf{ImageNet} & \textbf{EuroSAT} & \textbf{DTD} & \textbf{OxfordPets} & \textbf{Average}\\
\midrule    
\textcolor{gray}{\textbf{ZS-CLIP (baseline)}} & \textcolor{gray}{60.31} & \textcolor{gray}{26.83} & \textcolor{gray}{41.01} & \textcolor{gray}{81.82} & \textcolor{gray}{52.49}\\
\midrule

\textbf{\href{https://huggingface.co/CompVis/stable-diffusion-v1-4}{StableDiffusion-1.4}} (from main paper) & \textbf{61.84} & 45.57 &  50.59 & 85.34 & {60.84} \textcolor{ForestGreen}{(+8.35\%)} \\

\midrule

\textbf{\href{https://huggingface.co/spaces/ai-forever/Kandinsky2.1}{Kandinsky2.1}} & 61.83 & 44.96 & 49.17 & \textbf{85.47} & 60.36 \textcolor{ForestGreen}{(+7.87\%)} \\

\textbf{\href{https://huggingface.co/prompthero/openjourney-v4}{OpenJourney-4}} & 61.81 & 45.00 & \textbf{50.71} & 85.17 & 60.67 \textcolor{ForestGreen}{(+8.18\%)} \\
\textbf{\href{https://huggingface.co/darkstorm2150/Protogen_v2.2_Official_Release}{Protogen-2.2}} & 61.82 & \textbf{48.67} & 50.35 & 85.26 & \textbf{61.52} \textcolor{ForestGreen}{(+9.03\%)} \\

\bottomrule
\end{tabular}
\label{tab:t2i}
\end{table}

\section{Fine-tuning \textit{SuS-X}}
Despite our work's main focus being the training-free adaptation regime, we explore some preliminary results with fine-tuning \textbf{\textit{SuS-X}} on a few datasets. We compare both the training-free and the fine-tuned variants of \textbf{\textit{SuS-X}} with other CLIP adaptation methods that use full or partial (parameter-efficient fine-tuning) in~\cref{full-ft}. We note for some datasets, full/partial fine-tuning methods perform better than training-free \textbf{\textit{SuS-X}}. However, due to the domain gap between StableDiffusion/LAION-5B curated data and real test data, the gains are not large (confirming prior work~\cite{he2022syntheticclipsyn,sariyildiz2023fake}).
Further, we note that full fine-tuning and \textbf{\textit{SuS-X}} are complementary, allowing a large boost in performance for \textbf{\textit{SuS-X-F}}. On the other hand, we emphasise that the goal of our work is to keep the approach \textit{flexible} and \textit{scalable}---one can apply \textbf{\textit{SuS-X}} to an arbitrary number of rare categories \textit{without training}. This training-free approach can particularly benefit when the categories of interest vary frequently, \textit{rendering repetitive fine-tuning inefficient}. Moreover, fine-tuning forces the model to fit a very specific task distribution, enforcing forgetting of the model's pre-trained performance on a wide array of tasks. Since \textbf{\textit{SuS-X}} only requires target task class names and does not fine-tune the model, we can cache the task-specific \textit{support sets} a-priori and switch them dynamically based on the task at hand, without causing catastrophic forgetting of CLIP's pre-trained knowledge.

\begin{table}[h]
\centering
\caption{\textbf{Fine-tuning methods vs \textbf{\textit{SuS-X}}.}}
\begin{tabular}{c|c|ccc|cc}
\toprule
\multirow{2}{*}{\textbf{Method}} & \textcolor{gray}{{ZS-CLIP}} & {FT-CLIP} & {CoOp~\cite{zhou2022learningcoop}} & {CLIP-Adapter~\cite{gao2021clip}} & {\textit{\textbf{SuS-X}}} & \textbf{\textit{SuS-X-F}} \\
 & \textcolor{gray}{\scriptsize{(No adaptation)}} & \scriptsize{(Full fine-tuning)} & \scriptsize{(PromptTuning)} & \scriptsize{(Adapters)} & \scriptsize{\textbf{(Ours)}} & \scriptsize{\textbf{(Ours)}} \\

\midrule    
 
 \textbf{ImageNet} & {\textcolor{gray}{60.31}} & {60.35} & {60.96} & {61.61} & {\underline{61.89}} & {\textbf{63.22}} \\

 \textbf{EuroSAT} & {\textcolor{gray}{26.83}} & {55.37} & {52.12} & \underline{57.00} & {{44.23}} & {{\textbf{59.22}}} \\

 \textbf{DTD} & \textcolor{gray}{41.01} & \underline{50.35} & 45.66 & {49.29} & {49.23} & {\textbf{52.30}} \\

 \textbf{OxfordPets} & \textcolor{gray}{81.82} & 84.51 & {85.99} & 85.06 & \underline{86.59} & \textbf{87.77} \\     
 
\bottomrule
\end{tabular}
\label{full-ft}
\end{table}

\end{document}